\PassOptionsToPackage{table,dvipsnames}{xcolor}
\documentclass{applemlr}

\usepackage{amsmath}
\usepackage{enumerate}
\usepackage{algorithm}
\usepackage{algpseudocode}
\usepackage{amsfonts}
\usepackage{amsthm}
\usepackage{cleveref}
\usepackage{diagbox}
\usepackage{colortbl}
\usepackage{amssymb}
\usepackage{xspace}
\usepackage{wrapfig}
\usepackage{adjustbox}
\usepackage{tabularx}
\usepackage{booktabs}
\usepackage{mathtools}
\usepackage{tikz}
\usepackage{enumitem}
\usepackage{silence}
\usepackage{dsfont}
\usepackage[table]{xcolor}
\usepackage[dvipsnames]{xcolor}
\usepackage{multirow}
\usepackage{makecell}
\usepackage{xfakebold}
\usepackage{amsmath,amsfonts,bm}

\def\eqref#1{equation~\ref{#1}}
\def\1{\bm{1}}

\DeclareMathAlphabet{\mathsfit}{\encodingdefault}{\sfdefault}{m}{sl}
\SetMathAlphabet{\mathsfit}{bold}{\encodingdefault}{\sfdefault}{bx}{n}

\newcommand{\E}{\mathbb{E}}

\definecolor{textgray}{HTML}{6E6E73}
\usetikzlibrary{positioning, calc}
\usetikzlibrary{decorations.pathmorphing}

\makeatletter
\patchcmd{\wrong@fontshape}{\@gobbletwo}{}{}{}
\makeatother
\numberwithin{equation}{section}
\makeatletter
\AtBeginDocument{
  \urlstyle{sf}
  
}
\makeatother

\definecolor{light}{RGB}{125, 125, 125}
\crefname{tcb@cnt@pbox}{code}{code}
\Crefname{tcb@cnt@pbox}{Code}{Code}
\crefname{assumption}{assumption}{assumption}
\Crefname{assumption}{Assumption}{Assumptions}

\newtcolorbox[auto counter]{pbox}[2][]{
  colback=white,
  title=Code~\thetcbcounter: #2,
  #1,fonttitle=\sffamily,
  fontupper=\sffamily,
  arc=2pt,
  colframe=bgcolor,
  coltitle=fgcolor,
  colbacktitle=bgcolor,
  toptitle=0.25cm,
  bottomtitle=0.125cm
}

\makeatletter
\newcommand\applefootnote[1]{%
  \begingroup
  \renewcommand\thefootnote{}%
  \renewcommand\@makefntext[1]{\noindent##1}%
  \footnote{#1}%
  \addtocounter{footnote}{-1}%
  \endgroup
}
\makeatother

\definecolor{cverbbg}{gray}{0.90}

\usepackage{amsmath,amssymb,amsthm,booktabs,graphicx,xcolor,tabularx,array}
\usepackage{tikz,pgfplots}
\pgfplotsset{compat=1.18}
\usetikzlibrary{arrows.meta,positioning,calc,fit,backgrounds}
\usepackage{placeins,microtype}
\usepackage{capt-of,float}
\usepackage{hyperref,url}

\definecolor{joint}{HTML}{9A5CA8}
\definecolor{jointlt}{HTML}{D5A3DC}
\definecolor{ctr}{HTML}{62A9C8}
\definecolor{warm}{HTML}{D47A9F}
\definecolor{corebg}{HTML}{F7EEF8}
\definecolor{pubblue}{HTML}{2F5D9B}

\newcommand{\model}{CoRe}
\newcommand{\oss}[1]{\textcolor{teal!80!black}{#1}}%
\newcommand{\tcite}[1]{{\scriptsize\citep{#1}}}%
\newcommand{\published}[1]{\textcolor{pubblue}{#1}}
\newcommand{\corerow}{\rowcolor{corebg}}
\newcolumntype{Y}{>{\raggedright\arraybackslash}X}

\DeclareRobustCommand{\nrev}[1]{#1}
\DeclareRobustCommand{\polish}[1]{#1}
\DeclareRobustCommand{\edit}[1]{#1}

\title{CoRe: Jointly Learning Contextual Representations for Understanding and Generation}

\author[1,2,\dagger]{Hangke Sui}
\author[1]{Chang Li}
\author[1]{Yuyan Li}
\author[1]{Mohammad Haris Baig}
\author[1]{Daniel Ulbricht}

\affiliation[1]{Apple}
\affiliation[2]{UIUC}

\contribution[\dagger]{Work done during an internship at Apple.}

\abstract{
\polish{We introduce \model{}, in which understanding and generation jointly shape one contextual representation for images and one for text, trained from scratch.
Each representation is read for understanding, conditions the other modality, and is the space in which that modality is generated by a continuous flow from Gaussian noise.
With 383M parameters trained on CC12M and no pretrained language model, text encoder, or VAE at inference, \model{} matches a same-data contrastive-only baseline in global understanding and reaches COCO-30K FID 10.10, comparable to much larger generators built on pretrained components.
In controlled comparisons at equal data and architecture, \model{} generates better in both directions (text to image and image to text) than a representation shaped by understanding alone and then frozen, and follows prompts better than generating in a space that understanding does not shape.
In the other direction, with an attentive segmentation probe, on monocular depth, and when read by a frozen language model, \model{} surpasses the contrastive-only baseline: generation adds information that stronger readouts reach.
Because generation runs inside the understanding representation, frozen real-image probes read and steer intermediate states and the model ranks its own samples without extra training. Because both modalities use continuous flows, both also distill to a few steps.
Our controls indicate why one representation suffices: understanding organizes the representation semantically, while generation makes it retain more information.}

\vspace{-1.3em}
}
\metadata[Correspondence]{\sffamily Hangke Sui: \url{hangkes2@illinois.edu}; Chang Li: \url{chang_li2@apple.com}; Yuyan Li: \url{yli999248@apple.com}}
\begin{document}

\maketitle

\vspace{-6pt}
\begin{figure}[H]\centering
\includegraphics[width=\linewidth]{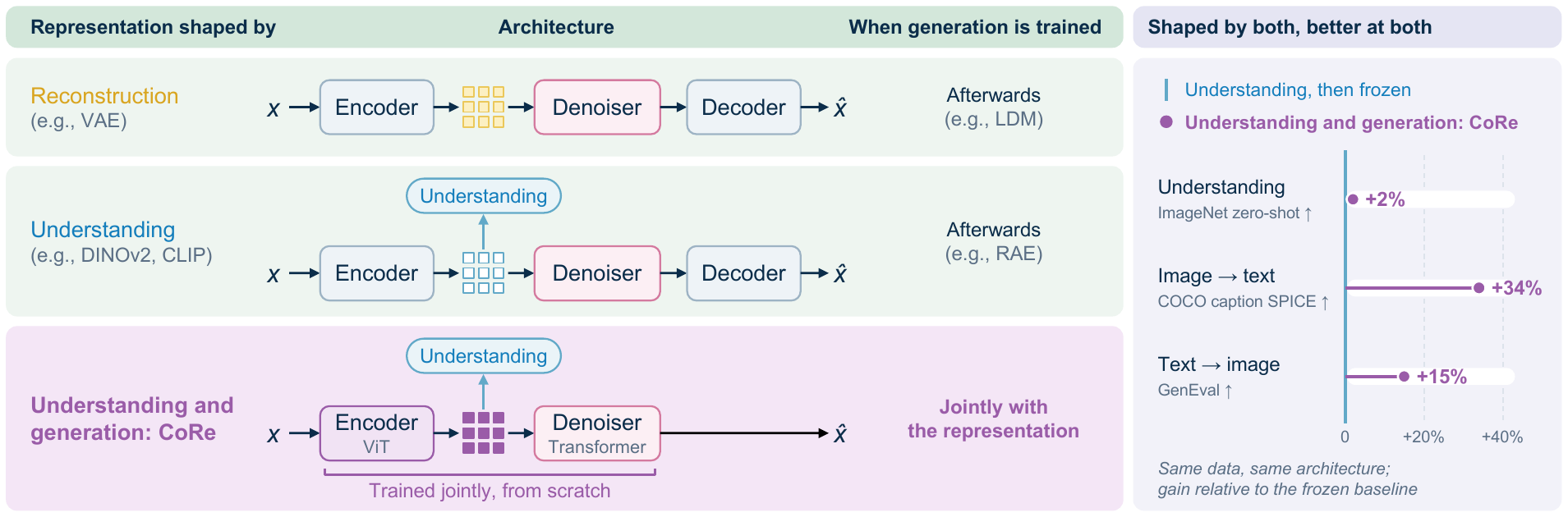}
\vspace{-4pt}
\caption{\textbf{\polish{\model{}: one representation per modality, shaped by understanding and generation together.}} \emph{Left:} \polish{when generation runs in a learned representation space, that space is usually trained beforehand, by reconstruction or by understanding, and then frozen, relying on a separate decoder.} \model{} shapes one contextual representation from scratch, for images and for text, \polish{and generates directly inside it with no separate decoder}. \emph{Right:} at equal data and architecture, joint shaping \polish{preserves global understanding, improves generation in both directions over a frozen understanding representation (Tables~\ref{tab:understanding}--\ref{tab:t2i}), and strengthens understanding under stronger readouts (Section~\ref{sec:analysis})}.}
\label{fig:teaser}\label{fig:paradigms}
\end{figure}

\section{Introduction}
\label{sec:intro}
\begin{center}\vspace{-4pt}\itshape ``What I cannot create, I do not understand.''\upshape\ ---Richard Feynman\vspace{-4pt}\end{center}
Feynman’s remark describes a demanding test for a learned representation. Generating an image \nrev{of one} red hat requires knowing what a hat is, what red is, and what “one” means, and rendering all three
coherently, while describing that image requires recovering the same facts in language. Generation is not the opposite of understanding but one of its most exacting uses. This suggests a representation-level question: \emph{can the same contextual representation support both understanding an observation and generating it?}

Today the two are learned apart. \emph{Understanding} means recovering what an observation contains from its frozen representation: its category, a matching caption, a label for every pixel, or the answer to a question. Such representations are learned by alignment or self-supervision \citep{radford2021clip,dinov2}. \emph{Generation}\polish{, when run in a learned representation space, usually operates in one trained beforehand and then frozen}: a latent shaped by reconstruction \citep{rombach2022ldm} or a representation shaped by understanding \citep{zheng2025rae}. 

We ask instead how the space itself should be learned: can one representation be shaped by understanding and generation together, and does each then benefit? \model{} learns such a representation from scratch, for images and for text, and generates in it. To our knowledge, it is the first model to learn the representation space of both modalities jointly with generation in that space. At equal data and architecture, \model{} preserves understanding and improves both image-to-text and text-to-image generation over a representation shaped by understanding and then frozen (Figure~\ref{fig:paradigms}, right).

\polish{This runs counter to a common view that} the two objectives compete: understanding favors abstract
semantics, whereas generation requires fine detail, motivating decoupled
encoders or separate codebooks
\citep{wu2024janus,qu2024tokenflow,deng2025bagel}. Yet many ``details'' \polish{(color,
texture, boundaries, and spatial relations)} also matter for understanding
\citep{cimpoi2014textures,geirhos2019texture,ibot}, and reconstruction need not
conflict with semantics when the representation has sufficient capacity
\citep{ma2025unitok}. We therefore separate two questions that are often
conflated: \emph{what information a representation retains}, and
\emph{how that information is organized} for downstream readout
\citep{bengio2013representation,xu2020usable}.
\polish{Generation pressures retention because what a decoder cannot recover is lost, while understanding pressures organization by pulling semantically related observations together. Rather than competing, these two pressures can shape a single representation.}

We build this view into \model{} (Figure~\ref{fig:architecture}).
Modality-specific encoders map images and text to contextual patch and token representations, each used in three roles: it is read for understanding, conditions generation of the other modality, and is \nrev{the space} in which that modality is generated. Global
readouts are aligned contrastively, while generation proceeds through
continuous trajectories in those same contextual spaces. Because these
representations are learned jointly and have no externally defined latent
targets, generative learning is anchored primarily to observables \polish{(pixels and
token identities),} with representation regression used only to support sampling.

Two aspects make this construction revealing. 
First, \emph{language is generated in the continuous space in which it is understood}.
Unlike diffusion over lookup embeddings or separately learned latents
\citep{li2022diffusionlm,dieleman2022cdcd,chen2026langflow,meshchaninov2026ldlm},
\model{} generates contextual token representations that are also pooled for
semantic alignment. These representations capture not only token identity
but also meaning in context: the same token, such as \emph{pitcher}, can
represent a baseball player in one caption and a water jug in another.
Generation therefore models tokens through their contextual representations,
rather than through a fixed embedding for each vocabulary entry.
Second, \emph{images are generated in a jointly learned contextual patch space}, rather than in pixels, a pretrained VAE latent, or a frozen representation \citep{rombach2022ldm,yu2024repa,zheng2025rae}. This allows one flow formulation and one conditioning mechanism to serve both image-to-text and text-to-image generation.

We keep the setting deliberately controlled: the encoders and two denoisers, 383M parameters in total, are trained jointly from
scratch on CC12M with public recaptions and alt-text
\citep{captionemporium2024cc12m}, and no pretrained language model, text encoder, or
VAE is part of the inference model. \polish{Unified models built on pretrained LLMs ask how to connect understanding and generation atop existing encoders and decoders, whereas \model{} asks how the representation itself should be formed. Training from scratch isolates this question from anything inherited from pretrained components, yielding design principles that can inform larger-scale pretraining.} \model{} matches a same-data, same-step
contrastive control on global understanding and reaches COCO-30K FID 10.10,
comparable to much larger generators using pretrained T5-XXL
encoders and VAEs \citep{li2026dream,fan2024fluid}. Prompt following and caption fidelity remain below systems with a pretrained language encoder
\citep{minit2i2026} or more caption data.

\textbf{Contributions.}
(i) \emph{One representation, shaped jointly by understanding and generation.} \model{} learns, from scratch and symmetrically for images and text, one contextual representation that is read for understanding, conditions generation, and is the space in which \nrev{the} generation trajectory runs (Section~\ref{sec:method}).
(ii) \emph{\polish{Joint shaping improves generation and strengthens understanding.}} At equal data and architecture, joint shaping improves generation in both directions, preserves global understanding, and leads \polish{on attentive segmentation, monocular depth, and frozen-LLM VQA} (Sections~\ref{sec:results} and~\ref{sec:analysis}).
(iii) \emph{Organization and retention.} \polish{Generating in a space shaped by understanding outperforms an unaligned space that reconstructs better, and the apparent dense linear-probe gap reverses under stronger readouts} (Section~\ref{sec:analysis}).
(iv) \emph{Reading, steering, and ranking generations, with few-step sampling.} Because generation runs in the representation that understanding reads, frozen probes read and steer a generation in progress \polish{(Figure~\ref{fig:steerstrip})} and the model ranks its own samples. Continuous generation in both modalities also distills to a few steps (Sections~\edit{\ref{sec:distillation} and~\ref{sec:trajectory}}).

\section{Related work}
\label{sec:related}
\textbf{Should understanding and generation share a representation?} Unified multimodal models combine understanding and generation through shared sequence models over discrete tokens \citep{ofa2022,wang2024emu3}, mixed autoregressive and diffusion objectives \citep{zhou2024transfusion,deng2025bagel,xie2025showo}, diffusion over both modalities \citep{swerdlow2025unidisc,yang2025mmada,li2024dualdiffusion}, or visual interfaces engineered for both uses \citep{wu2024janus,chen2025januspro,qu2024tokenflow,ma2025unitok,unispace2026}. These systems unify the architecture, the sequence model, or the task interface, usually on top of pretrained language models, VAEs, or vision encoders. We ask a different, representation-level question, with every model component learned from scratch.

\textbf{Which space generation runs in.} Pretrained semantic features improve diffusion as alignment targets or as the latent space itself \citep{yu2024repa,zheng2025rae}, and aligning or jointly tuning \nrev{a pretrained} autoencoder latent makes it easier to generate from\nrev{, whereas diffusion-only latent learning from scratch fails without reconstruction anchoring} \citep{yao2025vavae,leng2025repae}. Conversely, generative and reconstructive training yields transferable features and can sharpen understanding \citep{chen2020igpt,he2022mae}, including encoders trained through generative decoders \citep{preechakul2022diffae,hudson2024soda,chen2024ldae,li2023mage} or by captioning \citep{tschannen2023cappa}. UniDiffuser learns the conditionals and marginals that our routes train, but over frozen latents \citep{bao2023unidiffuser}. DREAM \citep{li2026dream} jointly optimizes contrastive understanding and masked generation over VAE tokens with a frozen T5 encoder. In these approaches generation runs in pixels, an autoencoder latent, or a frozen or separately regularized feature space. In \model{}, it runs in both modalities inside the representation that understanding reads.

\textbf{Continuous diffusion for language.} Continuous diffusion language models corrupt token embeddings or learned latents and decode categories \citep{li2022diffusionlm,dieleman2022cdcd,chen2026langflow,meshchaninov2026ldlm}. Their continuous space serves generation only. In \model{} the continuous text sequence is also the text representation for understanding, pooled for alignment and used to condition images. Continuity also makes flow-map and shortcut distillation applicable, compressing a continuous trajectory into a few jumps \citep{boffi2025flowmap,frans2025shortcut}.

\section{\model{}: jointly learned contextual representations}
\label{sec:method}
\begin{figure}[t]\centering
\includegraphics[width=1.03\linewidth]{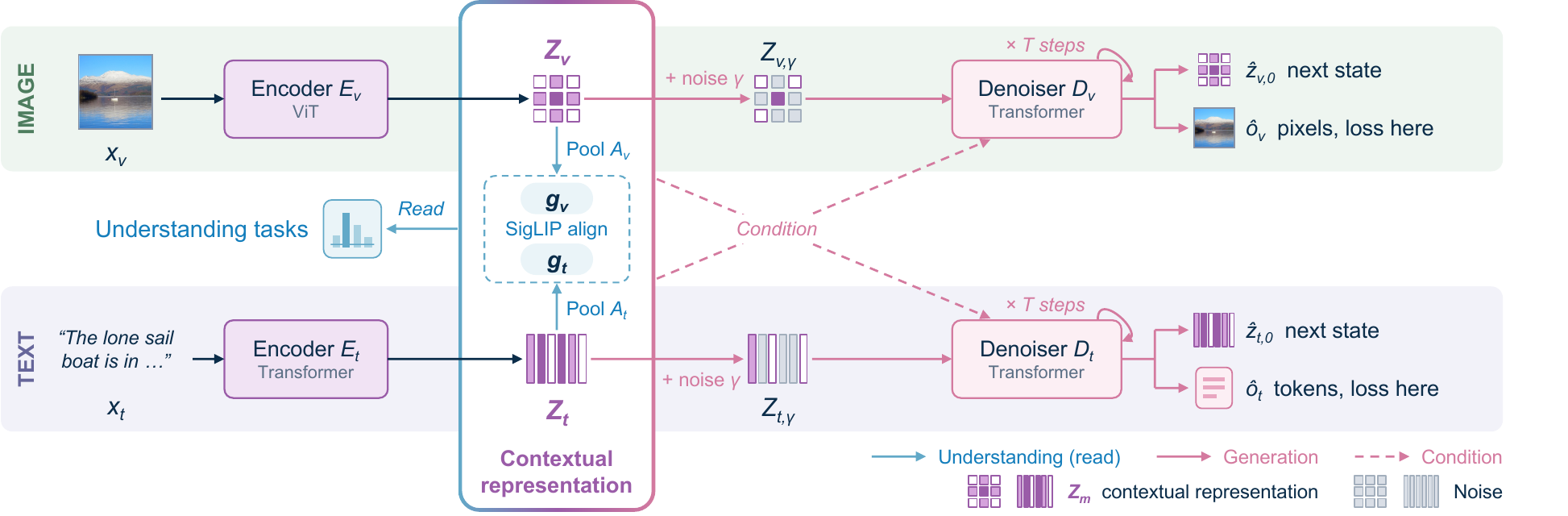}
\caption{
\textbf{One contextual representation per modality, in three roles.} The encoder $E_m$ ($m\in\{v,t\}$) maps an image or a caption to a contextual representation $Z_m$ and its pooled readout $g_m$. \emph{Understanding} reads them, and $g_v$ and $g_t$ are aligned. \emph{Conditioning:} $(g_m,Z_m)$ conditions the denoiser of the other modality. \emph{Generation} runs in $Z_m$ itself: from a noised $Z_m$, the denoiser $D_m$ predicts the clean representation (head $F_m$, the next sampling state) and the observation, pixels or tokens (head $B_m$), which anchors the training loss.}
\label{fig:architecture}
\end{figure}

\model{} learns one contextual representation for each modality, shaped jointly by understanding and generation, and uses it in
three ways: it is read for understanding, conditions generation of the other
modality, and is itself the space in which that modality is generated
(Figure~\ref{fig:architecture}). The key design choice is therefore not merely
to share an encoder, but to make understanding and generation operate on the
same learned representation.

\textbf{One contextual representation per modality.}
For a paired image and caption $(x_v,x_t)$ and modality
$m\in\{v,t\}$,
\begin{equation}
h_m = E_m(x_m),\qquad
Z_m = \operatorname{LN}\!\left(P_m h_m\right)
    \in \mathbb{R}^{N_m\times d},
\qquad
g_m = A_m(Z_m)\in\mathbb{R}^{d_g}.
\label{eq:state}
\end{equation}
Here $E_m$ is a transformer encoder over image patches or text tokens,
$P_m$ projects its features tokenwise to width $d$, and $A_m$ attention-pools
the sequence into a global representation $g_m$. The non-affine layer
normalization fixes the scale of each token relative to the noise used below.
Because $E_m$ uses self-attention, $Z_m$ is \emph{local in index but contextual
in content}: each image entry corresponds to a patch but depends on the full
image, and each text entry corresponds to a token but depends on the full
caption. The global representation $g_m$ is read from this same sequence.
We never align $Z_v$ and $Z_t$ token to token. Only their pooled readouts are aligned.
The image encoder is a 12-layer ViT-B/16 (width 768), the text encoder a 12-layer transformer (width 512), and the two denoisers are transformers of 12 and 6 blocks (Appendix~\ref{app:implementation}).

\textbf{Generation runs in the contextual representation.}
We corrupt the contextual representation itself with variance-preserving
Gaussian noise, parameterized by the log noise-to-signal ratio $\gamma$:
\begin{equation}
Z_{m,\gamma}
=
\alpha_\gamma Z_m+\sigma_\gamma\varepsilon,
\qquad
\alpha_\gamma^2=\operatorname{sigmoid}(-\gamma),\qquad
\sigma_\gamma^2=\operatorname{sigmoid}(\gamma),\qquad
\varepsilon\sim\mathcal N(0,I).
\label{eq:corrupt}
\end{equation}
The clean endpoint of this trajectory is therefore exactly the contextual
representation used for understanding. A modality-specific denoiser predicts
both the underlying observation and the clean representation:
\begin{equation}
H_m=D_m(Z_{m,\gamma},\gamma,C_m),\qquad
\widehat o_m=B_m(H_m),\qquad
\widehat Z_{m,0}=F_m(H_m).
\label{eq:heads}
\end{equation}
The observable head $B_m$ predicts RGB pixels for images and vocabulary logits
for text, while the representation head $F_m$ estimates the clean
representation used by the sampler. Both heads read the denoiser's features at every sampled noise level, so the observable head is trained on the states it meets at sampling time and no decoder is trained separately. 

\textbf{Observable targets anchor a moving representation.} The contextual representation $Z_m$ changes throughout training because its encoder is
learned jointly with the denoiser \nrev{and thus} has no fixed
ground-truth coordinates. If latent regression were the only objective, both
the prediction and its target could move together toward an easier solution
\citep{assran2023ijepa}. We therefore anchor generation primarily to fixed
observables \polish{(pixels and token identities)} and use regression to the detached
$Z_m$ only as auxiliary supervision for sampling. This lets the representation
evolve while keeping its generative meaning tied to what is actually observed. \polish{Together, contrastive alignment organizes $Z_m$ semantically while observable prediction keeps detail recoverable through $B_m$, letting both the space and its readout head emerge in one run without a pretrained autoencoder.}

\textbf{Cross, self, and null conditioning.}
To generate modality $m$ from the other modality $\bar m$, we condition on its contextual representation and pooled readout, $C_m=(g_{\bar m},Z_{\bar m})$: $g_{\bar m}$ enters through adaptive normalization, while $Z_{\bar m}$ supplies cross-attention keys and values. During training, each denoiser uses one of three routes: \emph{cross} ($p=0.7$), \emph{self} ($0.2$), or \emph{null} ($0.1$), with conditions $C_m^{\rm cross}=(g_{\bar m},Z_{\bar m})$, $C_m^{\rm self}=(g_m,\varnothing)$, and $C_m^{\rm null}=(\varnothing,\varnothing)$, where $\varnothing$ is a learned null input. For fixed encoders, paired observations induce a joint distribution over contextual representations,
\begin{equation}
p(Z_v,Z_t)
=
p(Z_v\mid Z_t)\,p(Z_t)
=
p(Z_t\mid Z_v)\,p(Z_v),
\label{eq:jointlaw}
\end{equation}
and the cross and null routes train the corresponding conditional and marginal distributions in both directions, while classifier-free guidance combines them at inference \citep{ho2022cfg}. The self route replaces the local condition with a null input while retaining the target modality's own pooled representation, teaching how much of that modality can be generated from its global summary alone. Thus, the same contextual representations define both the cross-modal conditionals and the unimodal marginals of the learned joint distribution.

\textbf{Joint objective.}
The same contextual representations are learned jointly for alignment and
generation. We optimize
\begin{equation}
\mathcal L
=
\mathcal L_{\rm align}
+
\mathcal L_{\rm img}
+
0.5\,\mathcal L_{\rm text},
\qquad
\mathcal L_{\rm align}
=
\operatorname{SigLIP}(\bar g_v,\bar g_t),
\quad
\bar g_m=\frac{g_m}{\lVert g_m\rVert_2}.
\label{eq:fullobjective}
\end{equation}
The alignment loss encourages paired images and captions to agree through their
pooled semantic representations, while leaving the contextual representations
themselves to be learned jointly \citep{zhai2023siglip}. Generation instead
anchors each contextual representation to its observable:
\begin{align}
\mathcal L_{\rm img}
&=
\mathbb E_{\gamma\sim q_v,\,r}
\Big[
\mathcal L_{\rm MSE}(\widehat o_v,x_v)
+
\mathbf 1[\gamma<\gamma_{\rm p}]
\big(
0.1\,\mathcal L_{\rm LPIPS}
+
0.01\,\mathcal L_{\rm P\text{-}DINO}
\big)
+
0.25\,\mathcal L_{{\rm rep},v}
\Big],
\nonumber\\
\mathcal L_{\rm text}
&=
\mathbb E_{\gamma\sim q_t,\,r}
\Big[
\mathcal L_{\rm CE}(\widehat o_t,x_t)
+
0.25\,\mathcal L_{{\rm rep},t}
\Big],
\label{eq:genlosses}
\end{align}
where the expectations are over sampled noise levels $\gamma$ and conditioning routes $r$, and
\begin{equation}
\mathcal L_{{\rm rep},m}
=
\frac{1}{N_m d}
\left\|
F_m(H_m)-\operatorname{sg}(Z_m)
\right\|_F^2 .
\label{eq:reploss}
\end{equation}
Pixels and token identities provide the fixed observable targets, while the detached
representation regression provides auxiliary supervision for the clean-representation
prediction used by the sampler. \polish{Importantly, gradients from the observable losses flow back into both encoders through the noised input $Z_{m,\gamma}$ and the conditioning pathway $C_m$. Only the regression target $\operatorname{sg}(Z_m)$ is detached, so generative supervision directly shapes $Z_m$ without collapsing it.} For images, frozen VGG16 and DINOv2 provide
low-noise perceptual supervision during training only
\citep{ma2026pixelgen,simonyan2015vgg,dinov2}. Thus, alignment specifies
\emph{which paired observations should agree semantically}, while generation
specifies \emph{what their contextual representations must keep recoverable}.
The visual noise proposal $q_v$ is fixed logit-normal
\citep{esser2024sd3}, and the text proposal $q_t$ follows LangFlow's learned noise
schedule \citep{chen2026langflow}. Remaining implementation details are given in
Appendix~\ref{app:implementation}.

\textbf{Inference.} A single encoder pass produces the contextual representation $Z_m$ and its
pooled readout $g_m$. Understanding tasks read these representations directly.
For cross-modal generation, the same source representation conditions the
denoiser, while the target modality starts from Gaussian noise and is iteratively
denoised in its contextual representation space \polish{using the clean-state estimates from $F_m$}. At the final step, $B_m$ maps the denoiser features to RGB pixels or
token logits. Thus no separate generative latent, such as
a VAE latent, is introduced for generation.

\section{\polish{Experiments: understanding and generation of images and text}}
\label{sec:results}
\polish{At equal data and architecture, shaping one contextual representation jointly by understanding and generation preserves contrastive-level global understanding while improving both image-to-text and text-to-image generation over a frozen understanding encoder. Moreover, both continuous flows distill to a few steps (Tables~\ref{tab:understanding}--\ref{tab:t2i}).} Each benchmark uses a fixed evaluation protocol for all models we evaluate, including released external models.

\textbf{Training setup.} All base models are trained from random initialization for 125K steps at batch size 8192 on CC12M images \citep{changpinyo2021cc12m} with the public LLaVA-NeXT recaptions \citep{captionemporium2024cc12m} and the original alt-text, one caption sampled per image presentation (\emph{CC12M-R}). \model{} has 383M parameters in total.

\textbf{Controlled variants.} We compare variants sharing the same training data and encoder architecture. \emph{Contrastive only} \polish{(understanding only, without generation)} trains the 143M encoders only with the SigLIP objective, providing a baseline without generative training. \emph{No contrastive loss} \polish{(generation only, without alignment)} drops only the alignment loss while retaining all other training losses, testing the role of explicit image--text alignment. \emph{Frozen contrastive $\to$ generation} \polish{(shaped by understanding alone, then frozen)} freezes the contrastive-only encoders and trains the two denoisers for another 125K steps, testing generation from fixed contrastive representations. \model{} + Gram/iBOT adds Gram anchoring and iBOT-style self-distillation on encoder patch features after 24K steps \citep{dinov3,ibot}, testing whether these regularizers improve dense spatial prediction. We also apply these regularizers to \emph{Contrastive only}, allowing comparison with and without generative training under the same structural regularization.

\textbf{Fine-tuning.} For text-to-image generation we fine-tune on a public 120K synthetic mixture \citep{minit2i2026}. The recipe ($^\sharp$ rows) first updates all \model{} + Gram/iBOT parameters on CC12M-R at $512^2$ for 10K steps, then trains only the image denoiser on the mixture for \nrev{76K steps}, preserving the first-stage model's understanding and captioning.

Across the following tables, \published{blue} denotes published results, \oss{teal} denotes our evaluations of released models or outputs, and black denotes our models and controls. ``--'' marks a result that the source does not report or an evaluation that does not apply. \polish{Appendices~\ref{app:implementation}--\ref{app:geometry} provide implementation details, evaluation protocols, external comparison accounting, controlled interventions, extended capability results, few-step distillation, iterative contextual refinement, and representation geometry measurements.}

\subsection{Understanding}
\label{sec:understanding}
Table~\ref{tab:understanding} evaluates \polish{visual recognition, cross-modal matching, and dense prediction, and we also evaluate the frozen text encoder on text-only benchmarks (Table~\ref{tab:textonly}).}

\textbf{Global classification.} ImageNet-1k zero-shot classification uses all 50K validation images and an 80-prompt ensemble, while full-train linear probes (90 epochs, LARS; \citealp{you2017lars}) use the pooled representation $g$ or average-pooled encoder features $h$. At equal data and training steps, \model{} reaches 50.63\% zero-shot and 73.48\% linear-probe accuracy on $g$, against 49.51\% and 72.34\% for the contrastive-only control. With Gram/iBOT, \model{} reaches 50.90\% and 74.31\%, respectively, both within 0.5 percentage points of the best results among the CC12M-trained references (COSMOS: 51.4\% zero-shot, LaSLIP: 74.4\% linear probe).

\textbf{Retrieval and compositional matching.} COCO Karpathy-5K retrieval \citep{lin2014coco} evaluates retrieval in both directions between 5K images and 25K captions, where we report R@1 in each direction and mean recall (mR) averaged over R@1, R@5, and R@10 in both directions. SugarCrepe \citep{hsieh2023sugarcrepe} tests whether image--text similarity prefers a correct caption over its compositional perturbations, reporting accuracy averaged over seven official splits. \model{} reaches 70.34\% mR on COCO retrieval and 80.6\% accuracy on SugarCrepe, compared with 69.75\% and 79.4\%, respectively, for the contrastive-only control. Together with the classification results, these comparisons suggest that adding generation preserves global recognition and image--text matching performance.

\nrev{\textbf{Text representation.} Evaluated without images, \model{}'s frozen text encoder\polish{, trained from scratch on CC12M image-caption pairs,} matches 400M-pair CLIP ViT-B/16 on SICK-R semantic relatedness (69.0 vs.\ 69.1) and exceeds it on caption-to-caption retrieval on COCO (35.5\% vs.\ 35.0\% R@1) and Flickr30k (62.5\% vs.\ 60.9\%; Table~\ref{tab:textonly}). Conditioned on $Z_t$ alone, the text denoiser also reconstructs test captions (BLEU-4 62, retrieving the exact source among 25K captions at 95\% R@1), showing that $Z_t$ organizes sentence meaning while retaining surface phrasing.}

\textbf{Dense spatial prediction.} \polish{On full-resolution ADE20K \citep{zhou2017ade20k}, joint learning lowers the mIoU of a linear probe by 0.9 on frozen $h$ (37.41 to 36.48) and by 5.5 on frozen $Z$ (30.15 to 24.69). The information is still present and is organized differently: an MLP probe closes most of the gap on $Z$, an attentive probe reverses it, and monocular depth favors \model{} under every probe (Section~\ref{sec:analysis}).}

\subsection{Image-to-text generation}
\label{sec:i2t}
Table~\ref{tab:i2t} evaluates captions on three axes: caption quality and hallucination, image-specific discrimination, and detailed coverage. Our models generate one full caption per image using a fixed 64-step DDPM sampling protocol.

\begin{table}[t]\centering
\caption{\textbf{Understanding} (higher is better). $g$: pooled representation; $h$: encoder features; $Z$: contextual representations. ZS: zero-shot classification; LP: linear probing; I$\to$T and T$\to$I: retrieval R@1; mR: mean recall. \emph{Inherited}: pretrained components the system uses. $^b$Published CC12M CLIP of \citet{fan2023laclip} / released SLIP-repository CC12M CLIP \citep{mu2022slip} evaluated under our protocol. $^\star$Equal to the contrastive-only encoders. $^\sharp$Fine-tuned model, evaluated at $512^2$. $^c$References: published backbone linear probes. Other notes: Appendix~\ref{app:tablenotes}.}
\label{tab:understanding}
\setlength{\tabcolsep}{2.4pt}
\resizebox{\linewidth}{!}{%
\begin{tabular}{@{}lrllrrrrrrrrr@{}}\toprule
&&&&\multicolumn{3}{c}{ImageNet-1k}&\multicolumn{3}{c}{COCO retrieval}&\multicolumn{2}{c}{ADE20K}&Sugar-\\
\cmidrule(lr){5-7}\cmidrule(lr){8-10}\cmidrule(lr){11-12}
Model & Params & Inherited & Data & ZS$\uparrow$ & LP $g^{c}\uparrow$ & LP $h\uparrow$ & I$\to$T$\uparrow$ & T$\to$I$\uparrow$ & mR$\uparrow$ & $h\uparrow$ & $Z\uparrow$ & Crepe$\uparrow$\\
\midrule
\multicolumn{13}{@{}l}{\emph{CC12M-trained references}}\\
LaSLIP-B/16 \tcite{fan2023laclip} &150M &none &CC12M-L &\published{49.7} &\published{74.4}& --& --& --& --& --& --&--\\
COSMOS-B/16 \tcite{kim2025cosmos} &150M &none &CC12M-R &\published{51.4} & --& --&\published{64.2}&\published{48.9}&\published{74.6}& --& --&--\\
CLIP-B/16$^b$ \tcite{radford2021clip} &150M &none &CC12M &\published{40.2} / \oss{37.41} &\published{70.3}& --&\oss{32.90}&\oss{22.50}&\oss{48.51}& --& --&\oss{72.1}\\
SLIP-B/16 \tcite{mu2022slip} &150M &none &CC12M &\oss{41.71} & --& --&\oss{37.58}&\oss{26.78}&\oss{53.14}& --& --&\oss{72.1}\\
FLUID-L \tcite{fan2024fluid} &$\sim$5.4B &T5-XXL, SD-VAE &CC12M &--&\published{48.1}& --&--&--&--&\published{22.1}& --&--\\
DREAM-B \tcite{li2026dream} &$\sim$5.2B &T5-XXL, SD-VAE &CC12M & --&\published{69.5}$^a$& --& --& --& --& --& --&--\\
DREAM-L \tcite{li2026dream} &$\sim$5.5B &T5-XXL, SD-VAE &CC12M & --&\published{72.7}& --& --& --& --&\published{36.8}& --&--\\
\midrule
\multicolumn{13}{@{}l}{\emph{Controls (same training data and encoder architecture)}}\\
Contrastive only &143M &none &CC12M-R &49.51 &72.34 &72.51 &60.16 &41.75 &69.75 &37.41 &30.15 &79.4\\
\quad + Gram/iBOT &143M &none &CC12M-R &50.63 &73.98 &74.06 &61.90 &43.65 &70.87 &40.00 &33.13 &79.6\\
No contrastive loss &383M &none &CC12M-R &1.18 &70.22 &71.72 &0.40 &0.19 &1.51 &35.80 &23.88 &53.8\\
Frozen contrastive $\to$ generation &383M &none &CC12M-R &49.51$^\star$ &72.34$^\star$ &72.51$^\star$ &60.16$^\star$ &41.75$^\star$ &69.75$^\star$ &37.41$^\star$ &30.15$^\star$ &79.4$^\star$\\
\midrule
\corerow \textbf{\model{}} &383M &none &CC12M-R &\textbf{50.63} &\textbf{73.48} &\textbf{72.82} &\textbf{60.78} &\textbf{43.12} &\textbf{70.34} &\textbf{36.48} &\textbf{24.69} &\textbf{80.6}\\
\corerow \textbf{\model{} + Gram/iBOT} &383M &none &CC12M-R &\textbf{50.90} &\textbf{74.31} &\textbf{73.62} &\textbf{61.08} &\textbf{44.06} &\textbf{70.88} &\textbf{38.27} &\textbf{25.45} &\textbf{80.0}\\
\corerow \quad + fine-tuning$^\sharp$ &384M &none &CC12M-R &51.49 &75.86 &74.19 &62.84 &45.34 &72.02 &39.41 &23.92 &80.6\\
\bottomrule
\end{tabular}}
\end{table}

\begin{table}[t]
\centering
\caption{\textbf{Image-to-text generation.}
One sample/image, 64-step DDPM (128 denoiser evaluations, NFE).
TrueMatch uses bags of 3/5/7 images.
$^c$COCO-finetuned open\_clip \citep{ilharco2021openclip}.
$^\sharp$Fine-tuned model, $512^2$.
$^e$Second number: text denoiser $D_t$ fine-tuned on COCO captions (encoders frozen), the COCO-style protocol of the COCO-trained rows$^{a,c}$. Other notes: Appendix~\ref{app:tablenotes}.}
\label{tab:i2t}
\begingroup
\setlength{\tabcolsep}{3.5pt}
\renewcommand{\arraystretch}{0.92}
\setlength{\aboverulesep}{0.3ex}
\setlength{\belowrulesep}{0.3ex}
\resizebox{\linewidth}{!}{%
\begin{tabular}{@{}lrlr>{\hspace{9pt}}r>{\hspace{9pt}}r>{\hspace{9pt}}r>{\hspace{9pt}}crr@{}}
\toprule
&&&\multicolumn{4}{c}{COCO 5K}&TrueMatch&DetailCaps&DOCCI\\
\cmidrule(lr){4-7}
Model & Params & Inherited
& SPICE$\uparrow$ & CIDEr-D$\uparrow$
& CHAIR$_i\downarrow$ & RefCLIP-S$\uparrow$
& R@1 3/5/7$\uparrow$ & CAPTURE$\uparrow$ & CAPTURE$\uparrow$\\
\midrule
BLIP-base$^a$ \tcite{li2022blip}
&224M &ViT-B, BERT
&\oss{17.53} &\oss{94.39} &\oss{3.2} &\oss{.808}
&\oss{42.9/28.7/25.8} &\oss{23.03} &\oss{25.83}\\
NDLB \tcite{gaur2025ndlb}
&244M &CLIP-B/32, GPT-2
& --& --& --& --&\published{67.7/58.7/57.9}& --&--\\
MAGVLT-B \tcite{kim2023magvlt}
&447M &VQ-GAN
&\published{14.3}$^b$ & --& --& --& --& --&--\\
CoCa-L/14$^c$ \tcite{yu2022coca}
&638M &none (LAION-2B)
&\oss{23.45} &\oss{123.05} &\oss{3.0} &\oss{.832}
&\published{52.2/40.4/38.2} &\oss{28.81} &\oss{30.89}\\
Florence-2-base$^d$ \tcite{xiao2024florence2}
&230M &DaViT, BART
&\oss{24.66} &\oss{133.02} &\oss{2.6} &\oss{.834}
&\oss{53.1/41.9/42.2} &\oss{46.92} &\oss{48.13}\\
Cambrian-1-8B \tcite{tong2024cambrian}
&$\sim$11.0B &Llama-3, 4 enc.
& --& --& --& --&\published{60.6/53.3/53.1}& --&--\\
InstructBLIP-13B \tcite{dai2023instructblip}
&14.2B &Vicuna, EVA-ViT-g
& --& --& --& --&\published{53.7/42.7/42.1}& --&--\\
LLaVA-NeXT-34B \tcite{liu2024llavanext}
&$\sim$34.8B &Yi, CLIP-L
& --& --& --& --&\published{57.9/46.9/47.9}& --&--\\
\midrule
No contrastive loss
&383M &none
&13.86 &8.66 &10.1 &.741
&57.0/49.0/47.6 &44.43 &45.84\\
Frozen contrastive $\to$ generation
&383M &none
&10.06 &3.63 &16.5 &.736
&58.3/46.3/43.9 &43.73 &44.46\\
\midrule
\corerow \textbf{\model{}}
&383M &none
&\textbf{13.53} &\textbf{9.10} &\textbf{11.2} &\textbf{.740}
&\textbf{58.9/53.3/48.4} &\textbf{43.62} &\textbf{45.66}\\
\corerow \textbf{\model{} + Gram/iBOT}$^e$
&383M &none
&\textbf{13.50}/20.01 &\textbf{9.13}/96.06 &\textbf{11.2}/5.1 &\textbf{.740}/.822
&\textbf{60.6/51.9/47.9} &\textbf{44.15} &\textbf{45.52}\\
\corerow \quad + fine-tuning$^\sharp$
&384M &none
&13.91 &9.45 &10.4 &.743
&62.1/51.0/48.2 &44.30 &45.32\\
\bottomrule
\end{tabular}%
}
\endgroup
\end{table}

\textbf{Correctness and hallucination.} We evaluate captions on COCO Karpathy-5K, with five reference captions per image. SPICE \citep{anderson2016spice} measures semantic agreement with the references. CHAIR$_i$ \citep{rohrbach2018chair} measures the percentage of object mentions that are hallucinated. RefCLIPScore \citep{hessel2021clipscore} measures agreement with both the image and the references using CLIP embeddings. \polish{At equal data and architecture, \model{} substantially outperforms the frozen contrastive $\to$ generation control, raising SPICE from 10.06 to 13.53 and cutting hallucinated object mentions from 16.5\% to 11.2\%.} Its zero-shot SPICE score trails several captioners trained on larger or COCO-specific caption datasets. The fine-tuned model reaches SPICE 13.91 and reduces hallucinated object mentions to 10.4\%. Fine-tuning only the text denoiser on COCO captions, the protocol of the COCO-trained baselines, raises SPICE to 20.01 and CIDEr-D to 96.06 and lowers hallucinated object mentions to 5.1\% (second numbers in Table~\ref{tab:i2t}).

\textbf{Image-specific detail.} Reference agreement can reward generic captions, so TrueMatch \citep{gaur2025ndlb} asks a frozen CLIP ViT-B/32 listener to pick each caption's source image from a bag of 3, 5, or 7 visually similar images. \polish{Among our models and controls, \model{} reaches R@1 of 58.9\%, 53.3\%, and 48.4\%, ahead of both frozen-contrastive and no-contrastive-loss. Against external systems, it exceeds} LLaVA-NeXT-34B at every bag size and approaches Cambrian-1-8B (which leads at sizes 3 and 7 and ties at size 5), while trailing NDLB, which is explicitly trained for self-retrieval.

\textbf{Detailed coverage.} DetailCaps-4870 (CAPTURE against three detailed references; \citealp{dong2024capture}), 
and DOCCI (CAPTURE against one long human description; \citealp{onoe2024docci}) ask whether a caption covers fine-grained content. \polish{Compared with the frozen-contrastive control, \model{} leads on DOCCI (45.66 vs.\ 44.46) and CompreCap (29.4 vs.\ 26.0 unified score under a Qwen2.5-7B-Instruct judge; Table~\ref{tab:comprecap}). Against external models, \model{} surpasses} BLIP-base, CoCa-L/14, and caption-prompted Florence-2 on CompreCap (22.2, 27.3, and 27.7) as well as BLIP and CoCa on DetailCaps and DOCCI, while trailing detailed-caption-prompted Florence-2 (trained on 5.4B generated annotations) and 7B--17B multimodal LLMs.

\begin{table}[t]\centering
\caption{\textbf{Text-to-image generation and reconstruction.} Our models use $256^2$ unless marked. For our non-distilled models, GenEval/DPG use 128 steps (257 NFE), and FID uses 64 steps (128 NFE); distilled models use the budgets shown in their row labels. rFID denotes reconstruction FID. $^\sharp$Fine-tuned model, $512^2$. Notes $^{a\text{--}c}$: Appendix~\ref{app:tablenotes}.}
\label{tab:t2i}
\setlength{\tabcolsep}{2.4pt}
\resizebox{\linewidth}{!}{%
\begin{tabular}{@{}lrllrrrrrr@{}}\toprule
&&&&\multicolumn{2}{c}{Prompt following}&\multicolumn{2}{c}{FID}&\multicolumn{2}{c}{Reconstruction}\\
\cmidrule(lr){5-6}\cmidrule(lr){7-8}\cmidrule(lr){9-10}
Model & Params & Inherited & Data & GenEval$\uparrow$ & DPG$\uparrow$ & COCO-30K$\downarrow$ & CC12M-50K$\downarrow$ & PSNR$\uparrow$ & rFID$\downarrow$\\
\midrule
\multicolumn{10}{@{}l}{\emph{External systems}}\\
SD1.5 \tcite{rombach2022ldm} &1.1B &CLIP-L, VAE &LAION &\published{.43} &\published{63.18} & --& --& --&--\\
SDXL \tcite{podell2023sdxl} &3.5B &CLIP-L/bigG, VAE &internal &\published{.55} &\published{74.65} & --& --& --&--\\
MAGVLT-B \tcite{kim2023magvlt} &447M &VQ-GAN &CC3M/12M+ & --& --&\published{12.08}$^a$ & --& --&--\\
MiniT2I-B/32 \tcite{minit2i2026} &601M &FLAN-T5-Large &CC12M-R &\published{.505} &\published{78.3} &\published{13.69} / \oss{13.42} &\oss{9.51} &-- &--\\
MiniT2I-B/16 \tcite{minit2i2026} &599M &FLAN-T5-Large &CC12M-R+120K &\published{.873} &\published{84.2} &\published{10.51}$^b$ / \oss{15.54} &\oss{11.35} &-- &--\\
FLUID-L \tcite{fan2024fluid} &$\sim$5.4B &T5-XXL, SD-VAE &CC12M & --& --&\published{9.62} &\published{4.53} & --&--\\
DREAM-L \tcite{li2026dream} &$\sim$5.5B &T5-XXL, SD-VAE &CC12M & --& --&\published{10.0} &\published{4.34} & --&--\\
SD-VAE \tcite{rombach2022ldm} &84M &-- &autoencoder &-- &-- &-- &-- &\oss{25.04} &\oss{3.05}\\
SDXL-VAE \tcite{podell2023sdxl} &84M &-- &autoencoder &-- &-- &-- &-- &\oss{26.10} &\oss{2.95}\\
\midrule
\multicolumn{10}{@{}l}{\emph{Controls}}\\
No contrastive loss &383M &none &CC12M-R &.193 &45.31 &23.20 &9.73 &26.91 &4.76\\
Frozen contrastive $\to$ generation &383M &none &CC12M-R &.400 &70.12 &12.44 &7.99 &19.26 &11.58\\
\midrule
\corerow \textbf{\model{}} &383M &none &CC12M-R &\textbf{.461} &\textbf{74.43} &\textbf{10.10} &\textbf{4.33} &\textbf{24.09} &\textbf{6.43}\\
\corerow \textbf{\model{} + Gram/iBOT} &383M &none &CC12M-R &\textbf{.448} &\textbf{73.49} &\textbf{10.20} &\textbf{4.37} &\textbf{24.03} &\textbf{6.24}\\
\corerow \quad + fine-tuning$^\sharp$ &384M &none &+120K &\textbf{\nrev{.806}} &\textbf{\nrev{79.29}} &\nrev{31.81}$^c$ &\nrev{15.27} &\nrev{25.35} &\nrev{6.29}\\
\corerow \quad\quad + distilled, 1 jump (2 NFE) &384M &none &+120K &\nrev{.782} &\nrev{75.96} &\nrev{27.12} &\nrev{17.27} &-- &--\\
\corerow \quad\quad + distilled, 8 jumps (9 NFE) &384M &none &+120K &\nrev{.797} &\nrev{78.14} &\nrev{36.05} &\nrev{18.93} &-- &--\\
\bottomrule
\end{tabular}}

\vspace{6pt}
\makeatletter\def\@captype{figure}\makeatother
{\setlength{\tabcolsep}{1pt}%
\begin{tabular}{@{}cccccccc@{}}
\includegraphics[width=.120\linewidth]{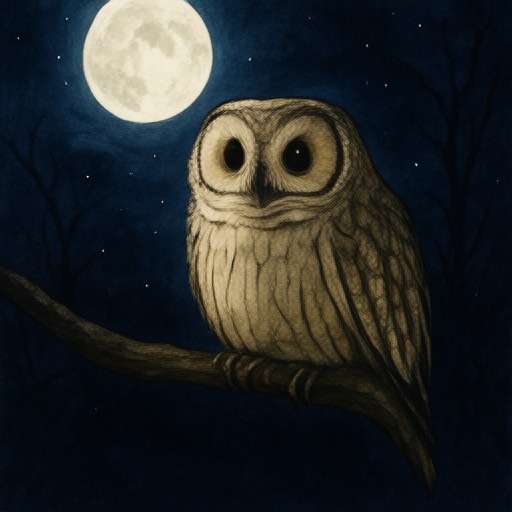} & \includegraphics[width=.120\linewidth]{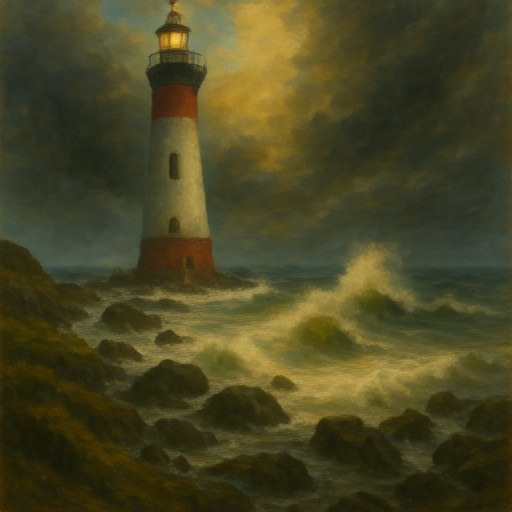} & \includegraphics[width=.120\linewidth]{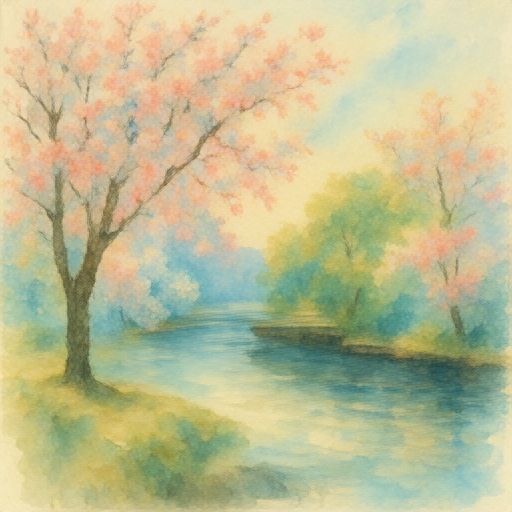} & \includegraphics[width=.120\linewidth]{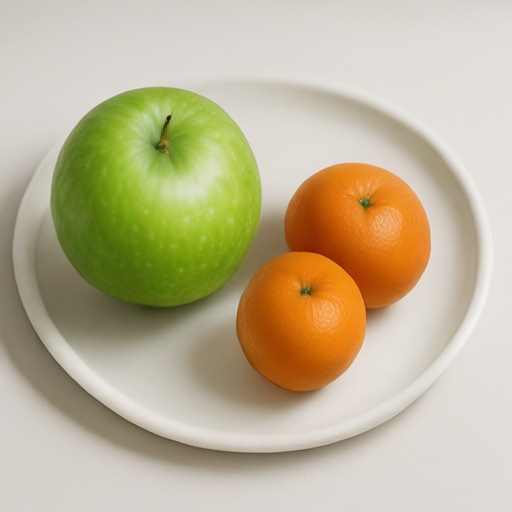} & \includegraphics[width=.120\linewidth]{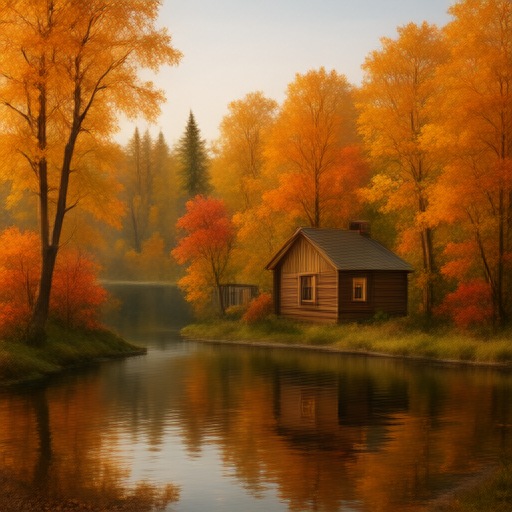} & \includegraphics[width=.120\linewidth]{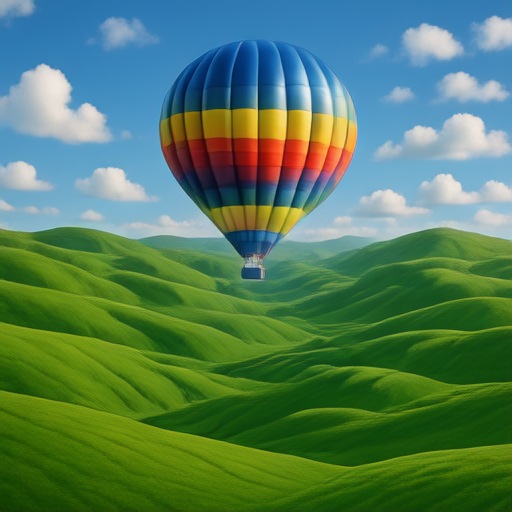} & \includegraphics[width=.120\linewidth]{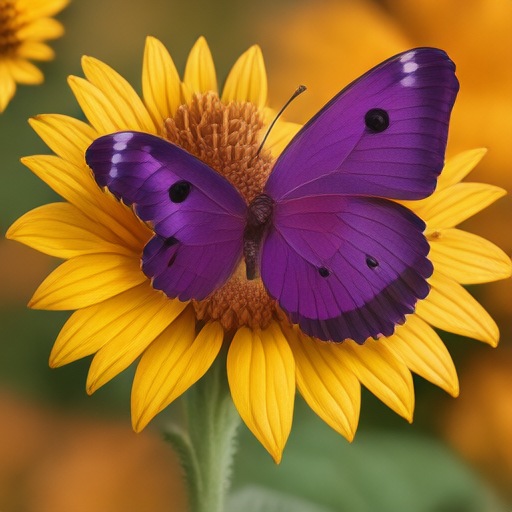} & \includegraphics[width=.120\linewidth]{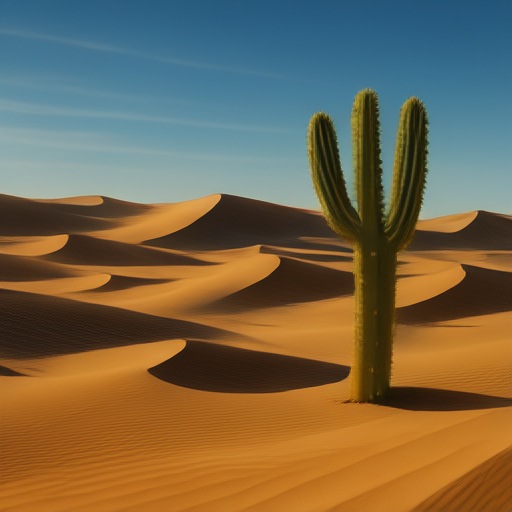}\\
\parbox[t]{.120\linewidth}{\centering\tiny An owl perched on a branch at night under a full moon} & \parbox[t]{.120\linewidth}{\centering\tiny An oil painting of a lighthouse on a rocky coast during a storm} & \parbox[t]{.120\linewidth}{\centering\tiny A watercolor painting of cherry blossoms along a quiet river} & \parbox[t]{.120\linewidth}{\centering\tiny A green apple and two oranges on a white plate} & \parbox[t]{.120\linewidth}{\centering\tiny A small wooden cabin beside a lake surrounded by autumn trees, reflected in the water} & \parbox[t]{.120\linewidth}{\centering\tiny A hot air balloon floating over rolling green hills} & \parbox[t]{.120\linewidth}{\centering\tiny A purple butterfly resting on a sunflower} & \parbox[t]{.120\linewidth}{\centering\tiny A desert landscape with sand dunes and a single cactus}\\[2pt]
\end{tabular}
\caption{\textbf{Text-to-image samples from \model{}} (fine-tuned model, $512^2$; sampler selected on held-out prompts). For each prompt, one of four seeds is chosen by the model's \emph{own} image--text score, not an external judge; all seeds of all 24 prompts are in Figure~\ref{fig:allseeds}, and samples for longer prompts in Figure~\ref{fig:dpgsamples}.}
\label{fig:t2ibest}
}
\end{table}

\subsection{Text-to-image generation}
\label{sec:t2i}
Table~\ref{tab:t2i} evaluates prompt following, image-distribution quality, and image reconstruction.

\textbf{Prompt following.} GenEval \citep{ghosh2023geneval} checks single objects, object pairs, counting, colors, relative position, and color attribution over 553 prompts with an object detector; DPG-Bench \citep{hu2024ella} evaluates entities, attributes, and relations through VQA questions over 1065 dense prompts. For both benchmarks, our non-distilled models generate four samples per prompt using a fixed sampler (128 steps, CFG 3), with prompts encoded by our own text encoder trained only on the training captions. Without fine-tuning, \model{} reaches .461 on GenEval and 74.43 on DPG\polish{, well ahead of the frozen-contrastive (.400 and 70.12) and no-contrastive-loss (.193 and 45.31) controls. It outperforms} SD1.5 on both and matches SDXL on DPG. \model{} trails MiniT2I-B/32 (.505 and 78.3), which uses a pretrained FLAN-T5-Large text encoder. Fine-tuning raises \model{} + Gram/iBOT from .448 to \nrev{.806} on GenEval and from 73.49 to \nrev{79.29} on DPG at $512^2$, without sacrificing the first-stage model's understanding or captioning, since the second stage updates only the image denoiser.

\textbf{Distribution quality.} Single-sample FID compares one generation per prompt with real images: zero-shot on COCO-30K (one caption per validation image) and in-distribution on 50K CC12M training images with their original alt-text. The FID sampler is selected once on disjoint COCO development prompts (64 steps, 128 NFE). \polish{Relative to the frozen-contrastive control (12.44 and 7.99), \model{} lowers FID to 10.10 on COCO-30K and 4.33 on CC12M-50K. Among external generators, this sits} close to the reported 10.0 and 4.34 of DREAM-L (which uses frozen T5-XXL and SD-VAE) and lower than MiniT2I-B/32 under the same protocol (13.42 and 9.51).

\textbf{Recoverability.} We encode 5K ImageNet validation images, add noise at the sampling schedule's lowest noise level, and reconstruct each image with one denoiser call conditioned on its own contextual representation. \polish{\model{} reaches 24.09\,dB PSNR, nearly 5\,dB above the frozen-contrastive control (19.26\,dB), whereas the no-contrastive-loss control reconstructs more accurately (26.91\,dB) but follows prompts poorly (GenEval .193), a contrast analyzed in Section~\ref{sec:analysis}. Against external autoencoders, \model{} lies within 1\,dB of SD-VAE (25.04\,dB).}

\subsection{Few-step generation}
\label{sec:distillation}
Because generation runs in a continuous contextual representation space, we can distill both image and text generation into a few steps while keeping the encoders fixed. We distill the \model{} + Gram/iBOT denoisers into shortcut students that jump between noise levels \citep{frans2025shortcut,boffi2025flowmap}, anchoring each student's final pixel or token predictions to the teacher's to preserve output quality.

\polish{\emph{Images.}} With eight jumps (9 NFE), the student of the fine-tuned image denoiser reaches GenEval \nrev{.797, against .806} for its 257-NFE teacher\nrev{, and a single jump (2 NFE) already reaches .782} (Table~\ref{tab:t2i}). \nrev{On the pretrained model, distilling from a stronger guidance trajectory (CFG~7) lets the 9-NFE image student exceed its 257-NFE teacher on GenEval (.485 vs.\ .448; Table~\ref{tab:distillscale}), while cutting single-image latency on one H100 from 4.75\,s to 173\,ms (38\,ms at one jump; Table~\ref{tab:latency}).}

\polish{\emph{Text.}} \nrev{Symmetrically, the eight-jump caption student (9 NFE) matches its 128-NFE teacher on caption quality and detailed coverage (SPICE 13.66 vs.\ 13.50, DetailCaps 44.86 vs.\ 44.15, DOCCI 46.54 vs.\ 45.52; 12.05 SPICE in one jump; Table~\ref{tab:distillcap}), generating a caption in 79\,ms at batch~1 (17\,ms at one jump)\polish{, faster} than autoregressive BLIP-base (107\,ms), Florence-2-base (136\,ms), and CoCa-L/14 (352\,ms).}
\section{How does joint learning shape representations?}
\label{sec:analysis}

Section~\ref{sec:results} shows that one contextual representation can support understanding, conditioning, and generation. We now examine how joint learning shapes this representation, distinguishing \emph{retention} \polish{(what information remains recoverable)} from \emph{organization} \polish{(how it is arranged and accessed)}. The controls of Section~\ref{sec:results} already separate the two: without the alignment loss, the representation retains the image (26.9 dB) but \polish{follows prompts poorly} (GenEval .193). Shaped by alignment alone and then frozen, it is organized but retains less (19.3 dB, GenEval .400). \polish{Below we examine four aspects of this interplay: semantic shaping of the generation space, dense retention across readouts, local neighborhood geometry, and capabilities unlocked by generating inside $Z$.}

\textbf{\polish{Shaping the generation space with understanding improves compositionality.}}
\label{sec:coexist}
We compare two matched continuations of the same checkpoint (Table~\ref{tab:endpointmain}). In the \emph{shared} arm, image generation uses the contextual representation $Z_v$ used for understanding, whereas in the \emph{separate} arm it uses a second projection of the same visual backbone, initialized as a copy of the original, while understanding continues to use $Z_v$ so that understanding neither shapes nor reads the second space. Although the separate representation reconstructs images better (25.03 vs.\ 23.38 dB), the shared representation achieves better compositional generation (GenEval .428 vs.\ .362, DPG 71.30 vs.\ 65.91), with similar global understanding. Better reconstruction therefore does not necessarily yield better conditional generation: \polish{a space must be semantically organized to compose text and visual concepts reliably.}

Giving the separate projection its own alignment loss \nrev{refines} this: it then generates nearly as well as the shared representation and never better (GenEval .423 and .411 on two training seeds, against .427 and .428), and trained from scratch it \nrev{is largely linearly predictable from} $Z_v$ ($R^2$ .89 / .95, Appendix~\ref{app:endpoint}). \nrev{The evidence thus supports shaping the generation space by understanding together with generation, not sharing as such}\polish{. Because a second aligned space brings no gain, one representation suffices.}

\begin{table}[t]
\centering
\caption{\textbf{Shared versus separate generation space.}
Both arms continue the same 24K-step checkpoint for 16K steps, generating in either the understanding representation $Z_v$ or a separate projection initialized as its copy. ADE20K uses the reduced-resolution diagnostic protocol.}
\label{tab:endpointmain}
\setlength{\tabcolsep}{4.5pt}\small
\begin{tabular}{@{}lrrrrrrrr@{}}
\toprule
Generated in & ZS & mR & SPICE & GenEval & DPG & PSNR & ADE lin. & ADE MLP\\
\midrule
\corerow Shared (\model{}) & 44.65 & 69.15 & 12.01 & \textbf{.428} & \textbf{71.30} &
23.38 & 15.4 & 26.0\\
Separate & 44.66 & 69.42 & 12.13 & .362 & 65.91 &
\textbf{25.03} & \textbf{21.9} & 26.6\\
\bottomrule
\end{tabular}

\vspace{5pt}
\makeatletter\def\@captype{figure}\makeatother
\normalsize \begin{tikzpicture}
\definecolor{rlCtrFill}{HTML}{A8D1E1}\definecolor{rlNavy}{HTML}{0B2D4A}\definecolor{rlSec}{HTML}{5B6F85}\definecolor{rlGrid}{HTML}{E6EBF1}\definecolor{rlAxis}{HTML}{A3AFBD}%
\begin{axis}[width=.78\linewidth,height=3.2cm,ybar=1.5pt,bar width=22pt,ymin=0,ymax=50,ytick={0,10,20,30,40,50},
 ylabel={ADE20K mIoU},symbolic x coords={Linear probe,MLP probe,Attentive probe},
 xtick=data,enlarge x limits=.19,axis lines*=left,
 x axis line style={rlAxis,line width=.6pt},y axis line style={draw=none},
 xtick style={draw=none},ytick style={draw=none},major tick length=0pt,
 ymajorgrids,grid style={rlGrid,line width=.5pt},
 yticklabel={\pgfmathprintnumber[assume math mode=true]{\tick}},
 xticklabel style={font=\sffamily\fontsize{7}{8}\selectfont,text=rlNavy},
 yticklabel style={font=\sffamily\fontsize{6.5}{7.5}\selectfont,text=rlSec},
 ylabel style={font=\sffamily\fontsize{6.5}{7.5}\selectfont,text=rlSec},
 legend style={at={(1.02,.5)},anchor=west,legend columns=1,draw=none,fill=none,legend cell align=left,
  font=\sffamily\fontsize{7}{8}\selectfont,text=rlNavy,row sep=1.5pt},
 legend image code/.code={\draw[#1] (0cm,-0.05cm) rectangle (0.17cm,0.12cm);},
 nodes near coords={\pgfmathprintnumber[fixed,zerofill,precision=1,assume math mode=true]{\pgfplotspointmeta}},
 nodes near coords style={font=\sffamily\fontsize{6.5}{7.5}\selectfont,text=rlNavy,inner sep=1.5pt}]
\addplot[fill=rlCtrFill,draw=ctr,line width=.4pt] coordinates {(Linear probe,30.15)(MLP probe,35.61)(Attentive probe,41.20)};
\addplot[fill=joint,draw=joint,line width=.4pt] coordinates {(Linear probe,24.69)(MLP probe,34.12)(Attentive probe,42.38)};
\addplot[fill=jointlt,draw=joint,line width=.4pt] coordinates {(Linear probe,25.45)(MLP probe,35.61)(Attentive probe,43.54)};
\legend{Contrastive only,\model{},\model{} + Gram/iBOT}
\end{axis}\end{tikzpicture}
\caption{\textbf{Fixed representations, different readouts.}
ADE20K mIoU on the frozen $Z$ under three probes of growing capacity (MLP and attentive: mean of three probe seeds). The gap to the contrastive-only control closes and then reverses: the dense information is retained, and what changes is how it is accessed.}
\label{fig:accessibility}
\end{table}

\textbf{A small nonlinear readout recovers most of the dense-prediction gap, and an attentive probe reverses it.}
\label{sec:retention}
The clearest tension appears in dense visual prediction. On ADE20K (Figure~\ref{fig:accessibility}), the jointly learned $Z$ is 5.5 mIoU below the contrastive-only control with a linear probe (24.7 vs.\ 30.2) and 1.5 below with an MLP (34.1 vs.\ 35.6), whereas with an attentive probe (two transformer blocks) \model{} leads at 42.4 vs.\ 41.2 (three probe seeds, Table~\ref{tab:ladderfull}).  The representations remain frozen throughout, indicating that much of the dense information remains recoverable through a simple nonlinear readout. The shared-versus-separate comparison shows the same pattern: a large linear-probe gap nearly disappears with an MLP (Table~\ref{tab:endpointmain}). These results locate much of the apparent tension in how information is accessed rather than whether it is retained\nrev{: a linear probe measures what is linearly arranged, whereas a stronger readout measures what is present.}

\nrev{Two further readouts reinforce this conclusion. On monocular depth estimation on NYU Depth v2 from frozen $Z$ (Table~\ref{tab:depth}), \model{} outperforms the contrastive-only control under every probe\polish{: linear ($\delta_1$ 58.5\% vs.\ 57.0\%), MLP (61.0\% vs.\ 57.8\%), and attentive (74.2\% vs.\ 72.7\%). Thus, the ADE20K linear-probe deficit is not a general loss of spatial geometry.} Likewise, when a frozen Qwen2.5-1.5B language model reads frozen visual features through a 2-layer MLP projector (Table~\ref{tab:llmreadout}, Appendix~\ref{app:llmreadout}), \polish{\model{} leads the same-data contrastive control on both $Z$ (64.6 vs.\ 63.0 VQAv2) and $h$ (65.6 vs.\ 64.9), surpassing DINOv2-B (trained on 142M images; 64.6) on $h$ and keeping its lead with a frozen 7B language model (70.0 vs.\ 68.7; full LoRA and CIDEr-D results in Appendix~\ref{app:llmreadout}).}}

\textbf{Joint learning reorganizes visual and textual neighborhoods differently.}
\label{sec:organization}
\polish{Contrastive-only visual patches strongly broadcast global image identity, whereas generative supervision pushes each patch to encode local appearance and geometry shared across images, consistent with the lower linear semantic readout alongside stronger depth and attentive-probe accuracy.} We measure neighborhood \emph{enrichment}: the fraction of a token's ten nearest neighbors sharing a label, divided by the fraction expected under random sampling. From the contrastive-only control to \model{}, same-image enrichment among visual patches falls from $83.3\times$ to $6.5\times$ \polish{(mean same-image patch cosine from .274 to .025; Appendix~\ref{app:geometry})}, and same-class enrichment among patches from other images falls from $9.0\times$ to $2.0\times$. Text instead preserves local lexical structure: same-word and same-family enrichment across captions remain nearly unchanged ($51.4\times$ to $50.8\times$ and $5.88\times$ to $5.80\times$, respectively; same-family neighbors exclude the same word). Connecting each text token to its 12 nearest neighbors yields an undirected graph whose largest connected component contains, on average, 97\% of sampled tokens for \model{}, compared with 68\% for the contrastive-only control. Contextual meanings also remain distinguishable: leave-one-out nearest-neighbor classification within each word achieves 98.8\% balanced sense accuracy, averaged over six polysemous words. Joint learning therefore produces different organizational changes across modalities: less source-centric visual neighborhoods, and preserved local lexical structure within a more connected text representation.

\textbf{Reading\nrev{, ranking,} and steering a generation in $Z$.}
\label{sec:trajectory}
Because generation runs in $Z$, probes \polish{and similarity scores} trained on encoder outputs \polish{apply directly to sampling states without adaptation.}

\polish{\emph{Probing sampling states.}} For \model{} + Gram/iBOT, the zero-shot class read after one denoiser call is correct for 54\% of the 424 prompts whose noun maps to an ImageNet class\nrev{. Across 36 tracked trajectories \polish{with the fine-tuned denoiser}, every state from the first step onward is nearest to its prompted class (final-state cosine 0.81 vs.\ 0.82 for real images), and a frozen language-model projector trained on real-image $Z$ captions the final generated state more accurately than the decoded and re-encoded image (CIDEr 93 vs.\ 82, Table~\ref{tab:uses}).}

\polish{\emph{Self-ranking without an external scorer.}} \nrev{Using the model's own pooled image--text cosine to select among four samples per prompt\polish{, with no external reranker,} raises GenEval from .436 to .501 on pretrained \model{} + Gram/iBOT \polish{and from .734 to .764 on the all-parameter fine-tune, matching an external CLIP reranker (.765; full image and caption self-ranking results in Appendix~\ref{app:ownselect} and Table~\ref{tab:ownselect5k}).}}

\begin{figure}[t]\centering
\includegraphics[width=\linewidth]{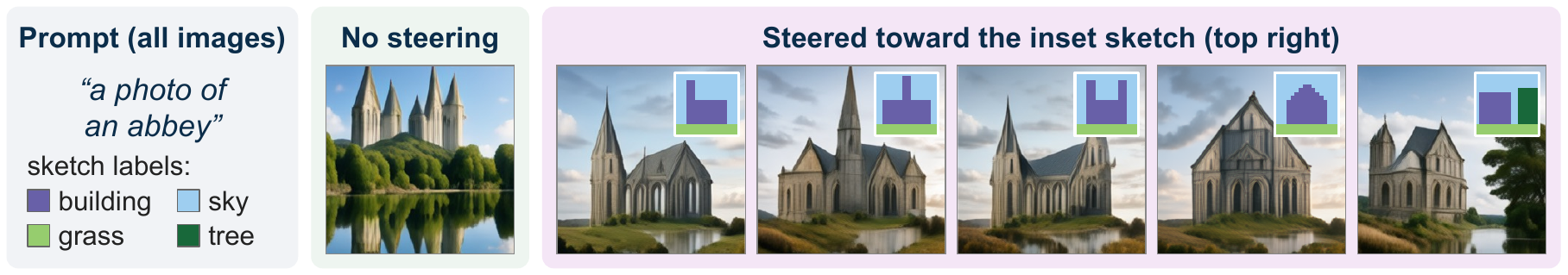}
\caption{\textbf{A probe trained for understanding steers generation.} \emph{Left:} no steering. \emph{Right:} the gradient of a frozen ADE20K linear probe, trained only on real images, pulls the early sampling steps toward the layout sketch in the inset. Nothing is trained for this: the probe applies because images are generated in the representation it reads.}
\label{fig:steerstrip}
\end{figure}

\polish{\emph{Training-free layout steering.}} \nrev{The} gradient of a frozen ADE20K linear probe, applied in the early steps, steers the image toward a layout sketch without training: \model{}'s images follow their sketch on 83\% of the pixels, against 50\% unsteered (Figure~\ref{fig:steerstrip}\nrev{; 82\% vs.\ 42\% with the fine-tuned denoiser). Guiding toward predicted layouts of reference images raises pixel agreement from 30\% to 55\%, lowers paired COCO-5K FID from 21.6 to 20.4, and reaches 86.7\% sketch agreement when combined with intermediate-state ranking at step 24 of 64 (Appendix~\ref{app:steer}).}

\section{Conclusion}
\model{} shows that \polish{one contextual representation per modality can be shaped jointly by understanding and generation, improving generation in both directions over a representation shaped by understanding alone and then frozen, and strengthening understanding under stronger readouts.} For each modality, this representation is read for understanding, conditions generation of the other modality, and is the space in which that modality is generated. Trained from scratch on CC12M, with 383M parameters and no pretrained language model, text encoder, or VAE in the inference model, \model{} matches same-data contrastive controls on global understanding while generating competitive images, though prompt following and caption fidelity remain below systems that inherit stronger text encoders or more caption data. More importantly, the controls \polish{indicate} why one representation suffices: understanding organizes \polish{the representation semantically, while generation makes it retain more information.}

\textbf{Limitations and outlook.} Our goal is to isolate this representation-level question rather than to scale the system. Our text encoder sees only CC12M captions, which makes counting and spatial prompts hard.
Larger scale and text-only data are compatible with the construction, since language is learned and generated in a representation of its own. \polish{We hope this controlled study informs larger-scale pretraining and future work on how one representation can retain and organize the information needed to both understand and create.}

\bibliographystyle{plainnat}
\bibliography{references}

\clearpage
% \appendix
\beginappendix   

\noindent\textbf{Appendix overview.}
\begin{itemize}\setlength{\itemsep}{1pt}\setlength{\parskip}{0pt}\setlength{\topsep}{2pt}
  \item[\textbf{A.}] Architecture, training recipe, and compute cost (Appendix~\ref{app:implementation}).
  \item[\textbf{B.}] Evaluation protocols for every benchmark (Appendix~\ref{app:protocols}).
  \item[\textbf{C.}] External comparisons: parameter, data, and protocol accounting (Appendix~\ref{app:external}).
  \item[\textbf{D.}] Controlled interventions: conditioning routes, objective balance, shared vs.\ separate space, and fine-tuning (Appendix~\ref{app:ablations}).
  \item[\textbf{E.}] Extended results, qualitative samples, own-score selection, frozen-LLM readout, and steering (Appendix~\ref{app:extended}).
  \item[\textbf{F.}] Few-step distillation and fixed-hardware latency (Appendix~\ref{app:distillation}).
  \item[\textbf{G.}] Supplementary study: iterative contextual refinement in text sampling (Appendix~\ref{app:iteration}).
  \item[\textbf{H.}] Representation geometry, readout ladders, and polysemy measurements (Appendix~\ref{app:geometry}).
\end{itemize}

\section{Architecture, training, and resource accounting}
\label{app:implementation}
\paragraph{Encoders and the contextual representation.}
The visual tower is a 12-layer ViT (width 768, 12 heads) on $16\times16$ patches of images in $[0,1]$, with no mean/std normalization and learned absolute positions (no class token). The text tower has 12 layers, width 512, and eight heads over a learned embedding table; token identities follow the T5 SentencePiece tokenizer \citep{raffel2020t5}, whose 32,100 entries are padded to a 32,128-way embedding and output layer. Blocks are pre-norm with GELU MLPs (ratio 4) and QK-RMSNorm. Each tower projects to width 256 and applies non-affine LayerNorm, giving $Z_m$. Eight learned queries attention-pool $Z_m$; their outputs are averaged and layer-normalized to the 512-d global readout $g_m$. Alignment uses the L2-normalized $g_m$; conditioning uses the layer-normalized output. Padding is excluded from encoder attention, pooling, cross-attention keys, and the token cross-entropy; the text denoiser's self-attention is unmasked, so pad slots are part of the language trajectory.

\paragraph{Denoisers and heads.}
The visual denoiser has width 768, 12 blocks, and 12 heads; the language denoiser has width 512, six blocks, and eight heads. Both use adaLN-zero modulation from an MLP of a sinusoidal embedding of $\gamma/4$ plus the projected global condition, and cross-attention to the local condition. The visual observable head $B_v$ reshapes the final $16\times16$ grid of trunk features and renders RGB with a small convolutional network: a $3{\times}3$ convolution to 128 channels, GroupNorm, SiLU, and a residual block, then four stages of nearest $\times2$ upsampling with convolution and residual blocks ($128{\to}64{\to}32{\to}32{\to}32$ channels), a final convolution to three channels, and a sigmoid (1.42M parameters). The text observable head is an untied linear layer to vocabulary logits. The representation heads $F_m$ are sibling linear layers to width 256. The tensor in which image generation runs is the same $Z_v$ used for local representation readout.

\paragraph{Language length and termination.}
A caption occupies a fixed canvas of 256 token slots: the tokenizer appends an end-of-sequence token and pads with zeros. All slots are noised and denoised in parallel; the cross-entropy covers valid tokens and the end token, and the representation regression covers all slots. At inference, each slot is read by argmax and everything after the first end token is discarded; captions without an end token are flagged. The text ladder terminates at $\gamma\approx4.3$, and the token logits are taken from the conditional branch of the last denoiser call; guidance acts on the clean-representation estimate only.

\paragraph{Complete objective.}
The implemented base objective is Equations~\ref{eq:fullobjective}--\ref{eq:genlosses}, i.e.\ $\mathcal L_{\rm align}+\mathcal L_{\rm img}+0.5\,\mathcal L_{\rm text}$ with representation-regression weights 0.25 (image) and $0.5\times0.25=0.125$ (text), plus the text-schedule fit $0.1\,\mathcal L_{\rm sched}$ on detached losses, which updates only the schedule parameters. $\mathcal L_{{\rm rep},m}$ is a mean over batch, positions, and channels.
SigLIP uses a learnable temperature initialized at $1/0.07$ (capped at 100) and bias initialized at $-10$, with negatives gathered across devices. Clean-representation targets are detached; the noisy input and source condition carry gradients into the encoders. Schedule fitting uses detached losses. The LPIPS term uses torchvision VGG16 features at the five LPIPS taps with channel unit-normalization and a spatial mean, averaged over taps and without LPIPS's learned channel weights; its weight 0.1 therefore corresponds to a rescaled, uncalibrated LPIPS. The P-DINO term averages the patchwise cosine distance between last-layer DINOv2-B features of predicted and target images (an $18\times18$ grid at 256 pixels). The perceptual weights (0.1, 0.01) and gate ($t\ge0.3$, i.e.\ $\gamma<\edit{\gamma_{\rm p}=}1.695$\edit{, the threshold $\gamma_{\rm p}$ of Equation~\ref{eq:genlosses}}) follow PixelGen \citep{ma2026pixelgen}. Both perceptual terms are evaluated on a subsample of 1024 images per training step (512 during fine-tuning).

The visual noise proposal is a truncated Gaussian $\gamma\sim\mathcal N(1.6,1.6^2)$ on $[-12,12]$, i.e.\ a logit-normal time distribution $\operatorname{logit}(t)\sim\mathcal N(-0.8,0.8^2)$ as in SD3, PixelGen, and JiT \citep{esser2024sd3,ma2026pixelgen,li2025jit}; it is fixed. The language proposal is LangFlow's learned truncated-Gumbel loss-curve schedule \citep{chen2026langflow}, initialized at location 0 and scale 2; $\gamma$ is drawn by stratified inverse-CDF sampling. Self-conditioning is disabled in the base models; its projection stays exactly zero. The frozen-encoder self-conditioning continuations are specified in Appendix~\ref{app:iteration}.

\paragraph{Optimization and data.}
Training uses the CC12M images \citep{changpinyo2021cc12m}, obtainable from the dataset's official URLs. Each image carries its alt-text and two synthetic captions from a public release \citep{captionemporium2024cc12m}: a long description generated by LLaVA-NeXT (Llama-3-8B) and a short version of at most 30 words condensed by Llama-3-8B. Each training example draws one of the three captions with probabilities 0.45/0.35/0.20 (long/short/alt-text), renormalized over the captions present; MiniT2I trains on the long captions only, so our caption mix differs from its setting. AdamW ($\beta_1=0.9$, $\beta_2=0.95$), learning rate $1.4\times10^{-3}$ with 4K linear warm-up and cosine decay to zero, gradient clipping at 1.0, EMA decay 0.999, global batch 8192, 125K training steps. The effective weight decay on encoder and denoiser matrices is 0.2; biases, normalization parameters, the patch embedding, the token embedding, positional tables, null codes, and the SigLIP scale/bias are excluded. Random resized crops use area $[0.5,1]$; horizontal flips are gated by caption compatibility.

\paragraph{Structural intervention.}
\model{} + Gram/iBOT resumes the joint run at training step 24K (optimizer, schedule, and EMA included) and continues to 125K with two added terms on backbone patch features: Gram anchoring toward the model's own 24K weights (weight 2) and an iBOT-style masked self-distillation loss with an EMA teacher (weight 1; 8192 prototypes, temperatures 0.1/0.04, block masking of 10--50\% of patches, a separate AdamW at $10^{-3}$) \citep{dinov3,ibot}. The iBOT head (8.4M parameters) and teacher are dropped at inference. Both terms act at full weight from training step 24,001, without warm-up. 

\paragraph{Fine-tuning.}
Every fine-tune starts from the \model{} + Gram/iBOT EMA weights with a fresh optimizer, with Gram and iBOT off. The synthetic data is MiniT2I's 120K mixture \citep{minit2i2026}: BLIP3o-60k \citep{chen2025blip3o}, DALL$\cdot$E~3 images \citep{opendatasets2023dalle3}, and ShareGPT-4o-Image images \citep{chen2025sharegpt4oimage}, each with its captions.

\emph{Final fine-tune at $512^2$} (the $^\sharp$ rows of the main tables). The positional tables are interpolated bicubically to a $32\times32$ grid (1024 image tokens; 384M parameters), and the visual noise proposal is shifted by $\ln 4$ in log noise-to-signal ratio for the fourfold token count, as in SD3 \citep{esser2024sd3}. A first stage trains all parameters for 10K steps on CC12M-R images whose short side is at least 512 pixels (batch 5120, learning rate $1.1\times10^{-4}$, 250 warm-up steps). The second stage then trains only the image denoiser and its null codes on the 120K mixture for \nrev{76K steps; the last 52K use no horizontal flips or random crops and follow a cosine schedule set for 64K steps, and the EMA weights at the end of these 52K steps are evaluated} (batch 5120, learning rate $1.1\times10^{-4}$, 100 warm-up steps), with both encoders, the text denoiser, and the text null codes frozen and the alignment loss switched off. Because nothing that understanding or captioning reads is updated, both are measured once on the first-stage model at its native $512^2$ input.

\emph{Controlled fine-tunes at $256^2$} (Appendix~\ref{app:finetune}). 5000 training steps on the 120K mixture at batch 8192, i.e.\ MiniT2I-B/16's sample budget (41M pair presentations, about 340 passes over the mixture), learning rate $1.4\times10^{-4}$ with 20 warm-up steps and cosine decay, EMA decay 0.999 (the evaluated EMA retains 0.7\% of the initial weights). One run trains all parameters; the other freezes both encoders, switches the alignment loss off, and trains the two denoisers and their null codes.

\subsection{Parameters and external resources}
\label{app:resources}
\begin{table}[ht]\centering\small
\caption{\textbf{Parameter accounting for the full base model.} Counts in millions. The renderer is part of the visual denoiser. Training-only components (perceptual networks, iBOT head and teacher, Gram anchor) and evaluators are excluded.}
\label{tab:params}
\begin{tabular}{@{}lr@{}}\toprule
Component & Parameters (M)\\\midrule
Visual tower, projection, and pooling &87.229\\
Text tower, projection, and pooling &55.731\\
Visual denoiser and heads &182.321\\
\quad of which convolutional renderer &1.423\\
Text denoiser and heads &57.845\\
Total, including null codes, SigLIP scale/bias, and schedule parameters &383.128\\
Total at $512^2$ (1024-entry positional tables in the visual tower and denoiser) &384.307\\\bottomrule
\end{tabular}
\end{table}

At inference the model uses only its own encoders and denoisers. Training additionally uses frozen VGG16 and DINOv2-B for decoded-image supervision \citep{simonyan2015vgg,dinov2}; these are training-only components, but they are not detached from the representation: their gradients pass through the image denoiser into the noisy representation, which stays attached to the image encoder (weights 0.1 and 0.01, only at $\gamma<1.695$). The T5 tokenizer defines token identities, but the embedding table and text encoder are learned from scratch. Synthetic captions carry knowledge from the LLaVA-NeXT captioner that produced them; we use the released captions \citep{captionemporium2024cc12m} as they are and did not re-audit them against the evaluation sets. Evaluation uses separate pretrained scorers---Inception for FID, DINOv2 and CLIP for feature distances, CLIP ViT-B/32 for TrueMatch and CLIPScore, Mask2Former for GenEval, mPLUG for DPG, and parsers, matchers, and a Qwen2.5-7B-Instruct judge for detailed captions---only for scoring (Appendix~\ref{app:protocols}).

\begin{table}[ht]\centering\small
\caption{\textbf{Training runs and exposure.} Every model we report, with what it trains and on what. Pair presentations count repeated examples.}
\label{tab:runs}
\setlength{\tabcolsep}{3pt}
\begin{tabularx}{\linewidth}{@{}>{\raggedright\arraybackslash}p{3.3cm} >{\raggedright\arraybackslash}X l r r@{}}\toprule
Model & Initialization; trained parameters & Data & Steps $\times$ batch & Presentations\\\midrule
\model{}; contrastive only; no contrastive loss & scratch; all (encoders only for contrastive only) & CC12M-R, $256^2$ & 125K $\times$ 8192 & 1.024B\\
\model{} + Gram/iBOT & \model{} at 24K; all, plus Gram/iBOT terms & CC12M-R, $256^2$ & 101K $\times$ 8192 & 1.024B in total\\
Frozen contrastive $\to$ generation & contrastive-only encoders; denoisers & CC12M-R, $256^2$ & 125K $\times$ 8192 & 2.048B in total\\
Final fine-tune, stage 1 & \model{} + Gram/iBOT; all & CC12M-R, $512^2$ & 10K $\times$ 5120 & +51.2M\\
Final fine-tune, stage 2 & stage 1; image denoiser and its null codes & 120K mixture, $512^2$ & \nrev{76K} $\times$ 5120 & \nrev{+389.1M}\\
Controlled fine-tunes (App.~\ref{app:finetune}) & \model{} + Gram/iBOT; all, or denoisers only & 120K mixture, $256^2$ & 5K $\times$ 8192 & +41.0M\\
Shared / separate generation space (App.~\ref{app:endpoint}) & \model{} at 24K; all & CC12M-R, $256^2$ & 16K $\times$ 8192 & +131M each\\\bottomrule
\end{tabularx}
\end{table}

Our base training sees about 1.85 times DREAM's pair presentations, on recaptioned CC12M; unique data and exposure are different resource dimensions (Table~\ref{tab:externaldata}). The visual denoiser costs 69.14 GFLOPs per forward pass at $256^2$ (MiniT2I-B/32: 313, a figure that includes its FLAN-T5-Large text encoder; we measure 311.9 for the same configuration). Wall-clock costs on one H100 are in Table~\ref{tab:latency}.

\paragraph{Training compute.} Table~\ref{tab:flops} counts model FLOPs (forward and backward, per training example at $256^2$; \texttt{torch} \texttt{FlopCounterMode} on each checkpoint's own configuration, matrix multiplies, convolutions and attention, no activation recomputation). The encoders and the alignment loss cost the same 205.8 GFLOPs in every arm; \model{}'s additional compute goes to the two denoisers (209.8 and 70.3) and the frozen perceptual networks (38.0), not to the encoders that understanding evaluates. A contrastive-only model would need about 318K steps instead of 125K to match \model{}'s training FLOPs; we did not train such a FLOP-matched control.
\begin{table}[!htbp]\centering\small
\caption{\textbf{Training FLOPs per example} (GFLOPs, forward + backward, $256^2$). Perceptual terms run on one eighth of each batch, Gram anchoring and iBOT on one quarter; the costs are averaged over the batch.}
\label{tab:flops}
\setlength{\tabcolsep}{4pt}
\resizebox{\linewidth}{!}{%
\begin{tabular}{@{}lrrrrrrr@{}}\toprule
Model & Encoders + SigLIP & $D_v$ & $D_t$ & VGG16 + DINOv2 & Gram + iBOT & Total & Ratio\\\midrule
Contrastive only & 205.8 & -- & -- & -- & -- & 205.8 & 1.00\\
\model{} & 205.8 & 209.8 & 70.3 & 38.0 & -- & 523.9 & 2.55\\
Contrastive only + Gram/iBOT & 205.8 & -- & -- & -- & 62.8 & 268.5 & 1.30\\
\model{} + Gram/iBOT & 205.8 & 209.8 & 70.3 & 38.0 & 62.8 & 586.6 & 2.85\\\bottomrule
\end{tabular}}\end{table}

\paragraph{Which losses move the image encoder.} Table~\ref{tab:gradshare} splits one training step's gradient on the image encoder (all 238 parameter tensors of $E_v$, $P_v$, $A_v$) by loss term, for \model{} + Gram/iBOT at 125K steps with its raw weights: two emulated training steps of 8192 real training pairs each, the trainer's loss weights, and training-distributed noise levels (the perceptual gate fired on 54\% of the supervised images). The DINOv2 term does reach the encoder, with a nonzero gradient on every tensor, but carries 0.11\% of its gradient norm (0.21\% when the gate fires; 0.13\% after Adam preconditioning with the checkpoint's own second moments) and is nearly orthogonal to the total. It acts mainly on the image denoiser, where it is 31\% of the gradient norm. Near convergence the encoder's gradient is dominated by SigLIP and iBOT, and the direct image-generation terms (pixel MSE, VGG, and the clean-representation regression) together contribute about 5\%. These are per-step shares at 125K steps, which are noise-dominated (the total gradient's cosine between the two steps is .12; a noise-free U-statistic estimate gives the same 0.12\% DINOv2 share); they say nothing about early training, and the retention results rest on the comparisons across models, not on gradient shares.
\begin{table}[!htbp]\centering\small
\caption{\textbf{Per-term gradient on the image encoder} (\model{} + Gram/iBOT, 125K steps; norm of each term's gradient as a fraction of the total gradient norm, and its cosine with the total; mean of two steps). Fractions sum to more than 100\% because the terms partly cancel.}
\label{tab:gradshare}
\setlength{\tabcolsep}{4pt}
\begin{tabular}{@{}lrr@{}}\toprule
Loss term & Fraction of norm & Cosine with total\\\midrule
SigLIP alignment & 92.4\% & $+.94$\\
iBOT & 33.3\% & $+.34$\\
Text denoising (token cross-entropy, through cross-conditioning) & 10.4\% & $+.20$\\
Image clean-representation regression & 2.6\% & $+.00$\\
Gram anchoring & 2.2\% & $-.01$\\
Pixel MSE & 1.4\% & $+.01$\\
VGG perceptual & 0.78\% & $-.02$\\
Text clean-representation regression & 0.48\% & $+.09$\\
DINOv2 perceptual & 0.11\% & $+.02$\\\bottomrule
\end{tabular}\end{table}

\section{Evaluation protocols}
\label{app:protocols}
Each benchmark has one frozen protocol, applied unchanged to every row we evaluate; released external models go through the same scorers, image sets, and views.

\subsection{Understanding}
\paragraph{ImageNet-1k.}
Zero-shot classification uses the entire 50K validation set, 1000 categories, and an 80-prompt ensemble. Full-train linear probing uses all 1,281,167 training images and evaluates on all 50K validation images. Three readouts are trained from the same frozen forward: average-pooled backbone $h$ (768 dimensions), average-pooled contextual representation $Z$ (256), and attention-pooled $g$ (512, ending in LayerNorm). Following the DREAM/MAE recipe \citep{li2026dream,he2022mae}, each head uses non-affine BatchNorm followed by a linear classifier, the LARS optimizer (SGD with momentum and a layer-wise adaptive learning rate; \citealp{you2017lars}) for 90 epochs, batch size 4096, base learning rate 0.1 scaled by batch/256, ten warmup epochs, and cosine decay. Random resized crops of size 256 are drawn from preprocessed 256-pixel image chunks. The primary score uses the last epoch. Inputs retain raw RGB $[0,1]$ scale. Confidence intervals resample validation images; they characterize test-sample uncertainty for a fitted head.

\paragraph{Retrieval, compositional matching, and dense prediction.}
COCO retrieval uses the Karpathy test split: 5000 images and 25,000 reference captions. We report R@1/5/10 in each direction and their mean. SugarCrepe uses its seven full official splits (1652/788/1406/245/666/2062/692 items) and ARO the full VG-Attribution and VG-Relation splits (28,748/23,937 items) \citep{hsieh2023sugarcrepe,yuksekgonul2023aro}; both score captions by global image--text similarity. The primary ADE20K evaluation uses full-resolution ADEChallengeData2016, with 20,210 training and 2000 validation images. The head is linear, as in DINOv3 \citep{dinov3}: channel dropout (0.1), BatchNorm, and a $1\times1$ convolution, i.e.\ one 150-class linear classifier shared by all patch tokens; its logits are bilinearly upsampled to pixel resolution. It fits frozen $h$ or $Z$ for 40K iterations, batch size 16, and 512-pixel crops. Sliding-window evaluation uses crop/stride 512/341. The learning rate is $10^{-3}$, as in the DINOv3 configuration; DREAM states $10^{-4}$ \citep{li2026dream,dinov3}. The last iteration supplies the primary score.

The readout-accessibility diagnostic uses a twofold-downsampled ADE20K package, 8000 training images, 512-pixel input, and 40 epochs. It compares linear SGD, linear AdamW with training-only ZCA, and MLP AdamW on fixed features. This reduced-resolution study isolates readout effects; the primary full-resolution protocol supplies the external comparison.

\textbf{Probe settings.} \emph{Reduced-resolution diagnostic} (Table~\ref{tab:ladder}): heads train on cached patch features (minibatches of 8192 patches, 40 epochs, learning rate 0.01, cosine annealing, weight decay $10^{-4}$) and are evaluated once, after the last epoch, on all 2000 validation images; one run per cell. Linear SGD uses momentum 0.9; ZCA whitening is fitted on training features; the MLP is Linear($C$, 512) $\to$ GELU $\to$ Linear(512, 150), trained with AdamW. \emph{Full-resolution protocol} (Table~\ref{tab:ladderfull}): the linear, MLP and attentive heads share the data, crops, batch size, 40K iterations and sliding-window evaluation above, with AdamW at learning rate $10^{-3}$, weight decay $10^{-3}$, and 1.5K warm-up iterations then cosine decay; we report the last iteration, with no selection on validation images. Each head starts with channel dropout (0.1) and BatchNorm. The linear head adds a $1\times1$ convolution to 150 classes, the MLP $1\times1$ convolutions $C\to2048\to150$ with GELU, and the attentive probe a projection to 512 channels, fixed 2-D sine--cosine positions and two pre-norm transformer blocks (eight heads) before the classifier. Probe seeds change only the head's initialization (number and spread in Table~\ref{tab:ladderfull}). NYU Depth v2 uses the same heads over 256 depth bins (10K iterations).

\subsection{Image generation and reconstruction}
\label{app:fid}
\paragraph{T2I sampling.}
GenEval and DPG use the native fitted visual ladder, 128 steps, CFG 3, and churn 1 at the model's native resolution ($256^2$; $512^2$ for the fine-tuned model). Each benchmark uses its prescribed prompts and seed multiplicity. Quantitative images receive no reranking, except for second GenEval numbers marked as own-score selection among the four samples per prompt, reported with the four-sample average of the same images in Appendix~\ref{app:ownselect}. GenEval uses 553 prompts and four images per prompt, evaluated by its object-focused scorer \citep{ghosh2023geneval}. DPG uses 1065 prompts and four images per prompt with the mPLUG VQA evaluator \citep{hu2024ella}. Overall DPG scores average the dependency-aware crop scores. Category columns use the last crop's category averages, while the overall score averages dependency-aware scores across all four crops. Each aggregation is reported under its released definition.

COCO FID uses 30K seeded sampled val2014 images, the first annotated caption for each image, and one generated image per prompt. Real references are the paired 30K images resized on the short side to 256 and center-cropped. Features come from pytorch-fid's Inception-2015 checkpoint \citep{heusel2017fid}. FD-DINOv2 and FD-CLIP are feature-distance diagnostics with separate numerical scales. Real-versus-real estimates characterize finite-sample bias: for fixed-recipe runs a pessimistic half-split of the reference set (4.405 at 15K per side), and for development-selected runs a disjoint 10K-versus-30K draw (4.387). Perfectly matched feature distributions have zero FID.

\paragraph{Development-selected COCO recipe.}
A disjoint 5K development set selects among 64/128 steps, CFG peaks 2/3/4/5, and constant/linear guidance, with churn 1 throughout. The rule minimizes development FID, then maximizes CLIPScore among configurations within 0.2 FID of that minimum. This selects 64 steps with guidance increasing from 1 toward 5 linearly in noise-level progress (development FID 20.181; minimum 20.034). The core-model selection is applied unchanged to every other checkpoint, once on the 30K test prompts. Each prompt receives one generated image. The reported 128 NFE is exact: the first step runs at guidance weight 1 and therefore needs one branch, the remaining 63 steps need two, and one final conditional evaluation renders pixels ($1+2\times63+1$). End-to-end timing additionally includes the source encoder.

\paragraph{Sampler for the $512^2$ models.} A 54-recipe sweep (guidance 3--7, constant or rising, 32--128 steps, churn 0/1, three terminal noise levels) on a held-out GenEval-style development set (439 prompts from the official prompt generator with a different seed, every test prompt removed), run on a $512^2$ fine-tune selects 64 steps, constant CFG 5, and churn 1; the top of the ranking is flat (CFG 5 and 6 within .001). On the test prompts it changes GenEval by less than .01 relative to the fixed recipe (fine-tuned model at 8K steps: .737 vs.\ .729) and raises DPG by 1.6--2.0 (79.75 vs.\ 77.79). The tables keep the fixed recipe; the qualitative samples use the selected one.

\paragraph{One recipe per benchmark.} Prompt-following benchmarks and FID therefore use different samplers, as in DREAM, which reports its best FID and CLIP score over constant and linear CFG schedules without stating a held-out selection set \citep{li2026dream}; our FID recipe is chosen on disjoint development prompts and run once on the test prompts. The core model's choice is applied unchanged to every other checkpoint; re-selecting guidance per checkpoint would favor them slightly (development FID 19.86 at CFG 3 vs.\ 20.38 for + Gram/iBOT). Swapping the recipes costs each metric: the FID-selected recipe gives GenEval .430 and DPG 74.52 for \model{} (fixed recipe: .461 and 74.43), and the fixed recipe gives COCO-30K FID 13.87 (selected: 10.10).

\paragraph{Training-distribution image synthesis.}
CC12M evaluation draws 50K reference images and a disjoint 50K real control from our CC12M training images (CC12M has no held-out split; DREAM does not state whether its 50K are held out), using original alt-text prompts, as DREAM does, and the development-selected recipe. $512^2$ generations are first brought to $256^2$ with the references' own transform (bicubic antialiased short-side resize and center crop). FID measures the training distribution; the disjoint real-versus-real estimate is 1.442.

\paragraph{Reconstruction.}
Reconstruction starts from the encoded image with low-noise corruption at $\gamma=-3.34$ (the visual ladder's terminal level), then predicts RGB with one denoiser call. Off-the-shelf VAEs are scored on the identical 5K images, preprocessing, and metrics, decoding their posterior mode. The 5K images are a fixed seeded draw from ImageNet-1k validation; each is resized on its short side to 256 (bicubic) and center-cropped to $256^2$ in $[0,1]$, with no training augmentation. PSNR is computed per image on $[0,1]$ and averaged; rFID compares the 5K reconstructions with the same 5K preprocessed originals (Inception-2015). The main tables use the full-condition route, which supplies the same image's $(g_v,Z_v)$ as global and cross-attention inputs in addition to the noised representation; the null route supplies learned null conditions. A 50K-image rFID is reported for \model{} and \model{} + Gram/iBOT. A stochastic contextual reconstruction and a deterministic autoencoder can be compared descriptively when sample counts and preprocessing are stated.

\subsection{Image-to-text generation}
\paragraph{A fixed development decision.}
The I2T recipe is selected on 1000 Karpathy validation images, hash-ordered after excluding all test IDs and images used by TrueMatch and the COCO portion of DetailCaps. We compare Heun-64 (CFG 5) with ancestral DDPM-64 (CFG 5, $\eta=1$), using each checkpoint's learned text ladder. Both disable self-conditioning and use argmax vocabulary readout of the conditional branch at the last ladder step. The decision maximizes the mean of SPICE and CAPTURE, subject to a CHAIR$_s$ degradation of at most five points; ties within 0.1 favor Heun. DDPM obtains 23.73 versus 23.24 and is frozen for all checkpoints.

Noise is generated per sample on a CPU fp32 stream seeded from dataset, image ID, sample index, and a fixed experiment seed. Evaluation batch size is 64. COCO uses a 256-pixel center crop; fine-grained and detailed-caption benchmarks use long-side resizing and centered gray padding to preserve boundary content. Scores use the complete EOS-normalized text; raw token IDs and truncation flags are retained.

\paragraph{Caption quality and hallucination.}
COCO uses 5000 test images with all five references and the official pycocoevalcap scorers, including CIDEr-D and SPICE \citep{vedantam2015cider,anderson2016spice}. CHAIR$_i$ is the fraction of mentioned object instances deemed hallucinated; CHAIR$_s$ is the fraction of captions containing a hallucinated object \citep{rohrbach2018chair}. CLIPScore uses ViT-B/32 and $2.5\max(0,\cos)$; RefCLIPScore also incorporates references \citep{hessel2021clipscore}. These scales are retained throughout. BLIP uses the released \texttt{blip-image-captioning-base} checkpoint (revision 82a3776), whose weights are converted from BLIP's pretrained ViT-B CapFilt-L model rather than its COCO caption fine-tune \citep{li2022blip}, with prompt ``a picture of'', beam size three, and at most 20 tokens as in BLIP's COCO evaluation.

\paragraph{Discrimination and detailed coverage.}
TrueMatch uses the official 254/104/93 bags of size 3/5/7 and the frozen OpenAI CLIP ViT-B/32 listener. Every bag image is a target once. The listener truncates at 77 tokens; ties count as hits following the official implementation. Our scorer reproduces the benchmark's released caption sets to within 0.1 points (HolisticCap 57.5/48.0/49.2 vs.\ 57.5/48.1/49.1).

DetailCaps-4870 uses three released references and CAPTURE 0.1.13 with FactualSceneGraph 0.5.0, its Flan-T5 parser, and MPNet matcher \citep{dong2024capture}. A parser limit of 64 tokens per sentence motivates reporting truncation. CompreCap uses its 560 annotated images, spaCy parsing, MiniLM matching, and a Qwen2.5-7B-Instruct judge in its stock configuration \citep{lu2025comprecap}, with greedy judge decoding and a single BOS token; external captions are scored under the same settings. For CAPTURE, the rescored external captions match the benchmark's three-reference README values (its arXiv table reports two-reference values: CogVLM 60.20, ShareCaptioner 59.55). DOCCI-CAPTURE transfers the CAPTURE metric to one human reference per image \citep{onoe2024docci} and is labeled as a metric transfer.

\section{External comparisons and resource context}
\label{app:external}
Here we give component and data accounting for every external model in the main tables, complete prompt-following profiles, and the image-distribution protocols behind each FID. Each model row cites its original work. \published{Blue}: published by the cited source. \oss{Teal}: open-source checkpoints or released outputs evaluated by us with our scorers. Parameter counts of released models were computed from tensor shapes in the official checkpoints (headers only), counting shared embeddings once.

\paragraph{Notes to Tables~\ref{tab:understanding}--\ref{tab:t2i}.}\label{app:tablenotes}
\emph{Table~\ref{tab:understanding}.} Classification, retrieval and SugarCrepe in \%; ADE20K in mIoU. CC12M-R: VLM-recaptioned CC12M (LLaVA-NeXT for ours, DreamLIP for COSMOS); CC12M-L: LLM-rewritten captions. Published LP/ADE results use each model's own probe protocol. $^a$DREAM-B at $\sigma{=}.45$. $^b$Retrieval and SugarCrepe from the released SLIP-repository CLIP. FLUID-L as reimplemented by DREAM. $^\sharp$The fine-tuned model has 1.2M additional positional parameters and a different resolution from the $256^2$ rows, so its higher zero-shot accuracy cannot be attributed to fine-tuning alone.
\emph{Table~\ref{tab:i2t}.} CHAIR$_i$ and TrueMatch R@1 in \%. $^a$Beam size 3, max.\ 20 tokens. $^b$CLIP reranking of 64 samples. $^d$Trained on 5.4B generated annotations; \texttt{\textless CAPTION\textgreater} for COCO, \texttt{\textless DETAILED\_CAPTION\textgreater} otherwise. $^e$Entries other than the second numbers use the pretrained model. CIDEr-D is length-sensitive (ours: 43 words; references: ${\sim}10$); \model{} scores substantially below autoregressive captioners.
\emph{Table~\ref{tab:t2i}.} GenEval on a 0--1 scale, DPG on 0--100, PSNR in dB. $^a$CLIP reranking of 32 samples. $^b$FID of the CC12M-pretraining checkpoint at $512^2$, before the 120K fine-tune. Second MiniT2I FIDs: released pretraining checkpoints, our protocol. $^c$After the 120K fine-tune (released fine-tuned MiniT2I-B/16: 27.35). Distilled rows: students of the fine-tuned model of the row above; with four jumps (5 NFE), they reach \nrev{.799 / 78.55 / 34.08 and CC12M-50K FID 16.60}; students of the pretrained model are in Table~\ref{tab:distillscale}. Distilled students are text-conditional samplers with the guidance scale distilled in; they share their teacher's encoders, so reconstruction (PSNR, rFID) is a property of the teacher rows.

\subsection{Model components and training data}
We count every parameter the reported system needs at inference, including frozen text encoders, language models, and image codecs; text modules and codecs are listed separately but are included in the total. \model{}'s 383M include both generative directions, its 114M text pathway, and its 1.42M renderer; its VGG/DINOv2 perceptual losses and Gram/iBOT head are training-only.

\begin{table}[!htbp]\centering\footnotesize
\caption{\textbf{System components (M parameters).} Totals include every component needed at inference. $\sim$: computed from released tensor shapes or configurations.}
\label{tab:components}
\setlength{\tabcolsep}{3pt}
\begin{tabularx}{\linewidth}{@{}>{\raggedright\arraybackslash}p{3.1cm} r Y Y@{}}\toprule
Model & Total & Language component & Visual component or image codec\\\midrule
\corerow \textbf{\model{}} &383.1 (384.3 at $512^2$)&55.7 encoder + 57.8 generator&87.2 encoder + 182.3 generator (incl.\ 1.42 renderer)\\\midrule
\multicolumn{4}{@{}l}{\emph{Understanding and captioning}}\\
CLIP / LaCLIP / LaSLIP / SLIP-B/16 \citep{radford2021clip,fan2023laclip,mu2022slip}&$\sim$150&$\sim$63&$\sim$86 ViT-B/16\\
COSMOS-B/16 \citep{kim2025cosmos}&$\sim$150&CLIP text tower&ViT-B/16\\
BLIP-base captioner \citep{li2022blip}&224.0&137.9 BERT-base decoder with cross-attention&86.1 ViT-B/16\\
NDLB \citep{gaur2025ndlb,mokady2021clipcap}&$\sim$244&124.4 GPT-2 + 31.5 mapper&87.8 CLIP ViT-B/32\\
CoCa-L/14 \citep{yu2022coca}&$\sim$638&331.7 (unimodal + multimodal)&306.7 ViT-L/14\\
Florence-2-base \citep{xiao2024florence2}&230&BART-style encoder--decoder&DaViT\\
LLaVA-1.5-7B \citep{liu2024llava15}&7063&6740 Vicuna-7B&CLIP ViT-L/14-336\\
ShareCaptioner \citep{chen2023sharegpt4v}&$\sim$8401&7322 InternLM-7B&$\sim$1079 EVA-ViT-g + Q-Former\\
Cambrian-1-8B \citep{tong2024cambrian}&$\sim$11049&8030 Llama-3-8B&$\sim$3018: four frozen encoders + SVA\\
InstructBLIP-13B \citep{dai2023instructblip}&$\sim$14192&$\sim$13017 Vicuna-13B&$\sim$1176 ViT-g + Q-Former\\
CogVLM \citep{wang2023cogvlm}&$\sim$17640&Vicuna-7B + visual expert&EVA-CLIP-E\\
LLaVA-NeXT-34B \citep{liu2024llavanext}&$\sim$34750&$\sim$34390 Yi-34B&CLIP ViT-L/14-336\\
GPT-4o \citep{openai2024gpt4o}&undisclosed&undisclosed&undisclosed\\\midrule
\multicolumn{4}{@{}l}{\emph{Image generation}}\\
MAGVLT-B \citep{kim2023magvlt}&447&integrated (371 transformer)&$\sim$76 VQ-GAN\\
MiniT2I-B/32 / B/16 \citep{minit2i2026}&601 / 599&341 frozen FLAN-T5-Large&none (pixels; 260 / 258 denoiser)\\
SD1.5 \citep{rombach2022ldm}&$\sim$1066&$\sim$123.1 CLIP-L text&$\sim$83.7 VAE ($\sim$859.5 U-Net)\\
SDXL base \citep{podell2023sdxl}&$\sim$3469&$\sim$123.1 + 694.7 CLIP-L + OpenCLIP-bigG&$\sim$83.7 VAE ($\sim$2567.5 U-Net)\\
DREAM-B / DREAM-L \citep{li2026dream}&$\sim$5176 / 5482&66 contrastive text + $\sim$4762 frozen T5-XXL&$\sim$83.7 SD kl-f8 ft-EMA (330 / 636 core, incl.\ text)\\
FLUID-L, DREAM impl.\ \citep{fan2024fluid,li2026dream}&$\sim$5416&$\sim$4762 frozen T5-XXL&$\sim$83.7 SD-VAE (570 core)\\
UniSpace \citep{unispace2026}&$\sim$17200&integrated: two Qwen3-8B experts, $\sim$8190 each&$\sim$836 PR-Qwen-ViT tokenizer + decoder\\\bottomrule
\end{tabularx}
\end{table}

The GenEval .55 for SDXL uses the base pipeline without the refiner (the official GenEval generation script loads only the base model). DREAM uses the frozen \texttt{google/t5-v1\_1-xxl} encoder and the SD kl-f8 ft-EMA VAE. UniSpace's 8B label is the active size per token of a model with two fully duplicated experts. The CoCa row is the COCO-finetuned open\_clip release; our TrueMatch run of it reproduces the published value within 0.6.\footnote{Component sources: \href{https://huggingface.co/stable-diffusion-v1-5/stable-diffusion-v1-5}{SD1.5}, \href{https://huggingface.co/stabilityai/stable-diffusion-xl-base-1.0}{SDXL base}, \href{https://huggingface.co/stabilityai/sd-vae-ft-ema}{SD-VAE}, \href{https://huggingface.co/google/t5-v1_1-xxl}{T5-XXL}, \href{https://huggingface.co/yjb6/UniSpace}{UniSpace}, \href{https://huggingface.co/MiniT2I/MiniT2I}{MiniT2I}.}

\begin{table}[!htbp]\centering\footnotesize
\caption{\textbf{Training data and inherited models.} Presentations count repeated examples.}
\label{tab:externaldata}
\setlength{\tabcolsep}{3pt}
\begin{tabularx}{\linewidth}{@{}>{\raggedright\arraybackslash}p{.28\linewidth} Y Y@{}}\toprule
System & Task-training data & Exposure / inherited resources\\\midrule
\corerow \textbf{\model{}}&CC12M; LLaVA-NeXT long/short recaptions and alt-text&1.024B presentations; frozen VGG/DINOv2 perceptual losses (training only)\\
\corerow \quad + final fine-tune&CC12M-R at $512^2$, then MiniT2I's 120K synthetic mixture&+51.2M, then \nrev{+389.1M} presentations (image denoiser only)\\
CLIP / LaCLIP / LaSLIP \citep{fan2023laclip}&CC12M; LaCLIP/LaSLIP add LLaMA caption rewrites&35 epochs, batch 8192\\
COSMOS \citep{kim2025cosmos}&10.0M CC12M pairs with DreamLIP synthetic captions&32 epochs, batch 4096; EMA self-distillation\\
DREAM \citep{li2026dream}&11.3M CC12M pairs, alt-text&49 epochs ($\sim$0.554B presentations); frozen T5-XXL, SD-VAE\\
MiniT2I-B/32 \citep{minit2i2026}&CC12M long recaptions; then the 120K mixture&JAX reference configuration: 128M + 20.48M presentations; the PyTorch port uses twice the batch (256M + 41M); the blog does not state which produced its numbers; frozen FLAN-T5-Large; $512^2$\\
MAGVLT \citep{kim2023magvlt}&CC3M, CC12M, SBU, VG ($\sim$17M)&pretrained VQ-GAN; CLIP reranking at evaluation\\
BLIP \citep{li2022blip}&129M pairs with CapFilt bootstrapped captions&ImageNet-pretrained ViT, BERT initialization\\
NDLB \citep{gaur2025ndlb}&COCO captions, then self-retrieval fine-tuning&frozen CLIP ViT-B/32, pretrained GPT-2\\
InstructBLIP / Cambrian-1 / LLaVA&instruction mixtures (13 datasets / 2.5M + 7M / 558K + 665K; LLaVA-NeXT 558K + 760K)&pretrained LLMs and vision encoders\\
ShareCaptioner / CogVLM \citep{chen2023sharegpt4v,wang2023cogvlm}&100K detailed GPT-4V captions on InternLM-XComposer / a filtered 1.5B image--text pool plus grounding and instruction data&pretrained LLMs and vision encoders\\
SD1.5 \citep{rombach2022ldm}&LAION subsets&$\sim$3.16B presentations; frozen CLIP-L text, VAE\\
SDXL \citep{podell2023sdxl}&undisclosed mixture&frozen CLIP-L and OpenCLIP-bigG text encoders, VAE\\
UniSpace \citep{unispace2026}&generation, editing, and understanding mixture&$\sim$510M presentations; Qwen3-8B and a frozen visual tokenizer\\\bottomrule
\end{tabularx}
\end{table}

\subsection{Prompt following: external profiles and the MiniT2I gap}
GenEval evaluates six constraints; DPG adds five question categories and a dependency-aware overall score. Published values keep their source precision; MiniT2I values follow its blog and JAX release. Our own profiles are in Table~\ref{tab:promptours}.

\begin{table}[!htbp]\centering\footnotesize
\caption{\textbf{GenEval and DPG-Bench profiles of external systems.} GenEval tasks in percent, overall as a fraction. SD rows: official GenEval release \citep{ghosh2023geneval} and \citet{hu2024ella}, Table~4; others: their papers or releases. Teal: the released MiniT2I-B/16 through our pipeline.}
\label{tab:promptext}
\setlength{\tabcolsep}{2.6pt}
\resizebox{\linewidth}{!}{%
\begin{tabular}{@{}lrrrrrrr|rrrrrr@{}}\toprule
&\multicolumn{7}{c|}{GenEval}&\multicolumn{6}{c}{DPG-Bench}\\
Model & Single & Two & Count & Color & Position & Attr. & Overall & Global & Entity & Attribute & Relation & Other & Overall\\\midrule
SD1.5 \citep{rombach2022ldm}&\published{97}&\published{38}&\published{35}&\published{76}&\published{4}&\published{6}&\published{.43}&\published{74.63}&\published{74.23}&\published{75.39}&\published{73.49}&\published{67.81}&\published{63.18}\\
SDXL \citep{podell2023sdxl}&\published{98}&\published{74}&\published{39}&\published{85}&\published{15}&\published{23}&\published{.55}&\published{83.27}&\published{82.43}&\published{80.91}&\published{86.76}&\published{80.41}&\published{74.65}\\
UniSpace \citep{unispace2026}&\published{98}&\published{92}&\published{69}&\published{88}&\published{83}&\published{73}&\published{.84}&\published{84.80}&\published{92.26}&\published{90.00}&\published{94.97}&\published{88.80}&\published{86.49}\\
MiniT2I-B/32, CC12M pretraining \citep{minit2i2026}&&&&&&&\published{.505}&&&&&&\published{78.3}\\
\quad + 120K fine-tuning&&&&&&&\published{.826}&&&&&&\published{82.3}\\
MiniT2I-B/16, + 120K fine-tuning \citep{minit2i2026}&&&&&&&\published{.873}&&&&&&\published{84.2}\\
\quad our scorers, $512^2$&\oss{99.7}&\oss{95.7}&\oss{79.1}&\oss{92.8}&\oss{81.5}&\oss{78.3}&\oss{.878}&&&&&&\oss{84.12}\\
\quad our scorers, downsampled to $256^2$&\oss{99.7}&\oss{95.2}&\oss{76.9}&\oss{93.4}&\oss{82.8}&\oss{80.0}&\oss{.880}&&&&&&\oss{84.65}\\\bottomrule
\end{tabular}}
\end{table}

MiniT2I-B/32 and \model{} both use 256 image tokens and the same LLaVA-NeXT CC12M recaptions \citep{captionemporium2024cc12m}. After CC12M pretraining, MiniT2I-B/32 reaches GenEval/DPG .505/78.3 and \model{} .461/74.43; after fine-tuning on the 120K mixture, MiniT2I-B/32 reaches .826/82.3 and our fine-tuned model \nrev{.806/79.29}. Three measurements narrow down the gap. First, the released MiniT2I-B/16 run through our pipeline reproduces its published scores (.878/84.12 vs.\ .873/84.2) and gives the same scores when its images are downsampled to $256^2$ (.880/84.65), so neither our scorers nor our output size penalize a model. Second, fine-tuning for MiniT2I-B/16's sample budget (5000 steps at batch 8192) reaches .732 at $256^2$ (Appendix~\ref{app:finetune}), so fine-tuning exposure alone does not close the gap. Third, the fine-tuned model's remaining deficit to MiniT2I-B/16 is concentrated in position (\nrev{58.3} vs.\ 81.5) and attribute binding (\nrev{67.5} vs.\ 78.3)\nrev{, with counting close (78.1 vs.\ 79.1)}, while single objects and colors are near ceiling for both (Tables~\ref{tab:promptext} and~\ref{tab:promptours})---the language-compositional tasks where a pretrained text encoder is most likely to help. The other differences are the frozen 341M FLAN-T5-Large text encoder versus our text encoder learned from captions alone, denoiser compute (313 vs.\ 69 GFLOPs per forward at 256 image tokens), caption mix, and the number of fine-tuning optimizer steps (40K vs.\ \nrev{76K}; 20.48M vs.\ \nrev{389.1M} presentations).\footnote{MiniT2I \href{https://peppaking8.github.io/\#/post/minit2i}{blog}; \href{https://github.com/PeppaKing8/minit2i-jax}{JAX} and \href{https://github.com/Hope7Happiness/minit2i-torch}{PyTorch} releases.}

\subsection{Image-distribution comparisons}
Single-sample FID depends on resolution, reference preprocessing, prompt field, and guidance selection, so each row keeps its protocol. DREAM does not state whether its CC12M evaluation images are held out or which CLIP backbone its similarity uses; our 100$\cos$ uses ViT-B/32. Numbers that select among several samples per prompt are not comparable and are omitted: DREAM-L with semantically aligned decoding (COCO 10.4, candidate count not stated for COCO) and OFA-L (10.5, COCO-finetuned, best of 24 by CLIP) \citep{li2026dream,ofa2022}.
\begin{table}[!htbp]\centering\small
\caption{\polish{\textbf{Zero-shot COCO-30K and training-distribution CC12M-50K FID.} Our rows: single sample per prompt, development-selected recipe (64 steps, 128 NFE), paired COCO references (short side 256, center crop), pytorch-fid Inception; the fine-tuned model generates at $512^2$ and is brought to $256^2$. External CC12M values from \citet{li2026dream}. Teal: released MiniT2I checkpoints sampled by us ($512^2$ brought to $256^2$, our references); where two COCO values are given, the first is published. $^a$128 steps (COCO-30K: 64 steps; CC12M at 64 steps: \published{4.57} / \published{29.1}), best of two guidance schedules per metric. $^b$Published against precomputed $512^2$ COCO statistics. $^c$CLIP reranking of 32 samples.}}
\label{tab:fidexternal}
\setlength{\tabcolsep}{5pt}
\begin{tabular}{@{}lrrrl@{}}\toprule
&COCO-30K&\multicolumn{2}{c}{CC12M-50K}&\\\cmidrule(lr){2-2}\cmidrule(lr){3-4}
Model & FID$\downarrow$ & FID$\downarrow$ & 100$\cos$ & \polish{Inherited}\\\midrule
FLUID-L, DREAM impl.\ \citep{fan2024fluid,li2026dream} &\published{9.62}&\published{4.53}&\published{30.0}&T5-XXL, VAE\\
DREAM-L$^a$ \citep{li2026dream} &\published{10.0}&\published{4.34}&\published{29.4}&T5-XXL, VAE\\
REPA, DREAM impl.\ \citep{yu2024repa} &--&\published{4.42}&\published{29.9}&published recipe\\
MAGVLT-B$^c$ \citep{kim2023magvlt} &\published{12.08}&--&--&VQ-GAN\\
MiniT2I-B/32, CC12M pretraining \citep{minit2i2026} &\published{13.69} / \oss{13.42}&\oss{9.51}&--&FLAN-T5-Large\\
MiniT2I-B/16, CC12M pretraining$^b$ \citep{minit2i2026} &\published{10.51} / \oss{15.54}&\oss{11.35}&--&FLAN-T5-Large\\
\quad + 120K fine-tuning &\oss{27.35}&\oss{13.00}&--&FLAN-T5-Large\\\midrule
\corerow \model{} &10.10&4.33&29.33&none\\
\corerow \quad + Gram/iBOT &10.20&4.37&29.30&none\\
\corerow \qquad + final fine-tune &\nrev{31.81}&\nrev{15.27}&--&none\\\bottomrule
\end{tabular}
\end{table}

Fine-tuning on the synthetic mixture moves every model away from both natural-image distributions: our mixture-only fine-tunes rise to 13--16 on CC12M-50K and 28--33 on COCO-30K (Table~\ref{tab:ftfull}), and the released MiniT2I-B/16 scores 27.35 on COCO under our protocol. MiniT2I's reported FID is that of its pretraining checkpoint.

\section{Controlled interventions in joint learning}
\label{app:ablations}
Every intervention below is evaluated on all three axes. Understanding: ImageNet zero-shot (ZS), COCO 5K retrieval mean recall (mR), the reduced-resolution ADE20K diagnostic on the contextual representation (MLP readout; the full-resolution protocol is reserved for the main models), and SugarCrepe (SC). Image-to-text: COCO 5K SPICE and CHAIR$_i$ with the frozen DDPM-64 recipe. Text-to-image: GenEval and DPG with the fixed 128-step recipe, COCO-30K FID with the development-selected recipe of the main tables, and ImageNet 5K reconstruction PSNR through the full-condition route.

\subsection{Which conditioning routes shape the representation?}
The self route conditions the target denoiser on its own global readout with a null local key. Removing it and redistributing its probability either proportionally or entirely to the null route leaves understanding and prompt following nearly unchanged but breaks full-condition reconstruction (Table~\ref{tab:routes}): with the target's own $(g_v,Z_v)$ supplied as condition, PSNR falls from 24.1 to 9--11~dB, while the null-conditioned reconstruction of the same models is unaffected (about 24.2~dB). Without the self route, the denoiser cannot use its own modality's representation as a condition, a failure invisible to benchmarks that condition on the other modality.

\begin{table}[h]\centering
\caption{\textbf{Conditioning-route ablation.} Route probabilities are (null, self, cross).}
\label{tab:routes}
\setlength{\tabcolsep}{3pt}
\resizebox{\linewidth}{!}{%
\begin{tabular}{@{}lrrrrrrrrrr@{}}\toprule
&\multicolumn{4}{c}{Understanding}&\multicolumn{2}{c}{Image $\to$ text}&\multicolumn{4}{c}{Text $\to$ image}\\\cmidrule(lr){2-5}\cmidrule(lr){6-7}\cmidrule(lr){8-11}
Routes & ZS & mR & ADE (MLP) & SC & SPICE & CHAIR$_i\!\downarrow$ & GenEval & DPG & FID$\downarrow$ & PSNR\\\midrule
\corerow \textbf{\model{} (.10, .20, .70)} &50.63 &70.34 &26.22 &80.6 &13.53 &11.2 &.461 &74.43 &10.10 &24.09\\
Self off, proportional (.125, 0, .875) &50.93 &70.79 &26.78 &79.1 &14.02 &10.7 &.439 &73.30 &10.13 &9.45\\
Self off, to null (.30, 0, .70) &50.44 &70.15 &26.62 &80.3 &13.74 &10.1 &.428 &72.80 &10.61 &10.93\\\bottomrule
\end{tabular}}
\end{table}

\subsection{How does the balance of objectives shape the representation?}
Table~\ref{tab:weights} sweeps the alignment weight with the generative weights fixed, from no alignment (weight 0, the no-contrastive-loss control, which still trains both denoisers) through 0.25, 1, and 4, together with the contrastive-only extreme (no generative loss). Once alignment is present, global understanding is nearly flat across a 16-fold range of its weight (ZS 49.9--51.2). Prompt following peaks at weight 1, and captioning degrades at weight 4 (SPICE 12.98). Reconstruction improves monotonically as alignment weakens (26.91~dB without it), and prompt following collapses without it (GenEval .193): alignment is what lets a text representation place a prompt. The default weight of 1 is the only setting that is near the best on all three axes. Adding Gram/iBOT to the contrastive-only model shows that the structural terms help either objective: its full-resolution ADE20K mIoU rises to 40.00 at the backbone and 33.13 at the representation (Table~\ref{tab:understanding}), so the joint model's linear-probe deficit persists under matched structural regularization, while its global readouts match (ZS 50.90 vs.\ 50.63, retrieval 70.88 vs.\ 70.87). COCO FID is nearly flat across aligned models (10.1--10.6) while prompt following varies (GenEval .427--.461), a reminder that distributional fidelity and prompt fidelity are distinct.

\begin{table}[h]\centering
\caption{\textbf{Objective balance.} Alignment weight $\lambda_{\rm align}$ with generative weights fixed; the contrastive-only control removes the generative losses instead. Gram anchoring alone (no iBOT, from 24K steps, $\lambda_{\rm Gram}=2$) leaves the dense and linear-probe readouts at the \model{} level (ImageNet linear probe on $h$ 72.34 vs.\ 72.82); the gains appear only with iBOT added (73.62). An iBOT-only arm was not run.}
\label{tab:weights}
\setlength{\tabcolsep}{3pt}
\resizebox{\linewidth}{!}{%
\begin{tabular}{@{}lrrrrrrrrrr@{}}\toprule
&\multicolumn{4}{c}{Understanding}&\multicolumn{2}{c}{Image $\to$ text}&\multicolumn{4}{c}{Text $\to$ image}\\\cmidrule(lr){2-5}\cmidrule(lr){6-7}\cmidrule(lr){8-11}
Objective & ZS & mR & ADE (MLP) & SC & SPICE & CHAIR$_i\!\downarrow$ & GenEval & DPG & FID$\downarrow$ & PSNR\\\midrule
No contrastive loss ($\lambda_{\rm align}=0$) &1.18 &1.51 &23.48 &53.8 &13.86 &10.1 &.193 &45.31 &23.20 &26.91\\
$\lambda_{\rm align}=0.25$ &51.20 &71.28 &26.49 &79.8 &14.14 &10.2 &.428 &72.24 &10.60 &25.29\\
\corerow \textbf{\model{}: $\lambda_{\rm align}=1$} &50.63 &70.34 &26.22 &80.6 &13.53 &11.2 &.461 &74.43 &10.10 &24.09\\
$\lambda_{\rm align}=4$ &49.87 &69.89 &27.12 &79.9 &12.98 &11.2 &.427 &71.80 &10.17 &22.71\\
Contrastive only (no generation) &49.51 &69.75 &27.21 &79.4 &-- &-- &-- &-- &-- &--\\\midrule
\corerow \model{} + Gram/iBOT &50.90 &70.88 &28.51 &80.0 &13.50 &11.2 &.448 &73.49 &10.20 &24.03\\
Contrastive only + Gram/iBOT &50.63 &70.87 &29.76 &79.6 &-- &-- &-- &-- &-- &--\\\bottomrule
\end{tabular}}
\end{table}

\subsection{Does generation need the understanding representation itself?}
\label{app:endpoint}
The controls above show that co-adaptation substantially improves generation; they do not show that the space generation runs in must be the representation that understanding reads. We therefore train a matched pair that differs in one configuration key. Both arms continue the \model{} checkpoint at 24K training steps for 16K steps at batch 8192 and $256^2$ (EMA initialization, fresh AdamW, learning rate $1.3\times10^{-3}$ with 1000 warm-up steps and cosine decay, same data order and seed). In the \emph{shared} arm the image denoiser generates $Z_v=\operatorname{LN}(P_vh_v)$, as in \model{}. In the \emph{separate} arm it generates $\operatorname{LN}(P^{\rm gen}_vh_v)$ through a second 768-to-256 projection (+0.2M parameters) initialized as a copy of $P_v$, so both arms start from the same function and differ only in where generation's gradient enters the encoder. Alignment, text conditioning, and every probe still read $Z_v$ in both arms.

\begin{table}[h]\centering
\caption{\textbf{Shared versus separate generation space.} Both arms: 16K training steps from the same 24K-step checkpoint, $256^2$, EMA weights. ADE20K: reduced-resolution diagnostic on the representation $Z_v$ with linear SGD / linear AdamW + ZCA / MLP readouts (Appendix~\ref{app:protocols}). Rank: participation-ratio effective rank of $Z_v$ on COCO patches. Reconstruction: ImageNet 5K, PSNR / FD-DINOv2. (b) Diagnostics: ImageNet validation-split probes on the pooled $g_v$ (40/10 images per class; $k$-NN at $k=20$, linear, 2048-unit MLP); ARO attribution / relation; linear ADE20K at the backbone and representation taps (a second run of the linear protocol); cosine between the clean-state estimate and the generation target at $\gamma=4$ / 5; mean training denoising loss over steps 8K--11K.}
\label{tab:endpoint}
\setlength{\tabcolsep}{3pt}
(a)\hfill\resizebox{.95\linewidth}{!}{%
\begin{tabular}{@{}lrrrrrrrrr@{}}\toprule
&\multicolumn{3}{c}{Global understanding}&\multicolumn{2}{c}{Dense representation}&I$\to$T&\multicolumn{3}{c}{Image generation}\\\cmidrule(lr){2-4}\cmidrule(lr){5-6}\cmidrule(lr){7-7}\cmidrule(lr){8-10}
Generated in & ZS & mR & SC & ADE lin / ZCA / MLP & Rank & SPICE & GenEval & DPG & Reconstruction\\\midrule
24K-step initialization &38.07 &65.88 &-- &17.0 / -- / -- &28.4 &-- &.372 &67.34 &--\\
\corerow \textbf{Shared} (\model{}) &44.65 &69.15 &79.8 &15.4 / 22.9 / 26.0 &21.8 &12.01 &\textbf{.428} &\textbf{71.30} &23.38 / 78.8\\
Separate &44.66 &69.42 &79.6 &\textbf{21.9} / \textbf{25.4} / 26.6 &30.0 &12.13 &.362 &65.91 &\textbf{25.03} / \textbf{57.0}\\\bottomrule
\end{tabular}}

\vspace{4pt}(b)\hfill\resizebox{.95\linewidth}{!}{%
\begin{tabular}{@{}lccccc@{}}\toprule
Generated in & ImageNet $k$-NN / lin / MLP & ARO & ADE tap study $h$ / $Z_v$ & $\cos(\widehat Z_0,Z_0)$ & Denoising loss\\\midrule
24K-step initialization &49.10 / 49.27 / 58.45 &-- &26.0 / 17.0 &.61 / .39 &--\\
\corerow Shared (\model{}) &54.94 / 55.47 / 63.97 &67.2 / 49.0 &26.6 / 15.5 &.56 / .34 &0.0713\\
Separate &54.79 / 55.51 / 63.43 &67.3 / 48.8 &28.1 / 21.9 &.48 / .28 &0.0809\\\bottomrule
\end{tabular}}
\end{table}

Generating in the understanding representation makes generation more compositional (Table~\ref{tab:endpoint}). The shared arm is ahead by .066 GenEval and 5.4 DPG (prompt-bootstrap 95\% intervals $[.043, .088]$ and $[4.40, 6.43]$; 10{,}000 paired resamples over prompts, stratified by GenEval task, none of which reverses the sign; this measures prompt sampling only, not training seeds; across three sampling seeds GenEval is $.4272\pm.0029$ against $.3616\pm.0024$, a gap of $.066\pm.004$), with the largest GenEval gains on two objects (50.3 vs.\ 36.1), counting (30.0 vs.\ 20.0), and color attribution (19.5 vs.\ 12.0)---the tasks that need text--image binding---and a lead in every DPG category (entity 80.2 vs.\ 76.2, other 66.0 vs.\ 60.8). Global understanding does not care where generation's gradient enters: both arms gain the same over the initialization (ZS $+6.6$, retrieval $+3.3$ and $+3.5$, ImageNet $k$-NN $+5.8$ and $+5.7$) and tie within 0.3 on SugarCrepe, ARO, and captioning. The separate space learns the better reconstruction code (+1.65~dB, FD-DINOv2 57.0 vs.\ 78.8), but that code is less predictable from noise: at $\gamma=4$ / 5 the clean-state estimate's cosine to its target is .48 / .28 against .56 / .34, and the mean denoising loss over training steps 8K--11K is 13\% higher (0.0809 vs.\ 0.0713; both targets are normalized, but they differ, so the loss comparison is only indicative). Whether the shared target is also of lower rank depends on the measure: by participation ratio it is (21.8 against 32.8 for the separate arm's target $\operatorname{LN}(P^{\rm gen}_vh_v)$), but 99\% of the separate target's variance lies in 39 components against 91 for the shared one. ``Lower rank is easier to denoise'' therefore does not simply explain the gap; a rank-matched separate space, or one with its own alignment loss, is the clean test. \nrev{Neither control varies rank alone, so rank-related effects are not ruled out.} The price of sharing is linear dense accessibility: the shared representation's effective rank falls below the initialization's (21.8 vs.\ 28.4; separate 30.0) and its linear-SGD ADE20K mIoU with it (15.4 vs.\ 21.9), a gap that whitening narrows to 2.5 points and an MLP nearly closes (26.0 vs.\ 26.6)---the same readout effect as in Section~\ref{sec:analysis}. The tap study places the cost at the representation: from the initialization, the linear representation tap falls from 17.0 to 15.5 in the shared arm and rises to 21.9 in the separate one, while the backbone tap rises in both (26.0 to 26.6 and 28.1). The pair is a controlled 16K-step continuation with one seed, not a full-length run.

\paragraph{A second projection with its own alignment loss, and training from scratch.} Table~\ref{tab:endpointladder} aligns the second projection on two training seeds (a) and trains both two-head designs from scratch (b). The aligned projection generates nearly as well as the single representation, never better (GenEval .423 and .411 vs.\ .427 and .428; from scratch at 48K steps .397 vs.\ .411), and \nrev{is largely linearly predictable from the understanding head ($R^2$ .89 / .95; .29 / .14 without alignment, and at most .40 in the continuation), which does not by itself make the two spaces equivalent}. On a second training seed from scratch, the aligned design is again behind at 24K steps (GenEval .328 vs.\ .369), and at 48K steps its zero-shot accuracy is 41.8\% against 43.4\%. The unaligned space stays behind on a second seed (GenEval $-.079$) and from scratch (.284 vs.\ .411 at 48K steps). The generation space is thus better when understanding shapes it together with generation; a second space shaped this way \nrev{brings no gain over the shared representation in these runs}, so one representation suffices.

\begin{table}[h]\centering\small
\caption{\textbf{A generation space that understanding shapes.} (a) 16K-step continuations of the 24K-step checkpoint \polish{(last three rows from a matched rerun)}. (b) From scratch, 24K steps: understanding reads $Z_U$, generation runs in $Z_G$; $R^2$: ridge regression between their patch tokens ($Z_U\!\to\!Z_G$ / $Z_G\!\to\!Z_U$). FID: COCO-30K.}
\label{tab:endpointladder}
\setlength{\tabcolsep}{4pt}
\begin{tabular}{@{}lrrrrrrr@{}}\toprule
(a) Generation space & ZS & mR & GenEval & DPG & FID & PSNR & ADE lin.\\\midrule
\corerow Shared (\model{}) &44.65 &69.15 &.428 &71.30 &10.75 &23.38 &15.4\\
Separate &44.66 &69.42 &.362 &65.91 &15.24 &25.03 &21.9\\\midrule
\corerow Shared (\model{}), \polish{matched rerun} &44.95 &68.84 &.427 &71.06 &10.99 &23.36 &15.5\\
Separate, aligned (extra understanding head) &44.65 &68.96 &.423 &70.13 &10.96 &23.20 &19.5\\
\quad second training seed &44.88 &68.56 &.411 &69.99 &11.04 &23.20 &19.4\\\bottomrule
\end{tabular}

\vspace{4pt}
\begin{tabular}{@{}lrrrrrr@{}}\toprule
(b) From scratch, 24K steps & ZS & mR & GenEval & DPG & FID & $R^2$\\\midrule
\corerow \model{} (one head) &38.07 &65.88 &.372 &67.34 &16.90 &--\\
Aligned $Z_G$ + understanding head $Z_U$ &36.34 &64.82 &.341 &66.03 &18.98 &.89 / .95\\
Separate $Z_G$ (generation only) &37.22 &65.72 &.226 &53.23 &40.75 &.29 / .14\\\bottomrule
\end{tabular}
\end{table}

\subsection{What does fine-tuning change?}
\label{app:finetune}
Table~\ref{tab:ftfull} compares the pretrained model with three fine-tunes on MiniT2I's 120K synthetic mixture (Appendix~\ref{app:implementation}): all parameters for MiniT2I-B/16's sample budget; the same with both encoders frozen, so that only the denoisers and their null codes train; and the recipe of the main tables, which trains only the image denoiser after a $512^2$ continuation on CC12M-R.

\begin{table}[h]\centering
\caption{\textbf{Fine-tuning on the 120K synthetic mixture.} FID: development-selected recipe; CAPTURE: DetailCaps. $^=$Encoders identical to the pretrained model. $^\sharp$Understanding and captioning read at the model's native $512^2$ input; the second stage leaves every parameter they read untouched. $^\natural$Same $512^2$ first stage, then all parameters for 8K steps with half of each batch from CC12M-R; read at $512^2$. $^\dagger$Second number: own-score selection among the four GenEval samples (Appendix~\ref{app:ownselect}), from a regenerated set whose four-sample average is .734.}
\label{tab:ftfull}
\setlength{\tabcolsep}{2.6pt}
\resizebox{\linewidth}{!}{%
\begin{tabular}{@{}lrrrrrrrrrr@{}}\toprule
&\multicolumn{4}{c}{Understanding}&\multicolumn{2}{c}{Image $\to$ text}&\multicolumn{4}{c}{Text $\to$ image}\\\cmidrule(lr){2-5}\cmidrule(lr){6-7}\cmidrule(lr){8-11}
Model & ZS & mR & LP $h$ & SC & SPICE & CAPTURE & GenEval & DPG & COCO FID$\downarrow$ & CC12M FID$\downarrow$\\\midrule
\corerow \textbf{\model{} + Gram/iBOT}, pretrained &50.90 &70.88 &73.62 &80.0 &13.50 &44.15 &.448 &73.49 &10.20 &4.37\\
\quad all parameters, 5K steps, $256^2$ &33.02 &54.52 &67.68 &64.2 &9.14 &34.52 &.732 / .764$^\dagger$ &79.25 &28.33 &13.28\\
\quad denoisers only, 5K steps, $256^2$ &50.90$^=$ &70.88$^=$ &73.62$^=$ &80.0$^=$ &10.86 &34.06 &.696 &78.10 &29.40 &13.24\\
\corerow \quad final: image denoiser only, $512^2{}^\sharp$ &51.49 &72.02 &74.19 &80.6 &13.91 &44.30 &\nrev{.806} &\nrev{79.29} &\nrev{31.81} &\nrev{15.27}\\
\quad joint: all parameters, 50\% CC12M-R replay, $512^2{}^\natural$ &45.60 &67.99 &73.70 &73.7 &13.03 &43.87 &.748 &78.26 &25.13 &5.22\\\bottomrule
\end{tabular}}
\end{table}

\begin{figure}[h]\centering
\includegraphics[width=\linewidth]{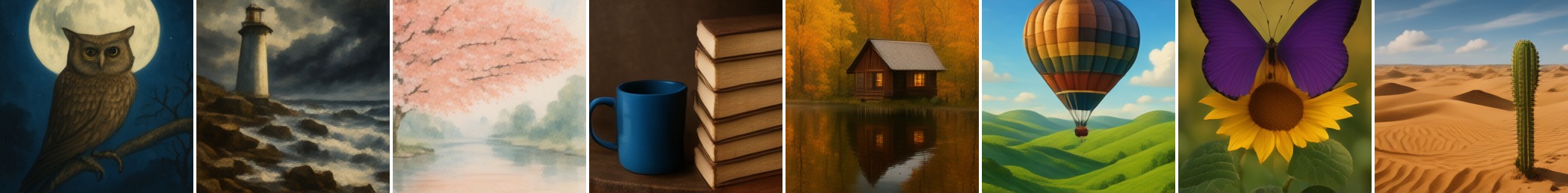}
\caption{\textbf{The all-parameter 5K fine-tune on \edit{eight prompts of Figure~\ref{fig:allseeds} (seven shared with Figure~\ref{fig:t2ibest})}.} $256^2$, 64 steps with CFG rising linearly from 1 to 5, churn 1; one of four seeds chosen by the model's own image--text score. This model reaches GenEval .732 but loses most of its understanding (Table~\ref{tab:ftfull}).}
\label{fig:ftsamples}
\end{figure}

Full fine-tuning buys prompt following (GenEval .448 $\to$ .732, DPG 73.49 $\to$ 79.25) at the cost of every other use\nrev{: zero-shot accuracy falls to 33.0\%, retrieval to 54.5, SPICE to 9.14, and DetailCaps CAPTURE to 34.52, while COCO-30K and CC12M-50K FIDs rise to 28.3 and 13.3 as generations adopt the mixture's style} (Table~\ref{tab:ftfull}). Most of the understanding loss is over-fitting to the small mixture rather than a conflict between the objectives: an all-parameter fine-tune at $512^2$ that replays CC12M-R for half of each batch loses 5.9 zero-shot points \nrev{(51.49 $\to$ 45.60)} instead of 17.9. Freezing both encoders keeps understanding exactly and still reaches 87\% of the GenEval gain and 80\% of the DPG gain, so prompt following does not require moving the shared representation; captions still drift \nrev{(SPICE 10.86)}, because the text denoiser now trains on the mixture's prompts alone, and both FIDs rise as much as under full fine-tuning (samples: Figure~\ref{fig:ftsamples}). \polish{The main-table recipe therefore trains the image denoiser only: understanding and captioning are exactly those of its $512^2$ first stage, the representation stays bit-identical, and it reaches the highest GenEval of the table (\nrev{.806}).} \nrev{Training all parameters jointly with CC12M-R replay reaches GenEval .748 and the best CC12M-50K FID, 5.22, but zero-shot accuracy falls to 45.60 and retrieval to 67.99.} \emph{Overlap with the benchmarks.} Of the mixture's 119{,}239 captions, 5 GenEval prompts appear verbatim (all single-object, all from the BLIP3o part), and 22 GenEval prompts (4.0\%; 11 if word order must match) have a training caption with the same objects, counts, colors, and relation; no DPG-Bench prompt appears (largest token Jaccard 0.56). The mixture is, however, close to GenEval in format: 3{,}256 captions (2.7\%) are complete GenEval-template captions over GenEval's vocabulary (7{,}227 with any object words), and 35.5\% have at most 12 tokens, GenEval's length range. GenEval gains after fine-tuning therefore partly reflect in-format training; DPG gains (73.49 $\to$ \nrev{79.29} for the main-table model) are the cleaner evidence, and MiniT2I fine-tunes on the same mixture.

\subsection{Other generation spaces and components}
\label{app:components}
Table~\ref{tab:spaces} trains a new image denoiser (131M image--text presentations) in frozen generation spaces, among them the spaces of RAE-style generation \citep{zheng2025rae}, changing only the space or the text encoder; absolute values are thus below the fully trained model's. \polish{Among the spaces trained on the same data, the one shaped by understanding and generation is the best on all three metrics, and it has the lowest FID of all rows.} Pretrained spaces (DINOv2-B, on 142M images; SigLIP 2, on about 10B pairs; the FLUX.2 VAE) are ahead in prompt following at this schedule; SD-VAE and FLUX.1 latents reconstruct better (25.0 and 31.2~dB, against 24.0) and generate worse. FLAN-T5-Large in place of our text encoder gives no gain. \polish{Table~\ref{tab:decoders} replaces the denoiser's pixel head by a separately trained decoder: on the frozen-contrastive control it improves FID but not prompt following, and on \model{}'s representation the denoiser's own head is better on both.}

\begin{table}[h]\centering\small
\caption{\polish{\textbf{A separately trained decoder in place of the denoiser's pixel head.} Decoder: ViT-B trained on the same data with L1, LPIPS, and adversarial losses on RAE's schedule (discriminator on frozen DINOv2-B features; 20.5M image presentations). Each pair is measured in one pass; FID: COCO-30K.}}
\label{tab:decoders}
\setlength{\tabcolsep}{5pt}
\begin{tabular}{@{}llrrr@{}}\toprule
Representation & Pixels from & GenEval & DPG & FID$\downarrow$\\\midrule
Frozen contrastive (fully trained control) & denoiser's own head & .404 & 71.27 & 12.55\\
 & separate decoder & .386 & 70.88 & 11.02\\\midrule
\corerow \model{} + Gram/iBOT, frozen & denoiser's own head & .457 & 74.18 & 10.02\\
 & separate decoder & .430 & 74.19 & 10.90\\\bottomrule
\end{tabular}
\end{table}

\begin{table}[h]\centering\small
\caption{\textbf{The same denoiser in other generation spaces and with another text encoder.} 16K steps from a new initialization; all spaces frozen. SigLIP 2 and DINOv2-B use the released RAE decoders; our spaces need no separate decoder. VAE latents are standardized per channel, and the noise schedule is shifted for the FLUX latents (FLUX.1 +1.386, FLUX.2 +2.079). FID: COCO-30K.}
\label{tab:spaces}
\setlength{\tabcolsep}{4pt}
\begin{tabular}{@{}llrrr@{}}\toprule
Generation space & Text condition & GenEval & DPG & FID$\downarrow$\\\midrule
\multicolumn{5}{@{}l}{\emph{Trained on CC12M}}\\
Pixels (no pretrained component) & ours &.200 &50.2 &39.77\\
Contrastive only (12M pairs) & ours &.343 &67.8 &14.3\\
Contrastive only + Gram/iBOT (12M pairs) & ours &.322 &68.4 &13.88\\
\corerow \model{} + Gram/iBOT (12M pairs) & ours &.355 &69.6 &11.68\\
\corerow \model{} + Gram/iBOT (12M pairs) & FLAN-T5-Large &.346 &69.1 &11.77\\\midrule
\multicolumn{5}{@{}l}{\emph{Pretrained on other data}}\\
SigLIP 2 So400M + RAE decoder ($\sim$10B pairs) & ours &.412 &74.3 &13.45\\
DINOv2-B + RAE decoder & ours &.456 &71.5 &13.80\\
SD-VAE latent & ours &.319 &62.5 &20.09\\
FLUX.1 VAE latent & ours &.276 &59.6 &24.19\\
FLUX.2 VAE latent & ours &.415 &72.1 &13.20\\\bottomrule
\end{tabular}
\end{table}

On the frozen \model{} encoder, an autoregressive text decoder of the same size is the better COCO captioner (SPICE 15.8 vs.\ 13.5 for the text denoiser), and it does better on \model{}'s encoder than on the contrastive-only one (15.80 vs.\ 15.42). We generate text by diffusion for another reason: it makes $Z_t$ a space in which text is generated, as for images, rather than a representation with a generator added on top. Tokens are produced in parallel, and the distilled student needs two denoiser evaluations per caption (SPICE 12.05; 13.66 with nine, against 13.50 for the 128-evaluation teacher; Appendix~\ref{app:distillation}). Continuing the 24K-step checkpoint for 16K steps with only the token condition $Z$ gives GenEval .400 and COCO FID 11.09, only the pooled $g$ .373 and 11.51, and both .427 and 10.99; zero-shot accuracy barely changes (44.9, 44.5, 45.0).

\section{Extended capability results}
\label{app:extended}
This appendix expands the three main tables for the models and controls they report. Every row uses the protocols of Appendix~\ref{app:protocols}; the frozen-contrastive control shares the contrastive-only encoders and is listed only where it has its own generator.

\subsection{Global and dense probes}
\begin{table}[!htbp]\centering\small
\caption{\polish{\textbf{ImageNet-1k linear probes and ADE20K probes on frozen features.} ImageNet: full-train LARS, last-epoch top-1 at three taps. ADE20K: full-resolution mIoU at learning rate $10^{-3}$ with linear, MLP, and attentive probes (Appendix~\ref{app:protocols}); MLP and attentive: mean of three probe seeds, attentive with standard deviation.}}
\label{tab:dense}\label{tab:ladderfull}
\setlength{\tabcolsep}{3.5pt}
\resizebox{\linewidth}{!}{%
\begin{tabular}{@{}lrrrrrrrrr@{}}\toprule
&\multicolumn{3}{c}{ImageNet-1k linear probe}&\multicolumn{3}{c}{ADE20K on $Z$}&\multicolumn{3}{c}{ADE20K on $h$}\\\cmidrule(lr){2-4}\cmidrule(lr){5-7}\cmidrule(lr){8-10}
Model & $h$ & $Z$ & $g$ & Linear & MLP & Attentive & Linear & MLP & Attentive\\\midrule
Contrastive only &72.51 &61.78 &72.34 &30.15 &35.61 &$41.20\pm0.08$ &37.41 &40.05 &$43.30\pm0.06$\\
Contrastive only + Gram/iBOT &74.06 &63.91 &73.98 &33.13 &38.30 &$43.36\pm0.06$ &40.00 &42.75 &$45.14\pm0.12$\\
\corerow \model{} &72.82 &63.70 &73.48 &24.69 &34.12 &$42.38\pm0.47$ &36.48 &40.16 &$43.57\pm0.21$\\
\corerow \model{} + Gram/iBOT &73.62 &64.78 &74.31 &25.45 &35.61 &$43.54\pm0.23$ &38.27 &42.00 &$44.86\pm0.65$\\\bottomrule
\end{tabular}}
\end{table}

Linear probes on $h$ and $g$ of all aligned models lie within 2 points of one another, whereas the dense representation separates them: joint learning lowers ADE20K mIoU on $Z$ by 5.5 points (24.69 vs.\ 30.15). The no-contrastive-loss control keeps 71.72\% ImageNet accuracy on $h$ \polish{(63.39\% on $Z$, 70.22\% on $g$; ADE20K linear probe 35.80 on $h$ and 23.88 on $Z$)} with near-chance zero-shot transfer (Table~\ref{tab:understanding}): its information is retained but not organized for language.

On $Z$, \model{}'s linear-probe deficit to the contrastive-only control (5.5 mIoU) shrinks under an MLP and reverses under an attentive probe ($+1.18$, paired 95\% interval $[0.53, 1.89]$; linear $-5.45$ $[-5.77, -4.75]$, MLP $-1.49$ $[-1.87, -0.92]$; Table~\ref{tab:ladderfull}). On $h$, \model{} and the contrastive-only control differ by under one point. With the attentive probe, \model{} + Gram/iBOT and its matched control are level on $Z$ (43.5 vs.\ 43.4). Stopping the generation gradient through the noised representation raises the reduced-resolution linear probe (15.5 to 22.3) but lowers GenEval (.427 to .404) and reconstruction (22.2 to 19.5 dB on 256 images).

\begin{table}[!htbp]\centering\small
\caption{\textbf{Monocular depth on NYU Depth v2 \citep{silberman2012nyu} from frozen $Z$.} Probes of Table~\ref{tab:ladderfull} over 256 depth bins; linear: three probe seeds, others: one. RMSE in metres; $\delta_1$ in \%. Difference: \model{} minus control in $\delta_1$, paired 95\% interval over test images.}
\label{tab:depth}
\setlength{\tabcolsep}{5pt}
\begin{tabular}{@{}lrrrrrr@{}}\toprule
&\multicolumn{2}{c}{Linear}&\multicolumn{2}{c}{MLP}&\multicolumn{2}{c}{Attentive probe}\\\cmidrule(lr){2-3}\cmidrule(lr){4-5}\cmidrule(lr){6-7}
&RMSE$\downarrow$&$\delta_1\uparrow$&RMSE$\downarrow$&$\delta_1\uparrow$&RMSE$\downarrow$&$\delta_1\uparrow$\\\midrule
Contrastive only &0.810 &56.95 &0.806 &57.79 &0.608 &72.73\\
\corerow \model{} &0.787 &58.53 &0.758 &61.04 &0.590 &74.18\\
Difference in $\delta_1$ &\multicolumn{2}{r}{$+1.58$ $[1.01, 2.14]$} &\multicolumn{2}{r}{$+3.25$ $[2.64, 3.90]$} &\multicolumn{2}{r}{$+1.45$ $[0.54, 2.38]$}\\\bottomrule
\end{tabular}
\end{table}

On monocular depth (795 training, 654 test images), \model{}'s $Z$ is ahead of the contrastive-only control \polish{across all three probes} (Table~\ref{tab:depth}). The ADE20K linear-probe deficit is thus not a general loss of dense information in $Z$.

\subsection{Cross-modal and compositional matching}
\begin{table}[!htbp]\centering\small
\caption{\textbf{COCO 5K retrieval, SugarCrepe, and ARO.} Retrieval R@1/5/10 in each direction. SugarCrepe (\%): R/S/A are replace/swap/add. ARO (\%): VG-Attribution and VG-Relation, chance 50; VG-Relation is at chance for every model and is shown only for completeness.}
\label{tab:crossmodal}
\setlength{\tabcolsep}{2.4pt}
\resizebox{\linewidth}{!}{%
\begin{tabular}{@{}lllrrrrrrrrrr@{}}\toprule
&\multicolumn{2}{c}{COCO retrieval}&\multicolumn{8}{c}{SugarCrepe}&\multicolumn{2}{c}{ARO}\\\cmidrule(lr){2-3}\cmidrule(lr){4-11}\cmidrule(lr){12-13}
Model & I$\to$T & T$\to$I & R-obj & R-att & R-rel & S-obj & S-att & A-obj & A-att & Mean & Attr. & Rel.\\\midrule
Contrastive only &60.16/83.10/89.52 &41.75/67.21/76.78 &92.1 &83.6 &70.2 &69.0 &80.2 &81.3 &79.5 &79.4 &66.4 &48.7\\
Contrastive only + Gram/iBOT &61.90/83.46/90.12 &43.65/68.35/77.71 &93.2 &83.2 &71.6 &69.8 &77.2 &82.6 &79.6 &79.6 &66.5 &49.3\\
No contrastive loss &0.40/2.06/3.72 &0.19/0.95/1.73 &57.8 &56.0 &52.9 &50.2 &47.9 &58.0 &54.0 &53.8 &50.6 &49.2\\
\corerow \model{} &60.78/83.00/89.64 &43.12/68.19/77.32 &93.2 &85.5 &71.6 &70.2 &81.2 &82.7 &79.5 &80.6 &63.6 &47.8\\
\corerow \model{} + Gram/iBOT &61.08/83.80/89.64 &44.06/68.91/77.80 &92.9 &84.5 &72.3 &69.0 &79.0 &83.0 &79.5 &80.0 &65.1 &49.1\\\bottomrule
\end{tabular}}\end{table}

\model{} matches the contrastive-only control on retrieval and matches or exceeds it on every SugarCrepe split. ARO is less favorable: attribute accuracy is 2.8 points lower for \model{} (63.6 vs.\ 66.4), and relation accuracy stays at chance for every model, including the contrastive-only control. At equal steps, \model{} is ahead of the contrastive-only control in zero-shot accuracy at all six checkpoints from 24K to 125K steps ($+0.6$ to $+1.3$ points), in few-shot classification on $g$ ($+1.9$ / $+1.7$ / $+1.5$ at 1 / 5 / 10 shots), on a second training seed at 24K steps (37.8 vs.\ 36.3), and in image-to-text retrieval of DOCCI's long descriptions (R@1 83.4 vs.\ 81.6; $+1.8$, 95\% interval $[+0.7, +2.8]$).

\subsection{Text-only readouts of the text representation}
\label{app:textonly}
\begin{table}[!htbp]\centering\small
\caption{\textbf{Text-only readouts of the frozen text encoder} (pooled representation; no image, no training). Caption-to-caption R@1: each test caption queries all others (COCO 25K, Flickr30k 5K captions), and a hit describes the same image. SICK-R: Spearman correlation ($\times100$) of cosine similarity with human relatedness (9{,}927 pairs).}
\label{tab:textonly}
\setlength{\tabcolsep}{5pt}
\begin{tabular}{@{}lrrrr@{}}\toprule
Text encoder & Pairs & COCO R@1 & Flickr30k R@1 & SICK-R\\\midrule
CLIP ViT-B/32 &400M &\oss{35.02} &\oss{61.26} &\oss{67.9}\\
CLIP ViT-B/16 &400M &\oss{34.95} &\oss{60.86} &\oss{69.1}\\
\corerow \model{} &12M &35.54 &62.48 &69.0\\
\corerow \model{} + Gram/iBOT &12M &35.66 &62.80 &69.1\\\bottomrule
\end{tabular}
\end{table}

Trained on 12M pairs, \model{}'s text encoder is level with CLIP's (400M) on sentence relatedness and above it on caption-to-caption retrieval (Table~\ref{tab:textonly}). The representation also generates text: from a caption's $Z_t$ alone, the text denoiser regenerates the sentence. On 5{,}000 COCO test captions (64-step recipe), BLEU-4 to the input is 62, 19\% are reproduced exactly, and 95\% of regenerations retrieve their source at R@1 among 25K captions.

\subsection{Prompt-following profiles}
\label{app:t2iresults}
\begin{table}[!htbp]\centering\small
\caption{\textbf{GenEval and DPG-Bench profiles of our models} (fixed 128-step recipe; four images per prompt). GenEval tasks in percent, overall as a fraction. External profiles are in Table~\ref{tab:promptext}.}
\label{tab:promptours}
\setlength{\tabcolsep}{2.6pt}
\resizebox{\linewidth}{!}{%
\begin{tabular}{@{}lrrrrrrr|rrrrrr@{}}\toprule
&\multicolumn{7}{c|}{GenEval}&\multicolumn{6}{c}{DPG-Bench}\\
Model & Single & Two & Count & Color & Position & Attr. & Overall & Global & Entity & Attribute & Relation & Other & Overall\\\midrule
No contrastive loss &62.5 &6.3 &11.6 &32.7 &0.5 &2.0 &.193 &72.95 &60.37 &69.70 &79.30 &37.2 &45.31\\
Frozen contrastive $\to$ generation &83.8 &47.7 &28.8 &49.2 &7.8 &22.8 &.400 &77.20 &79.14 &81.65 &89.21 &63.2 &70.12\\
\corerow \model{} &88.4 &54.5 &32.8 &67.0 &8.3 &25.8 &.461 &79.94 &82.91 &82.85 &90.29 &68.4 &74.43\\
\corerow \model{} + Gram/iBOT &91.6 &58.3 &30.0 &60.9 &8.0 &20.3 &.448 &79.64 &82.00 &81.85 &90.37 &68.4 &73.49\\
\corerow \quad all parameters fine-tuned, 5K steps, $256^2$ &99.1 &87.1 &58.1 &89.9 &41.5 &63.3 &.732 &80.85 &86.87 &85.44 &91.53 &74.8 &79.25\\
\corerow \quad final fine-tune, $512^2$ &99.7 &\nrev{90.4} &\nrev{78.1} &\nrev{89.6} &\nrev{58.3} &\nrev{67.5} &\nrev{.806} &\nrev{78.42} &\nrev{86.59} &\nrev{85.28} &\nrev{90.83} &\nrev{69.2} &\nrev{79.29}\\\bottomrule
\end{tabular}}
\end{table}

Without fine-tuning, prompt following fails mostly on relative position and attribute binding; fine-tuning on the synthetic mixture raises these most (final fine-tuned model: position 8.0 $\to$ \nrev{58.3}, attributes 20.3 $\to$ \nrev{67.5}), while single objects and colors approach their ceiling. DPG moves less, mainly on entity and attribute questions.

\subsection{Reconstruction}
\begin{table}[!htbp]\centering\small
\caption{\textbf{Low-noise reconstruction on 5K ImageNet validation images} (full-condition route; VAEs decode their posterior mode). Two independent draws of real images give rFID 3.058 at 50K (7.526 at 20K); reconstructions are paired with their own originals and can fall below this reference, and the 5K values are dominated by finite-sample bias.}
\label{tab:recon}
\setlength{\tabcolsep}{4pt}
\begin{tabular}{@{}lrrrrr@{}}\toprule
Model & PSNR & SSIM & LPIPS$\downarrow$ & rFID 5K$\downarrow$ & rFID 50K$\downarrow$\\\midrule
SD-VAE kl-f8 ft-EMA \citep{rombach2022ldm} &\oss{25.04} &\oss{0.711} &\oss{0.141} &\oss{3.05} &--\\
SDXL-VAE \citep{podell2023sdxl} &\oss{26.10} &\oss{0.749} &\oss{0.128} &\oss{2.95} &--\\\midrule
No contrastive loss &26.91 &0.784 &0.141 &4.76 &--\\
Frozen contrastive $\to$ generation &19.26 &0.492 &0.306 &11.58 &--\\
\corerow \model{} &24.09 &0.669 &0.206 &6.43 &2.31\\
\corerow \model{} + Gram/iBOT &24.03 &0.665 &0.203 &6.24 &2.20\\\bottomrule
\end{tabular}\end{table}

The encoded representation reconstructs within 1--2~dB of the two VAEs, which are deterministic and trained for reconstruction alone; at 50K images \model{}'s rFID falls to 2.31, so most of its 5K value (6.43) is finite-sample bias. Alignment costs fine detail (no contrastive loss: 26.91~dB), and a representation never shaped by generation reconstructs worst (19.26~dB).

\begin{figure}[!htbp]\centering
\includegraphics[width=.9\linewidth]{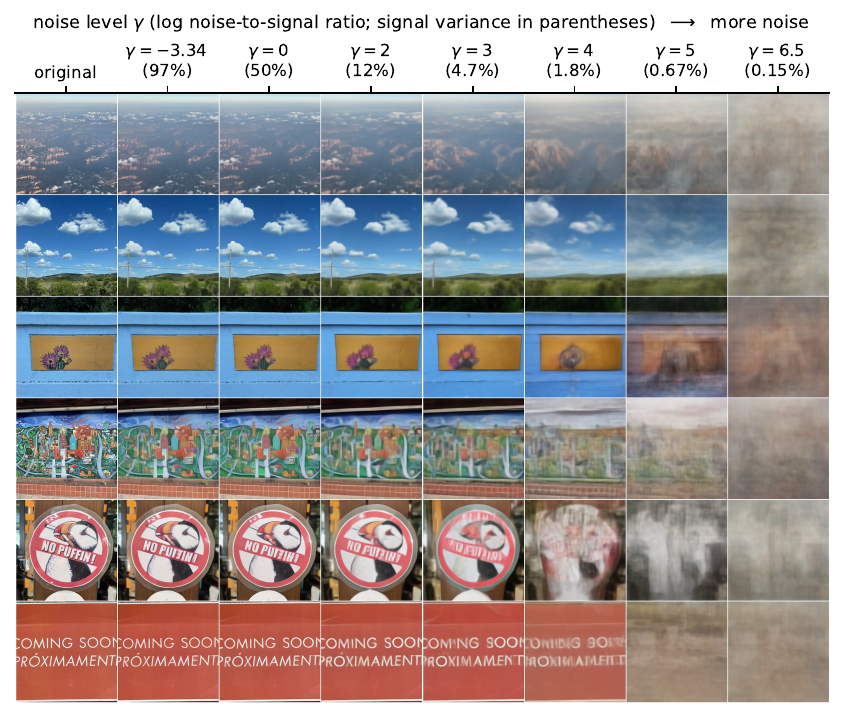}
\caption{\textbf{What the contextual representation retains as noise rises.} \model{}; each row is one person-free DOCCI image (CC BY 4.0): landscapes, a painting, a mural, and signs with benign text. Columns: the original, then one null-route denoiser call from the image's own representation corrupted at noise level $\gamma$ (top axis; signal variance $\operatorname{sigmoid}(-\gamma)$ in parentheses).}
\label{fig:gammapanel}
\end{figure}

\paragraph{What the representation retains as noise rises.} Figure~\ref{fig:gammapanel} decodes six images from their own contextual representation corrupted at increasing noise, with one noise draw per image and learned null conditions, so that the noised representation is the only image-side input. Global layout and color survive to $\gamma\approx3$ (under 5\% signal variance), legible text to $\gamma\approx2$, and fine lettering degrades already at the lowest noise level, consistent with the 24\,dB reconstruction level of Table~\ref{tab:recon}. Averaged over the six images, \model{} reaches 24.8/24.2/22.5/20.6/17.4/13.9/12.4\,dB at $\gamma=-3.34/0/2/3/4/5/6.5$; the no-contrastive-loss control is sharper at low noise (28.0\,dB at $\gamma=-3.34$) but indistinguishable from $\gamma=3$ on (20.9, 17.1, 14.3, 12.8\,dB): alignment costs fine detail, not coarse structure.

\subsection{COCO caption quality and hallucination}
\label{app:i2tresults}
\begin{table}[!htbp]\centering\small
\caption{\textbf{COCO Karpathy 5K captioning.} One DDPM-64 sample per image unless marked; CHAIR in percent. The human row scores one reference against the other four. Bottom: \model{} + Gram/iBOT choosing among four samples per image (main-protocol seeds) with its own image--text score; the length-matched rule picks the highest-scoring caption within two words of the seed-0 length when one exists.}
\label{tab:captionquality}\label{tab:ownselect5k}
\setlength{\tabcolsep}{4pt}
\begin{tabular}{@{}lrrrrrrr@{}}\toprule
Model & NFE & CIDEr-D & SPICE & CHAIR$_s\!\downarrow$ & CHAIR$_i\!\downarrow$ & RefCLIP-S & Words\\\midrule
Human reference &-- &89.28 &21.19 &3.5 &2.5 &.8175 &10.4\\
BLIP-base \citep{li2022blip} &-- &\oss{94.39} &\oss{17.53} &\oss{3.8} &\oss{3.2} &\oss{.8081} &\oss{6.2}\\\midrule
No contrastive loss &128 &8.66 &13.86 &18.8 &10.1 &.7414 &43.1\\
Frozen contrastive $\to$ generation &128 &3.63 &10.06 &29.8 &16.5 &.7358 &54.2\\
\corerow \model{} &128 &9.10 &13.53 &21.6 &11.2 &.7398 &42.6\\
\corerow \model{} + Gram/iBOT &128 &9.13 &13.50 &20.9 &11.2 &.7399 &43.8\\
\corerow \quad final fine-tune ($512^2$ representation) &128 &9.45 &13.91 &19.9 &10.4 &.7434 &43.4\\\midrule
\multicolumn{8}{@{}l}{\emph{Selection among four samples by the model's own readout (\model{} + Gram/iBOT)}}\\
Random of 4 &512 &9.70 &13.52 &-- &10.9 &.7396 &44.8\\
Own score, length-matched &512 &9.54 &\textbf{13.81} &-- &11.4 &.7423 &43.8\\
Own score, best of 4 &512 &7.43 &14.56 &-- &10.9 &.7458 &52.7\\\bottomrule
\end{tabular}\end{table}

CIDEr-D penalizes our long captions (43 words against about 10 for the references); scoring only the first sentence raises it to 14.80 for \model{} at nearly unchanged SPICE (13.44). The own-score selection rule was fixed in advance. At matched length it raises SPICE by 0.31 (paired 95\% interval $[0.20,0.42]$), CIDEr-D by 0.40 ($[0.18,0.64]$), and RefCLIP-S by .0024 ($[.0016,.0032]$) over the single sample, and beats random-of-four on SPICE by 0.29 ($[0.08,0.49]$); CHAIR$_i$ does not change significantly. Unconstrained selection prefers longer captions (+8.9 words), which raises SPICE by 1.06 but lowers CIDEr-D, so the length control is necessary. The rule costs four samples plus four caption encodings and one image encoding (about 40~ms on one H100, one percent of the sampling time).

\subsection{Detailed captioning}
\begin{table}[!htbp]\centering\small
\caption{\textbf{Detailed-caption benchmarks.} DetailCaps-4870: CAPTURE and its object/attribute/relation F1 (three references). DOCCI-5K: CAPTURE transferred to one human description. External rows: released captions rescored with the same evaluators \citep{dong2024capture}.}
\label{tab:detailed}
\setlength{\tabcolsep}{3pt}
\resizebox{\linewidth}{!}{%
\begin{tabular}{@{}lrrrrr@{}}\toprule
&\multicolumn{4}{c}{DetailCaps}&DOCCI\\\cmidrule(lr){2-5}\cmidrule(lr){6-6}
Model & CAPTURE & Object & Attribute & Relation & CAPTURE\\\midrule
ShareCaptioner-7B \citep{chen2023sharegpt4v} &\oss{59.80} &-- &-- &-- &--\\
CogVLM \citep{wang2023cogvlm} &\oss{60.06} &-- &-- &-- &--\\\midrule
No contrastive loss &44.43 &48.83 &41.79 &40.06 &45.84\\
Frozen contrastive $\to$ generation &43.73 &48.07 &41.71 &37.95 &44.46\\
\corerow \model{} &43.62 &48.01 &41.02 &39.17 &45.66\\
\corerow \model{} + Gram/iBOT &44.15 &48.56 &41.44 &39.88 &45.52\\
\corerow \quad final fine-tune ($512^2$ representation) &44.30 &48.83 &41.46 &40.06 &45.32\\\bottomrule
\end{tabular}}
\end{table}

On both benchmarks our models lie within 2 points of one another. They trail captioners built on 7B--17B pretrained language models by about 16 CAPTURE points on DetailCaps. For reference, the benchmark papers report a two-reference DetailCaps CAPTURE of \published{49.69} for LLaVA-1.5-7B \citep{liu2024llava15}. Parser truncation affects at most 2.1\% of our DetailCaps captions (5.5\% for the frozen-contrastive control, whose captions are longest).

\begin{table}[!htbp]\centering\small
\caption{\textbf{CompreCap} \citep{lu2025comprecap}: object coverage (\%), attribute and relation scores (0--5), and the unified score on its 560 images, with a Qwen2.5-7B-Instruct judge for every row. External rows: captions generated with the decoding settings of Table~\ref{tab:i2t}, or the captions released with the benchmark (GPT-4o, LLaVA-NeXT-34B).}
\label{tab:comprecap}
\setlength{\tabcolsep}{5pt}
\begin{tabular}{@{}lrrrr@{}}\toprule
Model & Object \% & Attribute & Relation & Unified\\\midrule
BLIP-base \citep{li2022blip} &\oss{29.82} &\oss{0.727} &\oss{1.209} &\oss{22.22}\\
CoCa-L/14 \citep{yu2022coca} &\oss{38.66} &\oss{0.929} &\oss{1.390} &\oss{27.29}\\
Florence-2-base, caption prompt \citep{xiao2024florence2} &\oss{39.13} &\oss{0.951} &\oss{1.404} &\oss{27.68}\\
Florence-2-base, detailed-caption prompt &\oss{69.85} &\oss{1.838} &\oss{2.022} &\oss{46.51}\\
LLaVA-NeXT-34B \citep{liu2024llavanext} &\oss{72.86} &\oss{2.059} &\oss{1.833} &\oss{47.29}\\
GPT-4o \citep{openai2024gpt4o} &\oss{72.81} &\oss{2.088} &\oss{1.974} &\oss{48.61}\\\midrule
No contrastive loss &46.27 &1.145 &1.167 &28.92\\
Frozen contrastive $\to$ generation &45.44 &0.951 &0.994 &25.97\\
\corerow \model{} &47.35 &1.159 &1.176 &29.36\\
\corerow \model{} + Gram/iBOT &45.71 &1.106 &1.149 &28.36\\
\corerow \quad final fine-tune ($512^2$ representation) &46.28 &1.134 &1.176 &28.92\\\bottomrule
\end{tabular}
\end{table}

On CompreCap (Table~\ref{tab:comprecap}), \model{} is above the frozen-contrastive control (unified 29.4 vs.\ 26.0; 95\% intervals $[28.2, 30.5]$ and $[25.0, 27.0]$) and above BLIP, CoCa, and Florence-2 with its caption prompt; captioners prompted or built for detailed descriptions are 17--19 points higher.

\subsection{Qualitative samples}
\begin{figure}[tbp]\centering
\includegraphics[width=\linewidth]{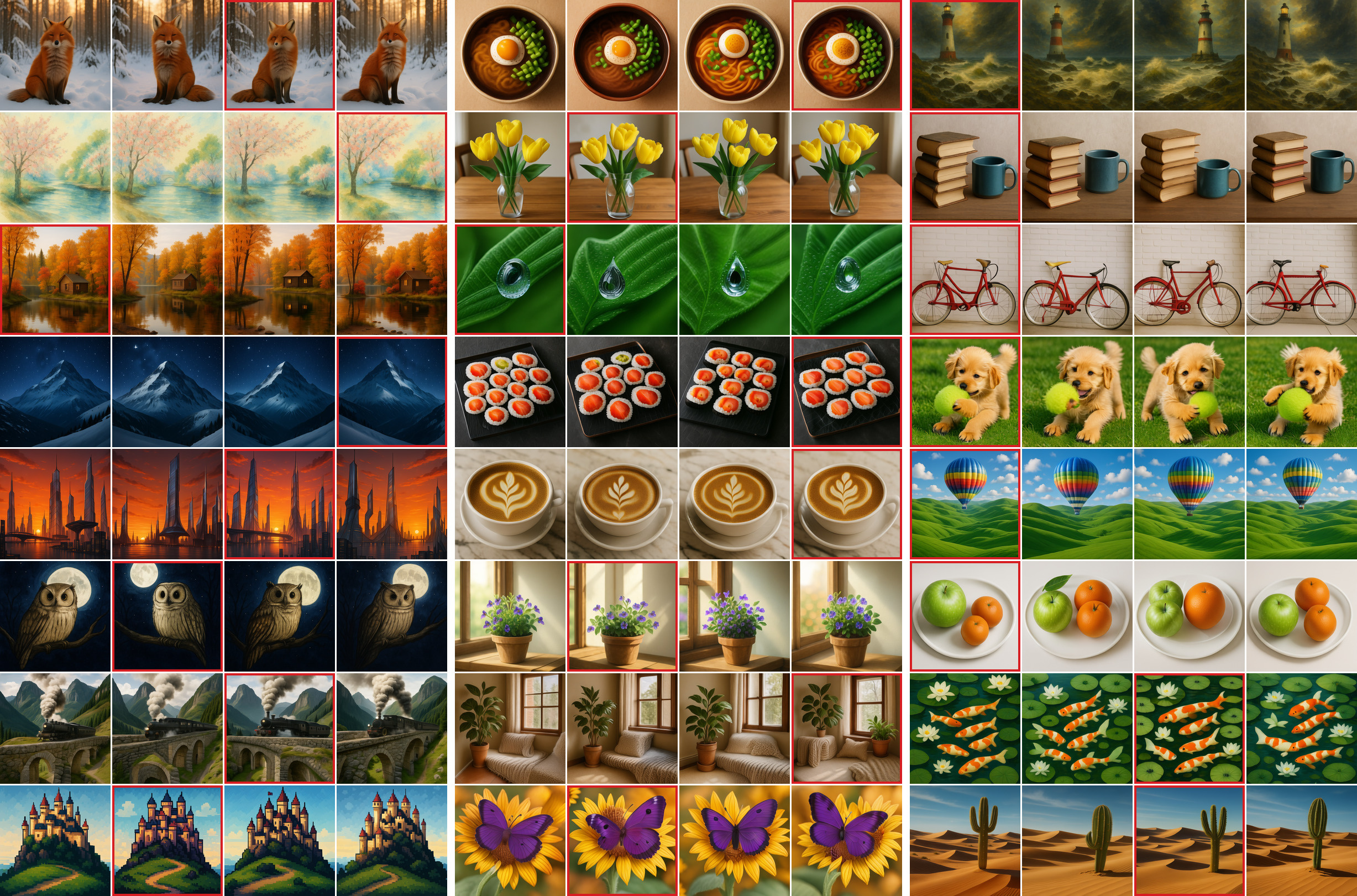}
\caption{\textbf{All samples, unselected.} Four seeds for each of 24 person-free prompts \edit{without rendered text} from the model and sampler of Figure~\ref{fig:t2ibest}; the seed chosen by the model's own image--text score is framed in red. Read left to right, top to bottom, three prompts per row: (1) A red fox sitting in a snowy forest at dawn; (2) A bowl of ramen with a soft-boiled egg and green onions, viewed from above; (3) An oil painting of a lighthouse on a rocky coast during a storm; (4) A watercolor painting of cherry blossoms along a quiet river; (5) Three yellow tulips in a clear glass vase on a wooden table; (6) A blue ceramic mug to the left of a stack of old books; (7) A small wooden cabin beside a lake surrounded by autumn trees, reflected in the water; (8) A macro photograph of a dewdrop on a green leaf; (9) A vintage red bicycle leaning against a white brick wall; (10) A snowy mountain peak under a starry night sky; (11) A plate of sushi arranged on a black slate board; (12) A golden retriever puppy playing with a tennis ball on the grass; (13) A futuristic city skyline at sunset with glass towers; (14) A cup of cappuccino with latte art on a marble table; (15) A hot air balloon floating over rolling green hills; (16) An owl perched on a branch at night under a full moon; \edit{(17) a potted plant with delicate small flowers featuring vibrant purple petals sits on a wooden windowsill [\dots] (a DPG-Bench prompt, shortened here);} (18) A green apple and two oranges on a white plate; (19) A steam locomotive crossing a stone bridge in the mountains; (20) A cozy reading nook with a window, a potted plant, and a knitted blanket; (21) A koi pond with orange and white fish among lily pads; (22) A pixel-art style castle on a hill; (23) A purple butterfly resting on a sunflower; (24) A desert landscape with sand dunes and a single cactus. Counts are not respected (prompt 5 shows more than three tulips in every seed), and the mug stands to the right of the books in every seed (prompt 6).}
\label{fig:allseeds}
\end{figure}

\begin{figure}[tbp]\noindent
\begin{minipage}[c]{.215\linewidth}\includegraphics[width=\linewidth]{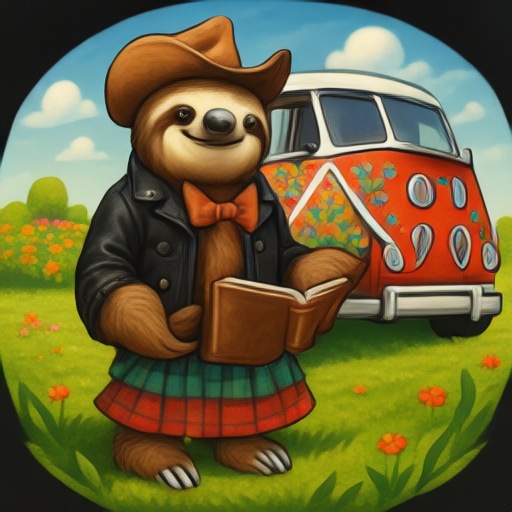}\end{minipage}\hspace{.01\linewidth}\begin{minipage}[c]{.265\linewidth}\raggedright\fontsize{5.6}{6.6}\selectfont\itshape A cheerful sloth, adorned with a black leather jacket and a brown cowboy hat, stands confidently on a patch of green grass. Its attire is completed with a red tartan kilt and a neatly tied bowtie. In one claw, it grasps a sturdy quarterstaff, while the other holds a large, ancient-looking book. Positioned a short distance behind the sloth, there is a gleaming Volkswagen van, its exterior decorated with vibrant flower patterns. The entire whimsical scene is captured through a wide-angle lens from a low vantage point, emphasizing the sloth's stature and the van's presence.\end{minipage}\hfill%
\begin{minipage}[c]{.215\linewidth}\includegraphics[width=\linewidth]{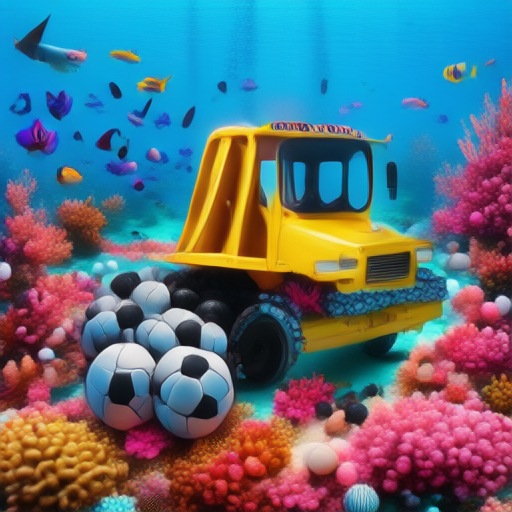}\end{minipage}\hspace{.01\linewidth}\begin{minipage}[c]{.265\linewidth}\raggedright\fontsize{5.6}{6.6}\selectfont\itshape A vibrant underwater scene where a yellow dump truck, brimming with black and white soccer balls, is whimsically placed among the colorful coral reef. The truck appears to be `scuba diving' in the clear blue waters, surrounded by schools of tropical fish and intricate sea life. The coral formations exhibit a multitude of hues, from bright pinks to deep purples, creating a playful contrast with the artificiality of the submerged vehicle.\end{minipage}\\[4pt]
\begin{minipage}[c]{.215\linewidth}\includegraphics[width=\linewidth]{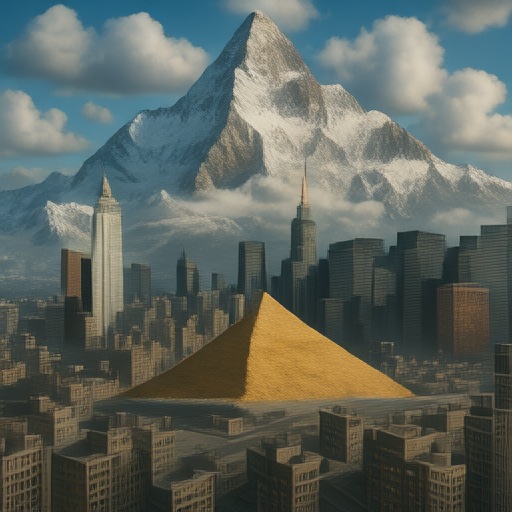}\end{minipage}\hspace{.01\linewidth}\begin{minipage}[c]{.265\linewidth}\raggedright\fontsize{5.6}{6.6}\selectfont\itshape A surreal landscape where the iconic skyline of downtown Manhattan, with its towering skyscrapers and bustling streets, is juxtaposed against the majestic backdrop of Mount Everest, its peak shrouded in snow and clouds. In the foreground, the ancient Great Pyramid of Giza stands solitary, its limestone blocks weathered to a golden hue, casting a long shadow in the direction of the urban architecture. This unusual combination of human-made wonders spans different continents and eras, creating a striking and imaginative scene.\end{minipage}\hfill%
\begin{minipage}[c]{.215\linewidth}\includegraphics[width=\linewidth]{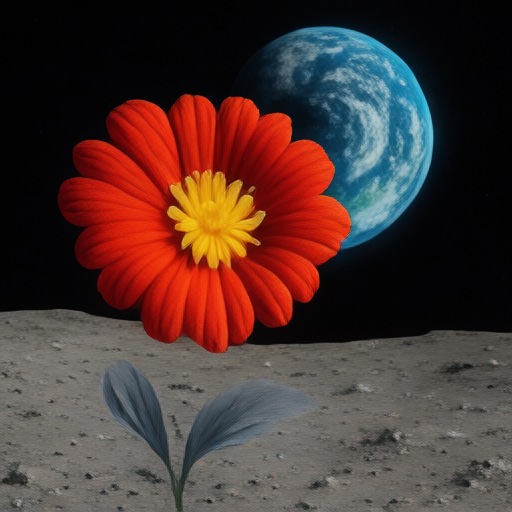}\end{minipage}\hspace{.01\linewidth}\begin{minipage}[c]{.265\linewidth}\raggedright\fontsize{5.6}{6.6}\selectfont\itshape A vibrant flower with large, crimson petals and a bright yellow center, standing in stark contrast to the moon's barren, grey surface. The flower's delicate texture is a surprising sight amidst the moon's craters and dust. In the background, the Earth can be seen rising, a swirl of blue and white, providing a breathtaking backdrop to this surreal lunar scene.\end{minipage}\\[4pt]
\begin{minipage}[c]{.215\linewidth}\includegraphics[width=\linewidth]{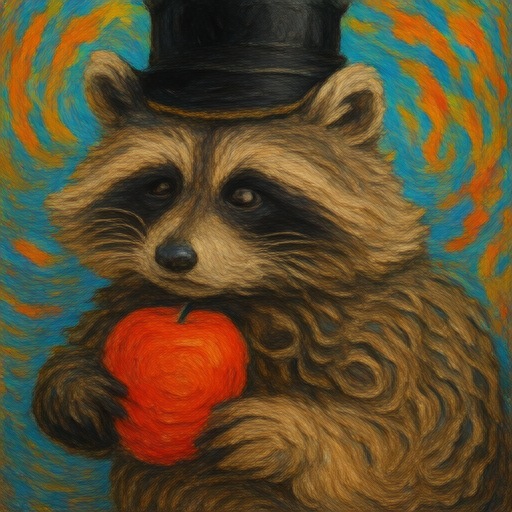}\end{minipage}\hspace{.01\linewidth}\begin{minipage}[c]{.265\linewidth}\raggedright\fontsize{5.6}{6.6}\selectfont\itshape A detailed oil painting that captures the essence of an elderly raccoon adorned with a distinguished black top hat. The raccoon's fur is depicted with textured, swirling strokes reminiscent of Van Gogh's signature style, and it clutches a bright red apple in its paws. The background swirls with vibrant colors, giving the impression of movement around the still figure of the raccoon.\end{minipage}\hfill%
\begin{minipage}[c]{.215\linewidth}\includegraphics[width=\linewidth]{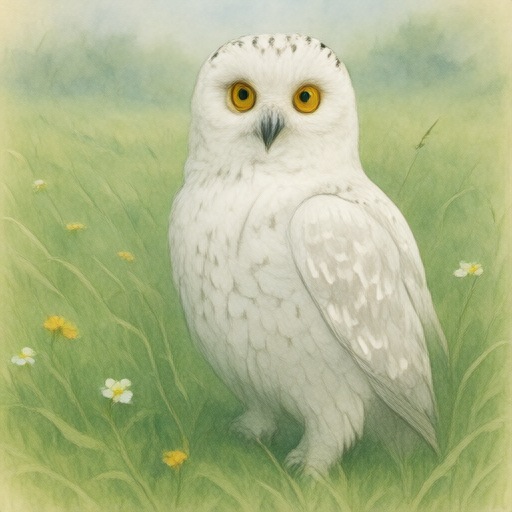}\end{minipage}\hspace{.01\linewidth}\begin{minipage}[c]{.265\linewidth}\raggedright\fontsize{5.6}{6.6}\selectfont\itshape A detailed watercolor painting that captures a majestic snowy owl with its pristine white feathers standing in the midst of a lush green field. The owl's bright yellow eyes are a stark contrast to the soft hues of the grass, and its feathers are intricately detailed, giving a sense of texture to the artwork. The field is dotted with wildflowers and the occasional blade of grass that sways gently, suggesting a light breeze in this tranquil scene.\end{minipage}\\[4pt]
\begin{minipage}[c]{.215\linewidth}\includegraphics[width=\linewidth]{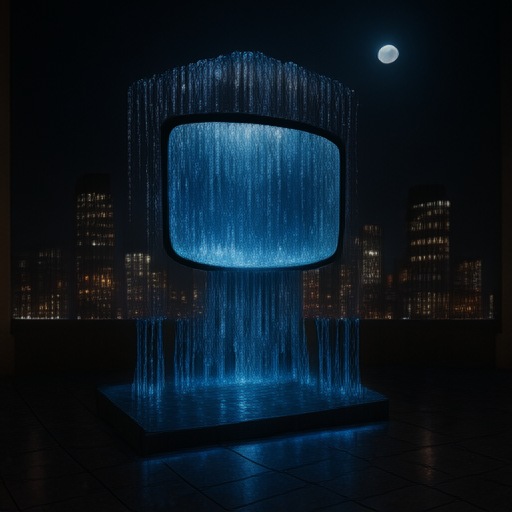}\end{minipage}\hspace{.01\linewidth}\begin{minipage}[c]{.265\linewidth}\raggedright\fontsize{5.6}{6.6}\selectfont\itshape An innovative water sculpture shaped like a flat-screen television stands in the center of a dimly lit room. The liquid crystal display, made entirely of cascading water, presents a luminous cityscape at night, with twinkling lights from skyscrapers and a reflection of the moon on the water's surface. Surrounding the unique installation, the floor is tiled in dark hues, accentuating the ethereal glow of the projected urban scene.\end{minipage}\hfill%
\begin{minipage}[c]{.215\linewidth}\includegraphics[width=\linewidth]{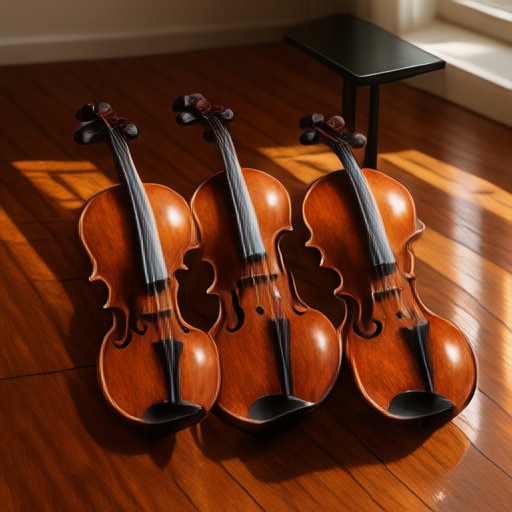}\end{minipage}\hspace{.01\linewidth}\begin{minipage}[c]{.265\linewidth}\raggedright\fontsize{5.6}{6.6}\selectfont\itshape Three violins with a glossy finish and delicate strings lying side by side on a polished hardwood floor. the instruments, each with a rich brown color and unique wood grain patterns, are positioned near a black music stand. natural light from a nearby window casts soft shadows around the violins, highlighting their elegant curves.\end{minipage}
\caption{\textbf{Samples for long prompts.} Each image is shown next to its full DPG-Bench prompt. Fine-tuned model of Table~\ref{tab:t2i} at $512^2$; for each prompt, the seed with the highest own image--text score among four.}
\label{fig:dpgsamples}
\end{figure}

\begin{figure}[tbp]\centering\scriptsize
\setlength{\tabcolsep}{3pt}\renewcommand{\arraystretch}{1.1}
\begin{tabularx}{\linewidth}{@{}m{1.55cm}X@{}}\toprule
Image & Verbatim caption from \model{} (DDPM-64, first sample)\\\midrule
\includegraphics[width=1.5cm]{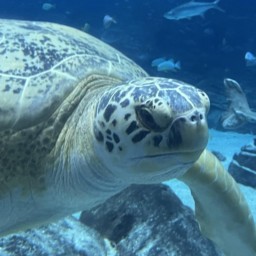} & A leatherwater sea turtle with its textured pattern and prominent sharp eyes rests in a blue ocean environment, surrounded by small fish.\\
\includegraphics[width=1.5cm]{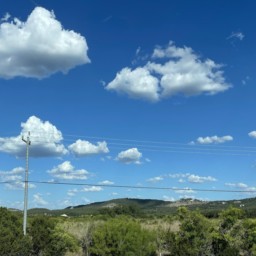} & A serene scene of a clear blue sky with fluffy white clouds, a lush green hillside, evoking tranquility and natural beauty beauty.\\
\includegraphics[width=1.5cm]{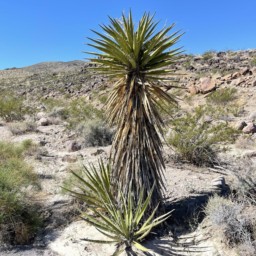} & A tall, spiky ta stands tall in a vast, sandy landscape, its needle-like leaves contrasting with the light and flat, rocky sand under a clear blue sky.\\
\includegraphics[width=1.5cm]{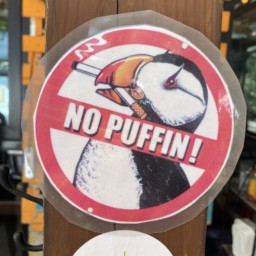} & A wooden sign with a stylized graphic of a gull wearing a helmet with the text ``no smoking'' above it. the sign is mounted on a wooden pole and there are two circular stickers with the text ``nonos'' in bold, capital letters. the background is blurred, but it appears to be a store setting with various light, possibly a window and a dining area. the sign is is likely sale, as indicated by the text anding on the,. it is not  legd.\\\bottomrule
\end{tabularx}\\[5pt]
\includegraphics[width=\linewidth]{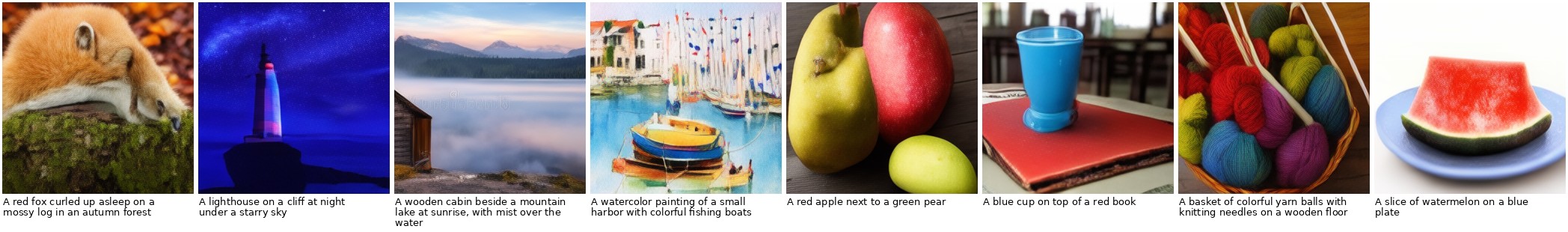}
\caption{\textbf{Both directions from one pretrained model.} \model{} after CC12M pretraining only. Top: verbatim captions of person-free DOCCI test images (the main-table DDPM-64 recipe, per-image seeds, first sample; defects such as repeated words are left in). Bottom: one of four seeds per prompt, chosen by the model's own image--text score, with the development-selected 64-step sampler\polish{ (web-watermark artifacts from pretraining data occasionally appear before synthetic fine-tuning)}.}
\label{fig:qualitative}
\end{figure}

Figure~\ref{fig:allseeds} shows all four seeds of every prompt behind Figure~\ref{fig:t2ibest}, without selection. Figure~\ref{fig:dpgsamples} shows samples for long prompts, and Figure~\ref{fig:qualitative} shows captions and images from the model after pretraining only.

\subsection{Own-score selection}
\label{app:ownselect}
Of GenEval's four samples per prompt, own-score selection keeps the one whose pooled image representation is closest (cosine) to the model's pooled prompt representation, with no external model. On the all-parameter fine-tune of Table~\ref{tab:ftfull}, it raises GenEval from .734 (four-sample average) to .764 (gain interval $[+.007, +.052]$), matching CLIP ViT-B/16 and SigLIP-B/16 rerankers (.765, .758); on the pretrained \model{} + Gram/iBOT, from .436 to .501 (four-sample average of a regenerated set; Table~\ref{tab:t2i}: .448).

\subsection{Frozen-feature readout with a language model}
\label{app:llmreadout}
\begin{table}[!htbp]\centering\small
\caption{\textbf{Frozen visual features read by a frozen language model.} Encoder $\to$ learned 2-layer MLP $\to$ frozen Qwen2.5-1.5B-Instruct \citep{qwen25}; one epoch of COCO captions, then one of VQAv2 \citep{goyal2017vqav2}. CIDEr-D: Karpathy test 5K; VQAv2: validation accuracy. External encoders: released weights, run by us.}
\label{tab:llmreadout}
\setlength{\tabcolsep}{3.5pt}
\begin{tabular}{@{}llrrrrr@{}}\toprule
&&&\multicolumn{4}{c}{VQAv2}\\\cmidrule(lr){4-7}
Frozen features & Pretraining data & CIDEr-D & overall & yes/no & number & other\\\midrule
No visual input (question only) &-- &-- &32.9 &-- &-- &--\\\midrule
DINOv2 ViT-B/14 &LVD-142M (images) &\oss{114.3} &\oss{64.6} &\oss{85.0} &\oss{45.5} &\oss{54.1}\\
\midrule
CLIP ViT-B/16 &CC12M (12M) &\oss{83.7} &\oss{54.6} &\oss{78.1} &\oss{37.5} &\oss{41.0}\\
SLIP ViT-B/16 &CC12M (12M) &\oss{101.3} &\oss{61.8} &\oss{83.0} &\oss{42.1} &\oss{50.8}\\
Contrastive only, $Z$ &CC12M-R (12M) &105.3 &63.0 &84.1 &44.8 &51.5\\
Contrastive only, $h$ &CC12M-R (12M) &108.5 &64.9 &85.1 &45.3 &54.4\\
\corerow \model{}, $Z$ &CC12M-R (12M) &108.1 &64.6 &85.0 &45.4 &54.0\\
\corerow \model{}, $h$ &CC12M-R (12M) &110.2 &65.6 &85.7 &46.0 &55.4\\
\corerow \model{} + Gram/iBOT, $Z$ &CC12M-R (12M) &109.9 &64.7 &85.1 &45.9 &54.1\\\bottomrule
\end{tabular}
\end{table}

Only the projector is trained, so row differences come from the visual features (the encoder is also frozen in \citealp{karamcheti2024prismatic,tong2024cambrian}). At equal data, \model{} is the better input on both taps ($+1.6$ VQAv2 and $+2.8$ CIDEr-D on $Z$; $+0.7$ and $+1.7$ on $h$); an independent implementation gives the same VQAv2 gap on $Z$ (63.2 vs.\ 61.6). On $h$, \model{} is above DINOv2-B, a frozen perceptual loss in our training (0.11\% of the encoder's gradient; Table~\ref{tab:gradshare}); on $Z$, it is level with DINOv2-B. With the language model trained as well (LoRA), \model{} stays above the contrastive-only control and DINOv2-B (VQAv2 70.0 vs.\ 69.3 and 69.8); with a frozen 7B language model reading $Z$, the order to the control is the same (70.0 vs.\ 68.7).

\subsection{Reading and steering a generation in $Z$}
\label{app:steer}
Because sampling runs in $Z$, probes trained on encoder outputs read every sampling state. With the fine-tuned denoiser, in 36 of 36 text-to-image trajectories every state from the first step on is nearest to the prompted ImageNet class (35 of 36 with the pretrained denoiser), and final states are as close to it as real images (cosine 0.81 vs.\ 0.82). The same probes steer sampling without training: at each step, the clean-state estimate $\hat Z_0$ moves down the gradient of the cross-entropy between the frozen ADE20K linear probe on $Z$ and a target layout. Steering the first 24 of 64 steps toward a layout sketch (hand-placed ADE20K labels on the $16\times16$ token grid; Figure~\ref{fig:steerstrip}), the image matches it on 83.1\% of pixels, against 49.6\% unsteered (four seeds; read by SLIP ViT-B/16 \citep{mu2022slip} with its own probe); steering all 64 steps follows weakly or loses realism.

\begin{table}[h]\centering\small
\caption{\textbf{Uses of one representation.} Probes and scores are defined on encoder outputs of real images and applied unchanged to generation; nothing is trained for these uses.}
\label{tab:uses}
\begin{tabular}{@{}p{.23\linewidth}p{.73\linewidth}@{}}\toprule
Use & Result\\\midrule
Probing sampling states & After the first denoising step, the zero-shot classifier recovers the prompted class for 54\% of the 424 (of 1{,}000) prompts whose object maps to an ImageNet class (\model{} + Gram/iBOT). A frozen language model with a projector trained on real-image $Z$ captions the final generated state better than the decoded and re-encoded image (CIDEr 93 vs.\ 82).\\
Self-ranking & Best of four samples by the model's own image--text score (\model{} + Gram/iBOT): GenEval .501 against .436 for the four-sample average of the same samples (95\% interval of the gain $[+.041, +.088]$).\\
Training-free guidance & A frozen ADE20K linear probe guides the early sampling steps toward a sketched layout: 83\% of pixels match the sketch, against 50\% without guidance (17 sketches, four prompts); 82\% vs.\ 42\% for \model{} + Gram/iBOT with its denoiser fine-tuned; 30\% $\to$ 55\% for layouts predicted on 200 reference images. On 5{,}000 COCO captions, guidance toward the layout of each caption's reference image lowers FID on these 5{,}000 images from 21.6 to 20.4 (paired 95\% interval of the change $[-1.7, -0.8]$).\\
Guidance + self-ranking & Keeping, of 16 seeds, the one the model scores closest to the sketch: 83.7\% $\to$ 87.6\% ($+3.9$, $[+2.5, +5.4]$); ranking at step 24 of 64 on the intermediate state: 86.7\%.\\\bottomrule
\end{tabular}
\end{table}

\section{\polish{Few-step distillation and inference cost}}
\label{app:distillation}
A continuous contextual representation admits a variable inference budget. We distill the \model{} + Gram/iBOT denoisers with the encoders fixed. Each student copies its modality's denoiser and adds a step-size embedding. Shortcut distillation trains dyadic jumps with a teacher-anchored finest level and EMA bootstrap targets; flow-map matching trains arbitrary start/end queries \citep{frans2025shortcut,boffi2025flowmap}. All students distill the guided, churn-free teacher ODE (guidance 7 for text, 3 for images); at inference the guidance is inside the student and each jump costs one denoiser evaluation.

\begin{table}[!htbp]\centering\small
\caption{\textbf{Adjusting inference computation.} Distilled students (own readout) versus the teacher (\model{} + Gram/iBOT) at matched NFE, which counts every denoiser evaluation including both guidance branches and the observable readout. FD-DINOv2 and CLIP-S on a 1K COCO draw (two seeds; a disjoint draw reproduces the students within 6.1 FD); GenEval on all 553 prompts $\times$ 4; SPICE on the first sentence of each caption (length-controlled; 1K images). Teacher rows use the best recipe found at each budget; the 64-step caption teacher is deterministic (with churn it reaches CLIP-S .791 and SPICE$_S$ 14.28); 
Wall-clock cost on one H100 is in Table~\ref{tab:latency}. FD-DINOv2 favors models trained with a DINOv2 loss, as the teacher and students all are, so it ranks rows within this study only.}
\label{tab:compute}
\setlength{\tabcolsep}{3.2pt}
\begin{tabular}{@{}lrrrr|lrrr@{}}\toprule
\multicolumn{5}{c|}{Text $\to$ image}&\multicolumn{4}{c}{Image $\to$ text}\\
Sampler & NFE & FD$_{\rm DINOv2}\!\downarrow$ & CLIP-S & GenEval & Sampler & NFE & CLIP-S & SPICE$_S$\\\midrule
Teacher, 2 steps & 5 & 913.4 & .752 & .309 & Teacher, 2 steps & 5 & .760 & 10.88\\
Teacher, 4 steps & 9 & 762.4 & .771 & .392 & Teacher, 4 steps & 9 & .778 & 12.37\\
Teacher, 64 steps & 129 & 679.6 & .774 & .433%
& Teacher, 64 steps & 129 & .782 & 13.08\\
\textbf{Student, 1 jump} & 2 & 931.6 & .745 & .317 & \textbf{Student, 1 jump} & 2 & .763 & 11.51\\
\textbf{Student, 4 jumps} & 5 & \textbf{775.1} & .764 & \textbf{.379} & \textbf{Student, 8 jumps} & 9 & \textbf{.781} & \textbf{13.43}\\
\textbf{Student, 8 jumps} & 9 & \textbf{730.6} & .773 & \textbf{.428} & \quad w/o KL anchor & 9 & .749 & --\\\bottomrule
\end{tabular}
\end{table}

\paragraph{Transport and output anchoring.}
For source and destination log-noise levels $\gamma_a>\gamma_b$, a predicted clean state defines the jump
\begin{equation}
 J_{a\to b}(u,\widehat z_0)
 =\alpha_b\widehat z_0+\frac{\sigma_b}{\sigma_a}(u-\alpha_a\widehat z_0).
\end{equation}
The student approximates teacher transport with a pseudo-Huber loss on clean states. Optional renoising replaces the implied noise by $\sqrt{1-r^2}\,\widehat\varepsilon+r\varepsilon'$ after non-final jumps; we use $r=1$ for multi-jump students. Because the observable head shares the denoiser trunk, its behavior can drift while the transport is fitted. We therefore add a readout anchor,
\begin{equation}
 \mathcal L_{\rm read}
 =\E_{(u,C)}\textstyle\sum_j\operatorname{KL}
 \big(p_{{\rm teacher},j}(\cdot\mid u,C)
 \,\|\,p_{{\rm student},j}(\cdot\mid u,C)\big)
\end{equation}
for text (weight 0.2), and the analogous pixel-space anchor for images (weight 1). Both heads are queried at the terminal noise level; half the anchored representations come from corrupted encoded observations and half from a detached one-jump student rollout. Text students use batch 2048, learning rate $10^{-4}$, and 20K training steps; image students use effective batch 2048 and 20K training steps. Tables report the final live weights at 20K training steps (the 14K checkpoint is 5--15 FD-DINOv2 better, but we keep the pre-chosen final checkpoint). An image student distilled in the same way from the fine-tuned model (teacher GenEval \nrev{.806}) reaches \nrev{.782, .799 and .797} with 2, 5 and 9 denoiser evaluations (Table~\ref{tab:t2i}); choosing among four samples with the model's own score gives \nrev{.824} at 9 evaluations (\nrev{.817} for the teacher under the same selection).

\paragraph{Evaluation.}
A fixed 1000-image draw from the COCO Karpathy test set with all five references, a disjoint 1000-image real control (FD-DINOv2 534.7, FID 58.1 against the draw), and a second disjoint data draw for replication. T2I prompts are the first reference captions. Student NFE is $k+1$ for $k$ jumps, including the observable readout; guided teacher NFE is $2s+1$ for $s$ steps. Teacher recipes (steps $\times$ guidance $\times$ churn) are swept separately at each budget. These 1K-image distances are dominated by finite-sample bias and are comparable only within this study.

\polish{Three findings. \emph{At a matched budget, the students beat the teacher's best sampler} (Tables~\ref{tab:compute} and~\ref{tab:distillscale}): on the 1K draw the eight-jump image student reaches FD-DINOv2 730.6 against 762.4 at 9 NFE, and at benchmark scale the four-jump student beats the teacher's best 5-NFE sampler on every metric; the 64-step stochastic teacher remains the quality ceiling. \emph{The readout anchor matters}: without it the eight-jump caption student's own-head CLIPScore falls from .781 to .749, and the pixel anchor improves FD-DINOv2 by 18--26 at every jump count; decoding the student latent with the teacher's readout is only 8--12 FD-DINOv2 better, so the student's own readout is not the main limit. \emph{A stronger-guidance target trades FID for alignment}: distilled from the CFG-7 trajectory, the 9-NFE student reaches GenEval .485, above every teacher sampler we scored (257-NFE teacher: .448; paired prompt bootstrap $+$.036, 95\% CI [.016, .057]), with a higher Inception FID.}

\begin{table}[h]\centering\small
\caption{\textbf{Matched-budget image generation on the main-table protocols.} Zero-shot COCO-30K (FID / FD-DINOv2 on the frozen test list), training-distribution CC12M-50K FID, GenEval (553 $\times$ 4) and DPG-Bench (1065 $\times$ 4); one sample per prompt. Teacher rows at 5, 9 and 17 NFE use the best recipe found at that budget on the 1K draw; the 128- and 257-NFE rows are the main-table recipes. Students: four jumps with the teacher's learned stops, one and eight jumps with linear stops (one jump is the same under both), own readout. \emph{CFG-7 target}: the same student recipe distilled from the CFG-7 teacher trajectory (10K training steps), linear stops at four and eight jumps. Bold: best at each budget. The 5--17-NFE teacher recipes and the students' stop schedules were selected on the 1K Karpathy-test draw, 752 of whose 1000 images are also in the COCO-30K test list (741 expected by chance).}
\label{tab:distillscale}
\setlength{\tabcolsep}{4pt}
\resizebox{\linewidth}{!}{%
\begin{tabular}{@{}llrrrrrr@{}}\toprule
 & & & \multicolumn{2}{c}{COCO-30K} & CC12M-50K & & \\
Model & Jumps / steps & NFE & FID$\downarrow$ & FD-DINOv2$\downarrow$ & FID$\downarrow$ & GenEval & DPG\\\midrule
Teacher & 2 steps, CFG 5 & 5 & 22.15 & 480.5 & 17.91 & .309 & 64.58\\
Teacher & 4 steps, CFG 7, churn 1 & 9 & \textbf{11.95} & 310.3 & 7.18 & .392 & 70.37\\
Teacher & 8 steps, CFG 7, churn 1 & 17 & 10.61 & 255.8 & 5.06 & .435 & --\\
Teacher & 64 steps, dev-selected & 128 & 10.20 & 221.5 & 4.37 & .433 & 73.20\\
Teacher & 128 steps, CFG 3, churn 1 & 257 & 13.66 & -- & -- & .448 & 73.49\\\midrule
Student & 1 jump & 2 & 16.06 & 495.7 & 9.53 & .317 & 67.38\\
Student & 4 jumps & 5 & \textbf{11.79} & 317.3 & \textbf{6.14} & .379 & 70.57\\
Student & 8 jumps & 9 & 12.45 & 279.1 & \textbf{6.07} & .428 & 72.09\\\midrule
Student, CFG-7 target & 1 jump & 2 & 15.38 & 463.0 & 8.89 & .329 & 67.53\\
Student, CFG-7 target & 4 jumps & 5 & 13.58 & \textbf{275.5} & 6.68 & \textbf{.457} & \textbf{72.16}\\
Student, CFG-7 target & 8 jumps & 9 & 15.04 & \textbf{259.2} & 7.21 & \textbf{.485} & \textbf{73.52}\\\bottomrule
\end{tabular}}\end{table}

\polish{Captions behave the same way: at 9 NFE the caption student is at the level of its 128-NFE teacher on the main-table protocols (SPICE 13.66 vs.\ 13.50; Table~\ref{tab:distillcap}), and one jump keeps most of it. On fixed hardware (Table~\ref{tab:latency}), the eight-jump caption student (79~ms at batch 1) is faster than BLIP-base (107~ms), Florence-2-base (136~ms), and CoCa-L/14 (352~ms), whose autoregressive decoders emit one token per step. Students add only the step embedding to the 383M base.}

\begin{table}[h]\centering\small
\caption{\textbf{Caption students on the protocols of Table~\ref{tab:i2t}} (\model{} + Gram/iBOT, pretrained).}
\label{tab:distillcap}
\setlength{\tabcolsep}{4pt}
\begin{tabular}{@{}lrrrrrcrr@{}}\toprule
 & NFE & SPICE & CIDEr-D & CHAIR$_i$ & RefCLIP-S & TrueMatch 3/5/7 & DetailCaps & DOCCI\\\midrule
Teacher, 64-step DDPM & 128 & 13.50 & 9.13 & 11.2 & .740 & 60.6/51.9/47.9 & 44.15 & 45.52\\
Student, 1 jump & 2 & 12.05 & 6.69 & 11.8 & .732 & 59.7/46.7/45.5 & 44.41 & 45.57\\
Student, 8 jumps & 9 & 13.66 & 11.32 & 11.9 & .737 & 58.9/50.0/48.4 & 44.86 & 46.54\\\bottomrule
\end{tabular}\end{table}

\begin{table}[h]\centering\small
\caption{\textbf{Fixed-hardware cost.} One dedicated H100 80GB, fp32 weights with TF32 matrix multiplies (the precision of all our evaluations), real COCO test images and captions, CUDA-synchronized wall-clock time, median of five (batch 1) or two (batch 64) calls after two warm-up calls. Latency at batch 1 includes guidance branches and the observable readout; encoding is listed separately because one encoded representation serves every task. External captioners use the decoding settings of Table~\ref{tab:i2t}.}
\label{tab:latency}
\setlength{\tabcolsep}{4pt}
\resizebox{\linewidth}{!}{%
\begin{tabular}{@{}lrrrr@{}}\toprule
Path & NFE & Latency, batch 1 (ms) & Throughput, batch 64 (/s) & Peak memory (GB)\\\midrule
\model{}: encode image & -- & 7.3 & 2284 & 3.1\\
\model{}: encode caption & -- & 8.2 & 4196 & 2.9\\\midrule
I2T teacher, DDPM-64, CFG 5 & 128 & 1079 & 18.1 & 9.5\\
I2T student, 8 jumps & 9 & 78.7 & 403 & 6.6\\
I2T student, 1 jump & 2 & 16.9 & 1848 & 6.5\\\midrule
T2I teacher, fixed 128 steps, CFG 3 & 257 & 4751 & 4.1 & 6.2\\
T2I teacher, dev-selected 64 steps & 128 & 2342 & 8.3 & 6.2\\
T2I student, 8 jumps & 9 & 172.5 & 117 & 6.2\\
T2I student, 4 jumps & 5 & 92.8 & 210 & 6.2\\
T2I student, 1 jump & 2 & 37.7 & 528 & 6.1\\\midrule
BLIP-base caption (beam 3) & -- & \oss{107} & \oss{72.8} & \oss{9.7}\\
Florence-2-base, \texttt{\textless CAPTION\textgreater} / \texttt{\textless DETAILED\_CAPTION\textgreater} & -- & \oss{136} / \oss{394} & \oss{26.2} / \oss{12.3} & \oss{14.9}\\
CoCa-L/14 caption & -- & \oss{352} & \oss{23.7} & \oss{7.2}\\\bottomrule
\end{tabular}}\end{table}

\section{Iterative contextual refinement}
\label{app:iteration}
\polish{As a supplementary study with frozen encoders, we test whether feeding back a step's prediction improves text sampling: re-encoding the predicted token posterior through the text encoder raises COCO SPICE by $+0.95$ (13.60 vs.\ 12.65) and lowers the non-word rate by 9\% at equal NFE, whereas revisiting the same noise level in an inner loop is interchangeable with ordinary steps at matched compute.}

The representation head produces a clean-state estimate at every step, which can be fed back as an additional input to the next denoiser evaluation through a zero-initialized projection (self-conditioning; \citealp{chen2023analogbits}). We compare three 15K-step continuations of \model{} + Gram/iBOT on CC12M with \emph{frozen encoders}: a matched control without self-conditioning; previous-step conditioning on the clean-state estimate $\widehat Z_{0}$; and previous-step conditioning on a \emph{re-encoded token posterior}, in which the predicted token probabilities form soft input embeddings that are passed through the frozen contextual text encoder. Self-conditioning is trained with probability 0.5 in both modalities. This input is separate from the self training route of Appendix~\ref{app:ablations}. Captions are scored on COCO test images with the native sampler (CFG 5), CLIPScore, SPICE, and the non-word rate $d$, the fraction of generated words with zero English word frequency (dropped-subword artifacts such as ``windowboard'').

\begin{table}[h]\centering\small
\caption{\textbf{Previous-step conditioning at equal NFE.} 500 COCO images, 32 steps (64 denoiser NFE), one caption per image, same seeds; SPICE from the COCO scorer on the full caption. Re-encoding adds one encoder pass per step.}
\label{tab:prevstep}
\begin{tabular}{@{}lrrrr@{}}\toprule
Conditioning (frozen encoders) & CLIP-S & $d\!\downarrow$ & Words & SPICE\\\midrule
Matched control &.7708 &.0694 &46.5 &12.65\\
Previous clean state $\widehat Z_0$ &.7751 &.0694 &47.0 &12.92\\
Re-encoded token posterior &\textbf{.7846} &\textbf{.0631} &49.1 &\textbf{13.60}\\\bottomrule
\end{tabular}\end{table}

Re-encoding raises CLIPScore by .0138 and lowers the non-word rate by 9\% over the matched control at no extra denoiser cost; feeding back $\widehat Z_0$ directly gains only .0043. Captions are 2.6 words longer, but the gain survives length control: scoring only the first sentence of each caption (equal length between the arms), re-encoding raises SPICE by 1.10 and CIDEr-D by 4.6 over the matched control (paired bootstrap 95\% intervals exclude zero; full captions: SPICE +0.95). The contrast between the two feedback paths is the interesting part: routing the model's prediction through its observable and back through its own contextual encoder is more useful than recycling the latent estimate, consistent with the observable being the stable coordinate system of an evolving representation (Section~\ref{sec:method}). The gain is specific to language: for images, previous-step conditioning raises FD-DINOv2 on 1K COCO prompts (680.5 $\to$ 777.6 for the re-encoded arm with 32 steps and churn 1).

\subsection{Loop depth versus sampling steps, with a compute control}
\label{app:loopdepth}
The re-encoded-posterior checkpoint above also supports an \emph{inner} loop: at each of the 32 outer diffusion steps, the same noise level is revisited $K$ times before advancing, each pass re-encoding the previous pass's token posterior. $K=1$ is the single-pass decode used everywhere else in this paper. Table~\ref{tab:loopnfe} asks whether this inner loop is worth its extra denoiser calls by comparing it against a control that spends the identical NFE as ordinary outer steps instead: at 128 total denoiser calls, 32 steps with $K=2$ scores .6550 against .6556 for 64 plain steps with no loop; at 256 calls, 32 steps with $K=4$ scores .6598 against .6592 for 128 plain steps. Every paired difference is within its bootstrap interval of zero. At matched compute the loop is interchangeable with ordinary steps: it is only occasionally, and never significantly, better.

\begin{table}[h]\centering\small
\caption{\textbf{Inner loops versus ordinary steps at equal compute.} Image-to-text, native sampler, self-conditioning carry on, 500 COCO images; score $=$ CLIPScore minus the per-caption rates of repeated words, dropped subwords, and invented names. Rows pair the inner-loop route against a plain-step control at the same total denoiser NFE.}
\label{tab:loopnfe}
\begin{tabular}{@{}rll@{}}\toprule
NFE & 32 steps $\times$ loop $K$ & plain steps, no loop\\\midrule
64 & 32$\times$1: .6493 & --\\
128 & 32$\times$2: .6550 & 64$\times$1: .6556\\
256 & 32$\times$4: .6598 & 128$\times$1: .6592\\\bottomrule
\end{tabular}\end{table}

\begin{figure}[!htbp]\centering\scriptsize
\setlength{\tabcolsep}{3pt}\renewcommand{\arraystretch}{1.15}
\begin{tabularx}{\linewidth}{@{}m{1.5cm}cX@{}}\toprule
Image & $K$ & Caption (re-encoded-posterior checkpoint, native sampler, 32 steps, CFG 5, carry on)\\\midrule
\includegraphics[width=1.4cm]{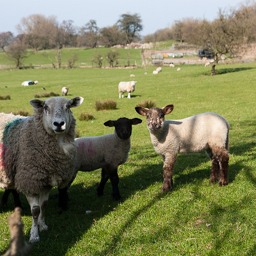} & 1 & Two sheep grazing in a green field.\\
 & 2 & Three sheep graze in a lush green field, surrounded by trees and clear sky.\\
 & 4, 8 & Three sheep graze in a lush green field, surrounded by trees and a clear blue sky.\\
\cmidrule(lr){1-3}
\includegraphics[width=1.4cm]{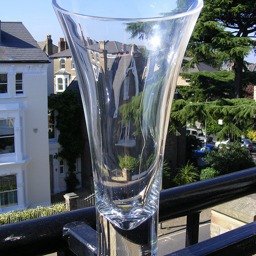} & 1 & A large, clear, cylindrical glass of water sits on a balcony, overlooking a city street scene with residential buildings and a clear blue sky.\\
 & 2, 4, 8 & A large, clear, cylindrical, clear vase sits on a balcony, overlooking a city street scene with residential buildings and a clear blue sky.\\
\cmidrule(lr){1-3}
\includegraphics[width=1.4cm]{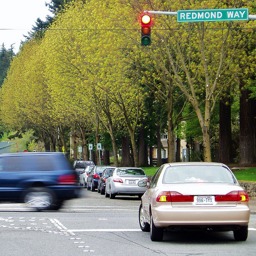} & 1 & A vibrant street scene on a city. the foreground is dominated by a white car, its its bluring of the right. the car's head ising. the road, a the,. thea white hues out on the right side of the road, marked with treesa the lights, adding a sense of the the., the right side of the scene. the is a traffic of a light, moving through the same direction. the trees are . the the side, their trees, adding a sense of order-madeity to the scene. the street is lined with several cars, their their their motion, rusting against the in the sky. the traffic sign, the's a green limit of visible, indicating thatas a the urban area. the sky overhead is a clear blue, suggesting a bright and sunny day and. below. overall, image encapsulates a typical of urban life, life where thea of of the life..\\
 & 2 & A street lined with cars, trees, and a green with a blurred red trafficwalk signal in the distance.\\
 & 4, 8 & A street lined with cars, trees, and street lights, with a blurred blue car driving motion in the distance.\\\bottomrule
\end{tabularx}
\caption{\textbf{Loop depth on three examples.} Ground truth (not shown): ``A herd of sheep are grazing in a green field''; ``A tall glass vase on a balcony''; ``A smaller car is stopped at the red light, as traffic drives on.'' Captions are verbatim, including their defects. Checkpoint: the frozen-encoder, re-encoded-token-posterior self-conditioning continuation of Appendix~\ref{app:iteration}. Quantitative compute control: Table~\ref{tab:loopnfe} and Appendix~\ref{app:loopdepth}.}
\label{fig:loopqual}
\end{figure}

Qualitatively, though, the two routes are not indistinguishable: since $K$ trades against ordinary steps only in aggregate score, a fixed 32-step budget with more inner loops still visibly changes individual captions, because the extra passes let a poor early posterior get re-encoded and corrected. Figure~\ref{fig:loopqual} shows this on three person-free COCO validation images (chosen from six person-free examples because each shows a clear change with $K$): $K=1$ can degenerate into repetition (a full paragraph of stutters on the street scene) or state a plausible but wrong detail (``glass of water'' for a vase, an undercounted ``two sheep''); by $K=2$ or $K=4$ the caption stabilizes and is usually more accurate. This is the same phenomenon Table~\ref{tab:loopnfe} scores in aggregate: more denoiser calls help, whether they arrive as loop depth or as steps, and the loop's only advantage is not needing a second solver step to spend them.

\section{Representation measurements}
\label{app:geometry}
\subsection{Readout ladder on fixed features}
For a declared target, retained information concerns the risk attainable from the representation by an unrestricted predictor; a fitted probe measures one practical readout of it. Table~\ref{tab:ladder} fits six readouts of increasing flexibility to the frozen contextual representation of every arm on the reduced-resolution ADE20K diagnostic (twofold-downsampled package, 8K training images, 512-pixel input, 40 epochs), all from one run per cell.

\begin{table}[h]\centering\small
\caption{\textbf{Readout ladder on the frozen contextual representation.} ADE20K mIoU, reduced-resolution diagnostic, one run per cell. The frozen-contrastive control shares the contrastive-only encoders; its row is an independent probe run on the same features, so the two rows differ only by probe-training noise.}
\label{tab:ladder}
\setlength{\tabcolsep}{3.5pt}
\begin{tabular}{@{}lrrrrrr@{}}\toprule
&\multicolumn{3}{c}{Linear SGD}&\multicolumn{2}{c}{Linear AdamW}&MLP\\\cmidrule(lr){2-4}\cmidrule(lr){5-6}
Model & raw & std. & ZCA & raw & ZCA & AdamW\\\midrule
Contrastive only &24.51 &25.24 &27.82 &27.30 &27.91 &27.21\\
Frozen contrastive $\to$ generation &24.50 &25.26 &27.79 &27.30 &27.90 &27.26\\
No contrastive loss &9.66 &15.49 &21.51 &18.26 &22.58 &23.48\\
\corerow \model{} ($\lambda_{\rm align}=1$) &15.83 &17.80 &23.04 &21.02 &23.59 &26.22\\
$\lambda_{\rm align}=0.25$ &12.21 &16.08 &20.91 &19.43 &22.02 &26.49\\
Self off, proportional &15.64 &17.31 &22.74 &20.77 &23.46 &26.78\\
Self off, to null &16.10 &18.00 &23.17 &20.84 &23.73 &26.62\\
Contrastive only + Gram/iBOT &27.54 &28.26 &31.00 &31.10 &31.26 &29.76\\
\corerow \model{} + Gram/iBOT &16.53 &18.59 &24.53 &22.69 &25.11 &28.51\\\bottomrule
\end{tabular}
\end{table}

Every generative arm gains more from a flexible readout than the contrastive controls do: the gap to the control shrinks from 8.67 (linear SGD) to 0.99 (MLP) for \model{}, and from 11.0 to 1.25 between the two Gram/iBOT models. In this diagnostic, the contrastive controls score higher with their linear probe than with the MLP, so the gaps at each model's best probe are 1.69 and 2.75. Table~\ref{tab:ladderfull} repeats the comparison under the full-resolution protocol, with probe seeds. Whitening is an invertible affine preprocessing, so its gain reflects regularized finite fitting rather than new information; the MLP additionally changes capacity. A matched study across feature taps (backbone, projection before LayerNorm, and final representation) gives best-readout gaps to the control of 1.55, 1.42, and 1.47 mIoU, while the best purely linear gap grows from 1.55 at the backbone to 4.32 at the representation: the narrow contextual representation is where linear accessibility is lost.

\subsection{Neighborhoods and connectivity}
\paragraph{Fixed-rank enrichment.} Normalize each nonzero contextual token to unit length. For a token $i$ with label $s_i$ (its source image, class, word, or word family), let $\mathcal N_k(i)$ be its $k$ nearest tokens within an eligible set $\mathcal E(i)$, and define
\begin{equation}
 E_s(k)=\frac1T\sum_{i}\frac{|\{j\in\mathcal N_k(i):s_j=s_i\}|/k}{b_s(i)},\qquad
 b_s(i)=\frac{|\{j\in\mathcal E(i):s_j=s_i\}|}{|\mathcal E(i)|},
\end{equation}
so that 1 is chance and $1/b_s$ the ceiling. The eligible sets exclude the trivial matches: all other tokens for the same image, tokens of other images for the class, tokens of other captions for the word, and tokens of other captions and other words for the family. Every model contributes exactly $k$ neighbors per token, so peaks at different radii or support cannot confound the comparison. The vision sample takes 24 ImageNet classes, 10 images per class, and 25 random patches per image (6000 tokens; each token has 24 same-image and 225 same-class, other-image companions), over three draws. The text sample is frequency-matched: at most 60 occurrences of each of 51 single-token words in seven families, one draw of 2711 occurrences shared by every model. As a scale check, the mean cosine between two patches of the same image is .274 for contrastive only and .025 for \model{} (.055 and .002 for different classes): \model{}'s patch representations are also more spread overall, which rank-based enrichment is insensitive to.

\paragraph{Sampled connectivity and relative detour.}
Construct a union-symmetrized $k$-nearest-neighbor graph on normalized tokens with Euclidean edge weights. $M$ is the largest-component fraction. From 48 sampled source vertices, compute shortest-path lengths $h_{ij}$; for reachable pairs with $d_{ij}>10^{-6}$, set $\rho_{ij}=h_{ij}/d_{ij}$ and
\begin{equation}
 D=\frac{\operatorname{median}_{s_i\ne s_j,\,\mathrm{reachable}}\rho_{ij}}
 {\operatorname{median}_{s_i=s_j,\,\mathrm{reachable}}\rho_{ij}}.
\end{equation}
$U$ is the unreachable between-label fraction over the same source/target pairs. The graphs use the frequency-matched text occurrences (1500 and 2500 occurrences $\times$ three seeds per $k$). Visual graphs are fully connected ($M=1$) for both models. A graph vertex is a token occurrence whereas a contextual representation is a whole sequence, so these statistics describe the sampled token cloud.

\begin{table}[!htbp]\centering\small
\caption{\textbf{Geometry at fixed rank.} Left: vision enrichment over chance among each patch's $k$ nearest patches (same image) and among its $k$ nearest patches of other images (same class); mean over three draws, s.d.\ ${\le}1.2$ and ${\le}0.31$. Middle: text enrichment among the ten nearest tokens of other captions (same word; ceiling 52.3) and of other captions and words (same family; ceiling 8.0). Right: text token graphs, mean $\pm$ s.d.\ over six draws; $M$ largest-component fraction, $U$ unreachable between-family fraction, $D$ relative between-family detour.}
\label{tab:geomsummary}\label{tab:graphrobust}
\setlength{\tabcolsep}{2.6pt}
\resizebox{\linewidth}{!}{%
\begin{tabular}{@{}lcccc|cc|ccccc@{}}\toprule
&\multicolumn{4}{c|}{Vision patches}&\multicolumn{2}{c|}{Text tokens}&\multicolumn{5}{c}{Text token graph}\\
&\multicolumn{2}{c}{Same image}&\multicolumn{2}{c|}{Same class}&Word&Family&\multicolumn{3}{c}{$k=12$}&$D$, $k=6$&$D$, $k=24$\\\cmidrule(lr){2-3}\cmidrule(lr){4-5}\cmidrule(lr){8-10}
Model & $k=10$ & $k=50$ & $k=10$ & $k=50$ & $k=10$ & $k=10$ & $M\uparrow$ & $U\!\downarrow$ & $D\!\downarrow$ & & \\\midrule
Contrastive only & 83.3 & 27.9 & 9.01 & 6.69 & 51.4 & 5.88 & .68$\pm$.02 & .53$\pm$.04 & 1.86$\pm$.12 & 2.77$\pm$.29 & 1.49$\pm$.06\\
\quad + Gram/iBOT & 86.8 & 29.5 & 9.33 & 6.93 & 51.4 & 5.92 & .71$\pm$.10 & .51$\pm$.16 & 1.73$\pm$.10 & 2.56$\pm$.49 & 1.40$\pm$.05\\\midrule
\corerow \model{} & 6.5 & 3.0 & 2.01 & 1.48 & 50.8 & 5.80 & .97$\pm$.03 & .08$\pm$.05 & 1.24$\pm$.07 & 1.38$\pm$.04 & 1.25$\pm$.04\\
\corerow \quad + Gram/iBOT & 6.7 & 3.1 & 2.15 & 1.54 & 50.5 & 5.81 & .98$\pm$.01 & .05$\pm$.05 & 1.13$\pm$.05 & 1.18$\pm$.05 & 1.21$\pm$.05\\\bottomrule
\end{tabular}}\end{table}

Under equal support the vision and text pictures differ (Table~\ref{tab:geomsummary}). Contrastive patch representations cluster by their source image, with a third of each patch's ten nearest patches from the same image, and by class; \model{}'s patches are far less tied to either, yet remain above chance, and Gram/iBOT, which raises backbone ADE20K mIoU by about two points (Table~\ref{tab:understanding}), leaves both unchanged. Text tokens keep same-word and same-family neighbors at the control's rate in every model, while both joint models form a nearly connected cloud with shorter between-family detours at every $k$; Gram/iBOT moves neither family appreciably, so connectivity is a property of the generative objective, not of the structural regularizer.

\subsection{Contextual sense separation}
\label{app:polysemy}
\begin{table}[!htbp]\centering\small
\caption{\textbf{Contextual sense separation.} Leave-one-out 1-NN accuracy (cosine) on COCO captions for six polysemous words (senses balanced; sense chance 47\%). \emph{Word}: 1-NN word identity among all target occurrences of 20 word--sense groups. \emph{Domain}: the nearest occurrence of a \emph{different} word shares the meaning domain (e.g.\ \emph{plate}/home plate $\to$ \emph{pitcher}/baseball; chance 11\%). representation $=Z_t$; backbone $=h_t$ before projection.}
\label{tab:polysemy}
\setlength{\tabcolsep}{4pt}
\begin{tabular}{@{}lrrrrrr@{}}\toprule
&\multicolumn{3}{c}{contextual representation $Z_t$}&\multicolumn{3}{c}{Backbone $h_t$}\\\cmidrule(lr){2-4}\cmidrule(lr){5-7}
Text tower & Sense & Word & Domain & Sense & Word & Domain\\\midrule
Lookup embedding (any model) & 47.2 & 100.0 & -- & 47.2 & 100.0 & --\\
Contrastive only & 97.9 & 99.9 & 43.2 & 97.9 & 100.0 & 49.2\\
No contrastive loss & 73.9 & 52.9 & 9.5 & 79.2 & 87.9 & 9.2\\
\corerow \textbf{\model{}} & \textbf{98.8} & \textbf{99.9} & 23.2 & 99.0 & 100.0 & 45.3\\
\corerow \textbf{\model{} + Gram/iBOT} & 97.3 & 100.0 & 20.0 & 97.8 & 100.0 & 39.3\\\bottomrule
\end{tabular}
\end{table}
We mine the 202K COCO val2014 captions for polysemous single-token words whose senses are identified by unambiguous cue words (e.g.\ \emph{trunk}: \{elephant\}, \{tree, log, stump\}, \{car, luggage, suitcase\}), keeping a caption for a sense only if it contains the target exactly once, cues of that sense, and no cue of another sense. Senses with at least 20 captions are kept (at most 100 per sense): \emph{plate} (dish, home plate), \emph{pitcher} (baseball, jug), \emph{trunk} (elephant, tree, car), \emph{glasses} (eyewear, drinkware), \emph{apple} (fruit, computer), and \emph{mouse} (computer, animal), plus single-sense groups for \emph{keyboard}, \emph{bank}, \emph{palm}, \emph{net}, \emph{bat}, \emph{fan}, and \emph{bow} that serve as cross-word neighbors (20 groups; 1,520 occurrences in which the word is a single token). For every occurrence we read the lookup embedding, the text backbone output $h_t$, and the representation $Z_t$. Sense accuracy is balanced leave-one-out 1-NN accuracy within a word, averaged over the six words (chance 47.2\%); a logistic probe trained on half the images and tested on the other half gives the same ordering. Word accuracy is 1-NN word identity over all occurrences. Domain accuracy asks whether the nearest occurrence of a \emph{different} word shares the meaning domain (baseball, dining, computing, animal, vehicle, plant, wearable; chance 11.1\%). Because senses are defined by cue words present in the caption, sense separation shows that the token representation absorbs its context; the comparison across objectives, not the absolute level, is the informative part.

\applefootnote{ \textcolor{textgray}{\sffamily Apple and the Apple logo are trademarks of Apple Inc., registered in the U.S. and other countries and regions.}}

\end{document}